\documentclass[10pt,twocolumn,letterpaper]{article}

\usepackage{amsmath,amsfonts,bm}

\def\eqref#1{equation~\ref{#1}}

\def\1{\bm{1}}

\def\vs{{\bm{s}}}

\DeclareMathAlphabet{\mathsfit}{\encodingdefault}{\sfdefault}{m}{sl}
\SetMathAlphabet{\mathsfit}{bold}{\encodingdefault}{\sfdefault}{bx}{n}

\usepackage[pagenumbers]{cvpr}

\usepackage{graphicx}
\usepackage{amsmath}
\usepackage{amssymb}
\usepackage{booktabs}
\usepackage[table]{xcolor}
\usepackage[super]{nth}
\usepackage{enumitem}
\usepackage[english]{babel}
\usepackage{blindtext}
\usepackage{multirow}
\usepackage{tabularx}
\usepackage{fancyvrb}
\usepackage[T1]{fontenc}
\usepackage{multicol}
\usepackage{listings}
\usepackage[pagebackref,breaklinks=true,colorlinks,citecolor=citecolor2,urlcolor=gray,linkcolor=linkcolor,bookmarks=false]{hyperref}

\usepackage[capitalize]{cleveref}
\crefname{section}{\S}{\S}
\Crefname{section}{Section}{Sections}
\Crefname{table}{Table}{Tables}
\crefname{table}{Tab.}{Tabs.}

\newlength\savewidth\newcommand\shline{\noalign{\global\savewidth\arrayrulewidth
\global\arrayrulewidth 1pt}\hline\noalign{\global\arrayrulewidth\savewidth}}
\newcommand{\tablestyle}[2]{\setlength{\tabcolsep}{#1}\renewcommand{\arraystretch}{#2}\centering\small}
\makeatletter\renewcommand\paragraph{\@startsection{paragraph}{4}{\z@}
{.4em \@plus1ex \@minus.2ex}{-.5em}{\normalfont\normalsize\bfseries}}\makeatother

\newcolumntype{q}[1]{>{\centering\arraybackslash}p{#1}}
\newcolumntype{x}[1]{>{\centering\arraybackslash}p{#1pt}}
\newcolumntype{k}[1]{>{\raggedright\arraybackslash}p{#1}}
\newcolumntype{y}[1]{>{\raggedright\arraybackslash}p{#1pt}}
\newcolumntype{v}[1]{>{\raggedleft\arraybackslash}p{#1}}
\newcolumntype{z}[1]{>{\raggedleft\arraybackslash}p{#1pt}}

\def\x{$\times$}
\definecolor{linkcolor}{HTML}{ED1C24}
\definecolor{citecolor}{RGB}{34,139,34}
\definecolor{citecolor2}{HTML}{0071bc}
\definecolor{defaultcolor}{gray}{0.9}
\definecolor{demphcolor}{gray}{.5}
\newcommand{\demph}[1]{\textcolor{demphcolor}{#1}}
\definecolor{baselinecolor}{gray}{.9}
\newcommand{\baseline}[1]{\cellcolor{baselinecolor}{#1}}

\begin{document}

\title{De-Diffusion Makes Text a Strong Cross-Modal Interface}
\author{
Chen Wei\textsuperscript{1,2}\quad Chenxi Liu\textsuperscript{1}\quad Siyuan Qiao\textsuperscript{1}\quad Zhishuai Zhang\textsuperscript{1}\quad Alan Yuille\textsuperscript{2}\quad Jiahui Yu\textsuperscript{1}
\\[0.5em]
\textsuperscript{1}Google DeepMind \qquad \textsuperscript{2}Johns Hopkins University
\vspace{-0.5em}
}

\twocolumn[{
\renewcommand\twocolumn[1][]{#1}
\maketitle

\begin{center}
\centering
\includegraphics[width=\linewidth]{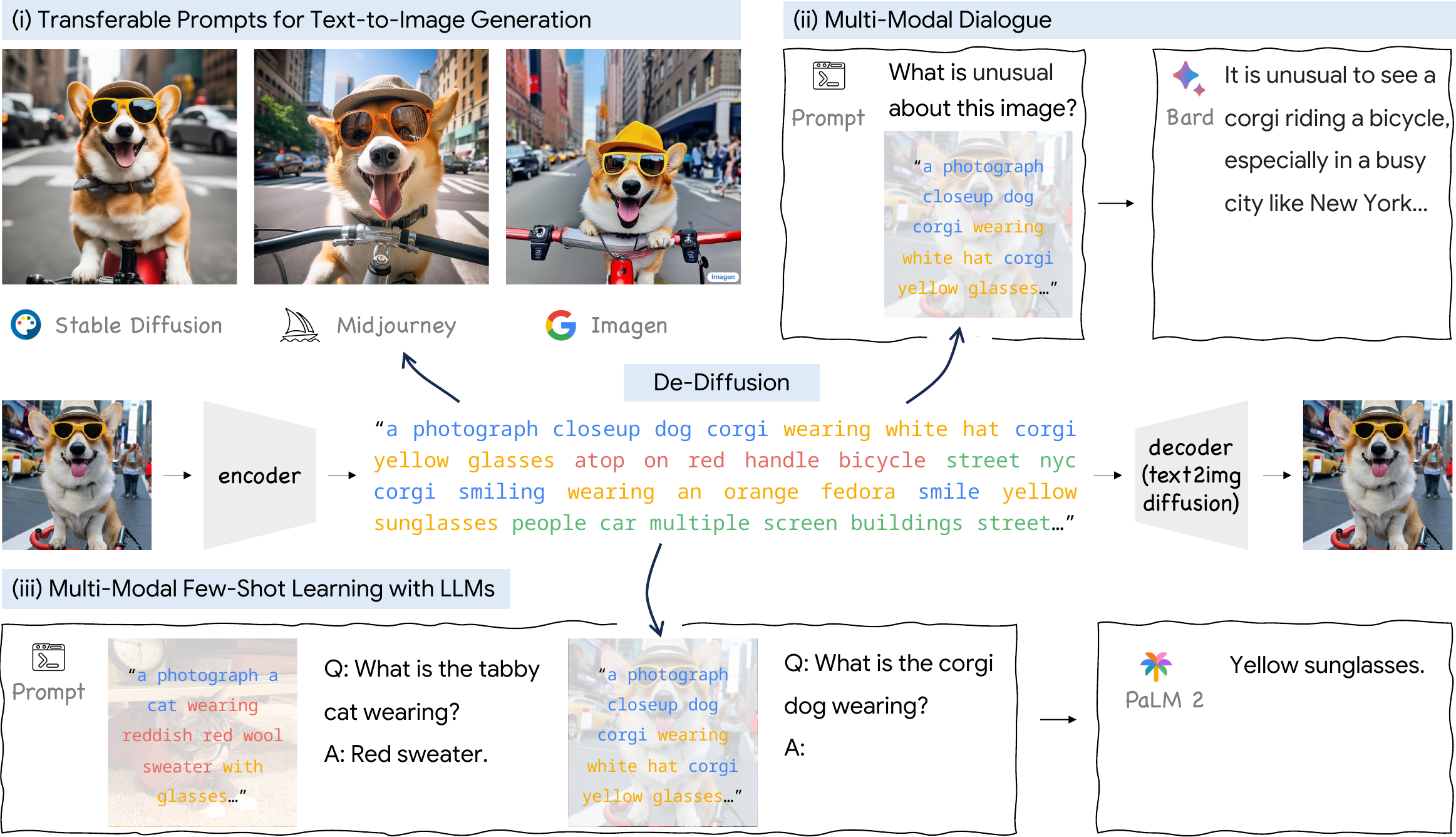}
\vspace{-15pt}
\captionof{figure}{
\small
De-Diffusion is an autoencoder whose decoder is a pre-trained text-to-image diffusion model. It encodes an input image into a piece of information-rich text, which mixes comprehensive semantic concepts present in the image to be a ``scrambled caption''. We group semantics by color for illustration. De-Diffusion text can act as a flexible interface between different modalities, for example, enabling diverse vision-language applications including: (\textit{i}) providing transferable prompts for different text-to-image tools, (\textit{ii}) enabling text-only chatbots, \eg, Bard~\cite{bard}, to engage in multi-modal dialogue, and (\textit{iii}) injecting image context into off-the-shelf large language models (LLMs), \eg, PaLM 2~\cite{palm2}, to perform open-ended visual question answering by prompting the LLM with few-shot examples.
}
\label{fig:teaser}
\end{center}
\vspace{14pt}
}]
\maketitle

\begin{abstract}
We demonstrate text as a strong cross-modal interface.
Rather than relying on deep embeddings to connect image and language as the interface representation, our approach represents an image as text, from which we enjoy the interpretability and flexibility inherent to natural language. We employ an autoencoder that uses a pre-trained text-to-image diffusion model for decoding. The encoder is trained to transform an input image into text, which is then fed into the fixed text-to-image diffusion decoder to reconstruct the original input -- a process we term \emph{De-Diffusion}. Experiments validate both the precision and comprehensiveness of De-Diffusion text representing images, such that it can be readily ingested by off-the-shelf text-to-image tools and LLMs for diverse multi-modal tasks. For example, a single De-Diffusion model can generalize to provide transferable prompts for different text-to-image tools, and also achieves a new state of the art on open-ended vision-language tasks by simply prompting large language models with few-shot examples.
Project page: \href{https://dediffusion.github.io/}{dediffusion.github.io}.
\end{abstract}

\section{Introduction}
We have witnessed LLM-powered products such as ChatGPT taking over the world by storm.
Nowadays many people are convinced of the benefits that LLMs can bring in understanding natural language conversations and assisting humans in creative tasks. However, what is the path forward? One clear direction and trend is towards \emph{multi-modality}, allowing the model to understand additional modalities such as image, video, and audio. GPT-4~\cite{gpt4} is a multi-modal model with impressive image understanding capabilities, and has recently rolled out to the public together with audio-processing capabilities. Gemini is also ``multi-modal from day one''~\cite{gemini}. Multi-modal models like these have a fundamental design choice to make, \ie, how different modalities should communicate and connect? In the context of this work, we rephrase the question as: what is the \emph{cross-modal interface}?

We argue that a good cross-modal interface should at least possess the following two properties: (1) \emph{content preserving}, \ie, signals from the original modality can be reconstructed from the interface representation to a high degree; (2) \emph{semantically meaningful}, \ie, the interface representation contains useful abstractions of the raw signals, so that understanding and reasoning can be performed more easily. Balancing these two properties is challenging, and in fact they can often be in contention with each other. For example, the raw signals from the original modality satisfy content preserving perfectly, but are lacking on the semantically meaningful front.

Ever since the deep learning era~\cite{dbn,hinton2006reducing,alexnet,imagenet}, \emph{deep embeddings} have been the go-to choice as cross-modal interface. They can be good at preserving image pixels if trained as an autoencoder~\cite{hinton2006reducing}, and can also be semantically meaningful, with the most recent exemplar being CLIP~\cite{clip}. In this paper, we do not argue that deep embeddings are a bad cross-modal interface \emph{per se}, but instead convey the idea that according to our experiments, \emph{text} can be a strong alternative cross-modal interface.

If we consider the relationship between the speech and text for a quick second, text has always been so natural of a cross-modal interface that we do not typically think of it as such. Converting the speech audio to text well \textit{preserves the content} such that we can reconstruct the speech audio with the mature text-to-speech technique. We are also confident that the transcribed text contains all the semantics information, in other words, \textit{semantically meaningful}. By analogy, we can also ``transcribe'' an image into text, which has the more familiar name of image captioning.
But when we compare typical image captions against the two properties of cross-modal interface, they do not preserve content well but only capture the most salient semantic concepts. In other words, image captions are more about precision than comprehensiveness~\cite{cococap,xie2021considered}, and it is hard to answer any and all visual questions from the short captions (\eg,~\cref{fig:viz-coco-1}).

While image captions do not make an ideal interface representation, we argue that precise \textit{and} comprehensive text, if attainable, remains a promising option, both intuitively and practically. Intuitively, humans rely on language to articulate our physical surroundings, engage in reasoning, and deliver solutions. In other words, we constantly ``transcribe'' information about the external world into language and use it as an interface for higher-level cognition~\cite{fodor1975language,chomsky2006language}. Practically, text is the native input domain for LLMs. Using text as the interface can avoid the need for adaptive training often required with deep embeddings~\cite{blip-2,flamingo}. Given that training and adapting top-performing LLMs can be prohibitively expensive~\cite{gpt4,palm2,flamingo}, text provides a modular design that opens up more possibilities. The question is, how can we attain precise and comprehensive text of images?

We resort to the classic autoencoding for a solution~\cite{hinton2006reducing}. Unlike common autoencoders, we utilize a pre-trained text-to-image diffusion model as the decoder, and naturally, with text as the latent space. The encoder is trained to transform an input image into text, which is then fed into the text-to-image diffusion model for decoding. To minimize the reconstruct error, the latent text, though often mixing semantic concepts together to be a ``scrambled caption'' of the input image, has to be both precise and comprehensive. No extra supervision is used other than images themselves.

Recent generative text-to-image models excel at converting arbitrary rich text of, \eg, tens of words, to highly detailed images that closely follow the prompts~\cite{glide,dalle-2,parti,imagen,stable}.
This essentially suggests the remarkable capability of these generative models to process complex text into visually coherent outputs. 
By employing one of these generative text-to-image models as the decoder, the optimized encoder explores the wide latent space of text and unpacks the enormous visual-language knowledge encapsulated within the generative model, embodying a foundational paradigm known as Analysis by Synthesis~\cite{atal1971speech,yuille2006vision,bever2010analysis}.

We show De-Diffusion text extensively captures semantic concepts in images, and, when used as text prompts, enables diverse vision-language applications (\cref{fig:teaser}). De-Diffusion text can generalize to be a transferable prompt for different text-to-image tools. Evaluated quantitatively by reconstruction FID~\cite{fid}, De-Diffusion text significantly outperforms human-annotated captions as prompts to a third-party text-to-image model~\cite{stable}. De-Diffusion text also enables off-the-shelf LLMs to conduct open-ended vision-language tasks by simply prompting LLMs with few-shot task-specific examples. We highlight De-Diffusion outperforms Flamingo~\cite{flamingo} on open-ended few-shot VQA~\cite{antol2015vqa} with 100$\times$ fewer learnable weights and without using interleaved image-text supervision. The results demonstrate De-Diffusion text effectively interconnects both human interpretations and various off-the-shelf models across domains. 

\section{Related Work}
\paragraph{Autoencoding} is a classical approach for learning representations~\cite{hinton2006reducing,autoencoder}. It uses an encoder to map the input into a compressed, meaningful representation, and a decoder to reconstruct the input from this representation to be as close as possible to the original. This simple autoencoding concept underpins many unsupervised representation learning algorithms across domains~\cite{hinton1993autoencoders,denoisingae,bert,vae,mae}. By forcing the model to compress then reconstruct the input, autoencoders discover useful structural representations of the data. For example, Neural De-Rendering~\cite{de-rendering} is a generalized autoencoder that utilizes a deterministic rendering function as the decoder and maps images into structured and disentangled scene descriptions. Inspired by its name ``de-rendering'', we name our approach ``De-Diffusion''. 

A specific type of autoencoder, VQ-VAE~\cite{vqvae,vqvae2} or discrete VAE~\cite{dall-e}, is designed to learn discrete, structured representations in the latent space. This can be especially useful for modeling data with categorical or symbolic attributes. These methods are now widely adopted in multi-modal models to tokenize images~\cite{dall-e,parti,vqgan,stable}. However, VQ-VAE's latent space is hidden and often entangled, requiring adaptive fine-tuning for downstream tasks. De-Diffusion also utilizes a discrete latent space. In contrast, we directly encode images into a sequence of text, which is directly interpretable. 

SPAE~\cite{spae} and LQAE~\cite{lqae} are two recent approaches that encode images into the vocabulary space of a fixed LLM. They jointly learn the encoder and decoder from scratch. Consequently, although the latent space is discrete text, it tends to act as a ``cipher code'' that only the co-trained decoder can interpret. This limits generalization to human understanding and off-the-shelf LLMs and text-to-image models. In contrast, De-Diffusion utilizes a pre-trained text-to-image diffusion model as the decoder, obtaining interpretable text as the latent representation.

\paragraph{How many words is an image worth?}
The adage ``a picture is worth a thousand words'' means that still images can convey complex and sometimes multiple ideas more effectively than a mere verbal description. Indeed, a single image can tell a story that would take many words to explain. The question, how many words is an image worth, is constantly explored by the computer vision community~\cite{fidler2013sentence,gabbay2021image,gal2022image,liu2023picture}. For example, ``An image is worth 16\x16 words'', or ViT~\cite{vit}, proposes to take the image patches as tokens (words) and process these tokens by Transformers~\cite{transformer}, which has become one of the standard vision backbones now. In this sense, our work can also been seen as ``An image is worth 75 words'', for we encode input images into a sequence of 75 tokens.

Several prior works also explore to use text to represent images~\cite{pica,lens} and combine with with LLMs. However, these works rely on multiple captioning and classification models, whose outputs are concatenated to be the text representation. Their performance is heavily dependent on the captioning and classification models, and we demonstrate in \cref{sec:applications} that even human-annotation captions can lack the extensive details covered in De-Diffusion text.

\paragraph{Vision-language models.} The breakthrough in NLP~\cite{bert,gpt-3,gpt4,t5,chinchilla,howard2018universal,flan}, especially their abilities to perform few-shot learning, has inspired a large body of vision-language work. A family of vision-language models is based on contrastive learning~\cite{contrastive}, where images and text are projected in to a same embedding space~\cite{clip,jain2021mural,align,ablef,pham2023combined,lit,coca}. De-Diffusion differs from contrastive models as we encode image as text, instead of deep embeddings. Another family of vision-language models fuses vision and language models by jointly training them with large-scale image-text data~\cite{flamingo,coca,idefics,anymal,dall-e,yu2023scaling,llava,pali}. In contrast, De-Diffusion takes a modular design with text as the representation, bypassing the heavy cost image-text data collection and jointly training large-scale vision and language models.

\section{Method}
\subsection{De-Diffusion for Text Representation}
\label{sec:de-diffusion}
\paragraph{Autoencoder.} Autoencoding is one of the classical methods for representation learning~\cite{hinton2006reducing,autoencoder}. An autoencoder first encodes an input $x$ into a latent representation $z$, then decodes $z$ back to $\tilde{x}$ for reconstruction. Both the encoder and the decoder are optimized so that the reconstructed input $\tilde{x}$ is as similar as possible to the original input $x$. By doing so, the compressed representation $z$ preserves the information in the input. Since no more supervision is required except the input itself, autoencoding is an unsupervised approach without the heavy burden of human annotation.

\begin{figure*}[t]
\centering
\includegraphics[width=0.7\linewidth]{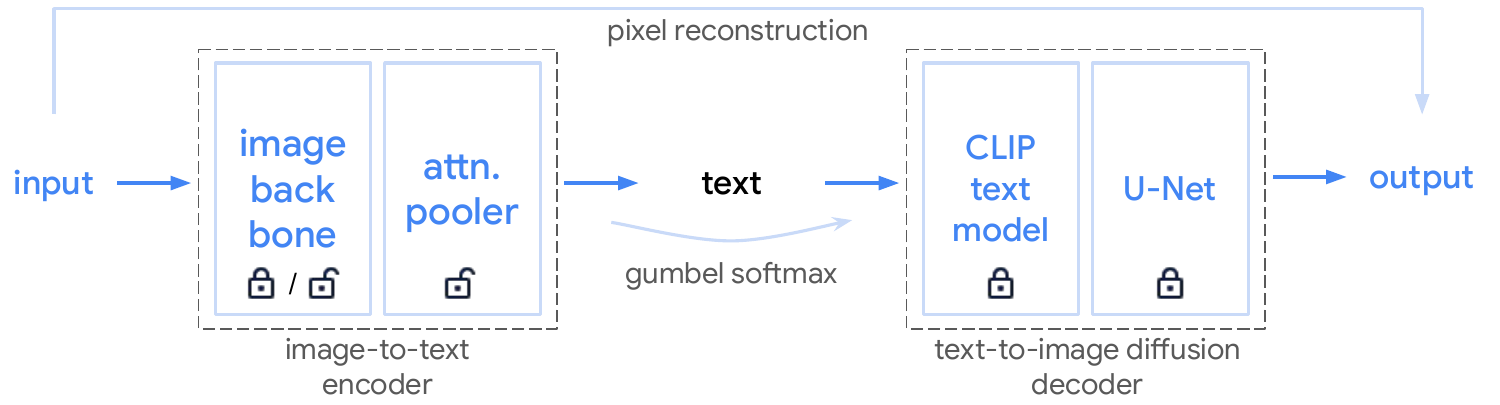}
\vspace{-7pt}
\caption{
\textbf{Architecture of De-Diffusion.}
The overall structure is an autoencoder, with (\textit{i}) a pre-trained text-to-image diffusion model as the decoder, (\textit{ii}) text as the latent representation, and (\textit{iii}) a image-to-text encoder consisting of a image backbone and an attentional pooler. Lock and unlock denote frozen and learnable weights, respectively. We use Gumbel-softmax~\cite{gumbel,gumbel2} for discrete text tokens.
}
\vspace{-15pt}
\label{fig:arch}
\end{figure*}

\paragraph{Text as the latent representation.} While autoencoders can learn compressed representations $z$ that preserve useful information, it is difficult to use the latent $z$ for downstream tasks without any additional training, let alone direct human interpretation. In this work, we propose to encode the input image into \textit{text}. Practically, the encoder compresses each image into a sequence of BPE-encoded text tokens~\cite{bpe}, where each token can take on a discrete value from the vocabulary. To faithfully reconstruct the image from the latent text, the text must precisely and comprehensively capture the semantic concepts present in the image, making a interface representation, in contrast to image captions that only focus on the most visually salient information.

\paragraph{Text-to-image diffusion as the decoder.} One potential concern is that the encoder might still encrypt the images into a cipher code that only the decoder can decipher, making human interpretation challenging. This is particularly likely when the encoder and the decoder are jointly trained. To mitigate this concern~\cite{lqae}, we introduce a pre-trained text-to-image diffusion model as the decoder, and dub our method as ``De-Diffusion''.

Text-to-image diffusion models, as the name suggested, learn the relationship between text and images from a large dataset of image-text pairs and excel at converting texts into highly detailed images. They already establish the projection from descriptive text to image, and we unpack this encapsulated knowledge by employing a frozen text-to-image diffusion model as the decoder. As illustrated in \cref{fig:arch}, the text-to-image diffusion model consists of a CLIP text encoder~\cite{clip} and a U-Net~\cite{unet}, and the codebook is then naturally the vocabulary of the CLIP text encoder.

When training De-Diffusion, we freeze the parameters of the text-to-image diffusion decoder. In each mini-batch, we expose the decoder with one randomly sampled noise level for each sample. This resembles the training procedure for diffusion models~\cite{ddpm}, except the parameters are fixed and the text conditions are outputs of the image-to-text encoder instead of the training data.

\paragraph{Image-to-text encoder.} The encoder maps the input image into text. It starts with an image backbone that extracts image features, followed by an attentional pooler~\cite{coca,attentional-pooler} that turns the features into output text tokens. The image backbone can be a pre-trained and frozen model that excels at image feature extraction. It can also be randomly initialized, supervised by the reconstruction objective during De-Diffusion training. We ablate the two choices in \cref{tab:image_backbone}.

The attentional pooler projects $n$ learnable queries to $n$ text tokens by a few Transformer blocks~\cite{transformer}. Each Transformer block consists of a self-attention layer over all the queries, a cross-attention layer to gather features from the image backbone, and an MLP layer. After the Transformer blocks, a linear layer projects the queries to discrete text tokens from the vocabulary of CLIP text encoder, in order to connect to the diffusion decoder. The $n$ queries are positional sensitive, meaning that each query corresponds to a specific position in the CLIP text encoder. The $n$ output text tokens, together with the special tokens [$\mathtt{SOS}$] and [$\mathtt{EOS}$], are then fed into the diffusion decoder. We ablate the effect of $n$, the number of text tokens, in \cref{tab:tokens}.

\paragraph{Optimization.} Same as other autoencoders, the training objective of De-Diffusion is to minimize the reconstruction error between the input image and the reconstruction from the pre-trained diffusion model. Specifically, both the loss function and the noise variance schedule strictly follow those of the pre-trained diffusion model~\cite{ddpm}. The training data of De-Diffusion only includes images, without human annotations or paired text descriptions.

Our model can be viewed as a special discrete autoencoder with discrete text tokens as the latent. Similar to other discrete autoencoders~\cite{dall-e,vqvae,vqvae2}, we use Gumbel-softmax~\cite{gumbel,gumbel2} as the continuous relaxation to back-propagate the gradients from the decoder through the discrete latent. The relaxation becomes tight as the temperature $\tau$\,$\rightarrow$\,$0$. We find that an annealing schedule of temperature $\tau$ is important for stable training.

To increase the information density and readability, we exclude all the punctuation in the vocabulary, which accounts for around 6\% of the original vocabulary of CLIP text encoder. As a result, only word tokens and number tokens are allowed. We ablation this design choice in \cref{tab:non-words}.

\subsection{Implementation Details}
\label{sec:training_details}
\paragraph{Text-to-image diffusion model.} The text-to-image diffusion model used for De-Diffusion training is based on Imagen~\cite{imagen}. The U-Net has 600M parameters with an embedding dimension of 256 and input resolution of 64\x64. The text encoder is from OpenCLIP ViT-H/14~\cite{openclip}. The training data is WebLI~\cite{pali}, an image-language dataset built from public web images and texts. We use v-prediction as the objective~\cite{vpred}, a batch size of 2048, and train for 3M steps. For reference, this text-to-diffusion model achieves an FID of 5.37 on 30K 64\x64 MS-COCO 2014 validation images.

\paragraph{Image backbone and attentional pooler.} We utilize a pre-trained CoCa ViT-L model with input resolution 288\x288 as the image backbone, and freeze it during De-Diffusion training~\cite{coca,vit}. This CoCa model is pre-trained on JFT-3B~\cite{jft} and ALIGN datasets~\cite{align}. Our attentional pooler is equipped with 75 queries, in addition to the [$\mathtt{SOS}$] and [$\mathtt{EOS}$] tokens to fully utilize the 77 context length defined by CLIP text encoder~\cite{clip,openclip}. The attention pooler has five Transformer blocks which are always randomly initialized.

\paragraph{Training of De-Diffusion.} The De-Diffusion training data also comes from WebLI~\cite{pali}, while only the images but not the text are used. The broad domain coverage of WebLI enables zero-shot and few-shot evaluations of De-Diffusion on downstream applications in the next section (\cref{sec:applications}). For memory efficiency, we use the Adafactor optimizer~\cite{adafactor} with $\beta_1$\,$=$\,$0.9$, $\beta_2$\,$=$\,$0.999$ and a decoupled weight decay ratio of 0.01. We train with a batch size of 2048 for 500K steps, taking around 2.5 days on 64 TPUv4 chips. The learning rate starts at 3e-4 and is annealed to 3e-6 with cosine decay~\cite{sgdr}, along with a 10K step warmup~\cite{1hour}. The Gumbel-softmax temperature begins from 2.0 and is exponentially annealed to 0.3 through the entire schedule, which we find is sufficient to close the gap between the continuous relaxation during training and the discrete inference.

\section{Experiments and Applications}
\label{sec:applications}
In this section, we introduce several applications of De-Diffusion text, ranging from transferable prompts for text-to-image tools and few-shot vision-language understanding. To demonstrate the versatility of De-Diffusion text across different tasks and domains -- that is, its ability to serve as a strong cross-modal interface -- all the applications use text from \textit{a single} De-Diffusion model detailed in \cref{sec:training_details}.

\subsection{Transferable Text-to-Image Prompt}

\begin{figure}[t!]
\centering
\includegraphics[width=0.95\linewidth]{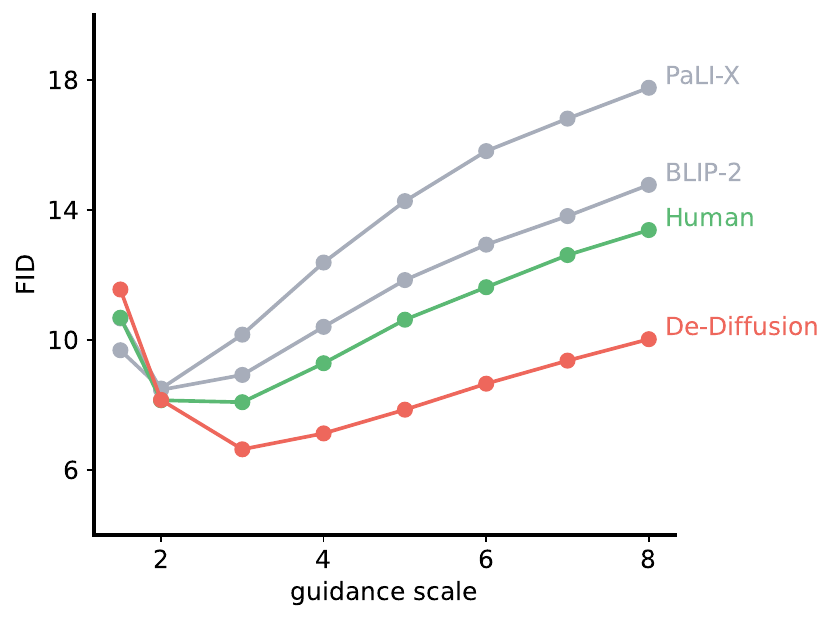}
\vspace{-10pt}
\caption{
\textbf{Evaluating different captioning methods by text-to-image reconstruction.}
The text-to-image model is a pre-trained Stable Diffusion v2-base model~\cite{stable}. We report FID ($\downarrow$) on 30K MS-COCO (2014) validation split with 256\x256 images. De-Diffusion obtains better FID than human-annotated captions, BLIP-2~\cite{blip-2} (fine-tuned on MS-COCO), and PaLI-X~\cite{pali-x} (a multi-task captioning model). Numerical results are provided in \cref{tab:fid-numbers}.
}
\label{fig:fid}
\end{figure}

Since De-Diffusion encodes an input image into text and decode it by a text-to-image diffusion model, it is trivial for De-Diffusion text to serve as a prompt suggestion to reconstruct an image by this specific text-to-image diffusion decoder. Furthermore, we demonstrate that De-Diffusion text is transferable to other unseen decoders, \ie, text-to-image tools, such as Imagen~\cite{imagen}, Stable Diffusion~\cite{stable} and Midjourney~\cite{midjourney}. This suggests that De-Diffusion text is not over-fitted to a single text-to-image decoder but generalizable across different text-to-image frameworks, which is crucial to make a cross-model interface.

\paragraph{Quantitative evaluation.}
We quantitatively evaluate the ability of De-Diffusion text to transfer to other text-to-image diffusion models and compare with traditional captioning methods. To do this, we develop a benchmark that uses a third-party pre-trained text-to-image model to reconstruct an image from either De-Diffusion text or captions. Specifically, we first obtain De-Diffusion text and captions for a given image. Both are then input into the third-party text-to-image model to synthesize the corresponding image. We compare the synthesized image to the original. Text containing more precise and comprehensive descriptions allows the model to produce images more similar to the original. By evaluating the similarity between original and synthesized images, our benchmark quantifies the precision and comprehensiveness of different methods.

We use the pre-trained Stable Diffusion v2-base~\cite{stable} as a generic text-to-image generator, whose weights and training data are oblivious to both De-Diffusion and captioning methods. We measure the similarity between original and synthesized 256\x256 images using FID (Frechet Inception Distance)~\cite{fid} on 30K images from MS-COCO 2014 validation split~\cite{cococap}. Image generation utilizes different classifier-free guidance~\cite{cfg} scales from 1.5 to 8.0, along with 50 steps of DDIM sampling~\cite{ddim}.

We evaluate De-Diffusion, human captions and two state-of-the-art image captioning methods, plotted in \cref{fig:fid}:

(\textit{i}) Human-annotated captions from MS-COCO provide a strong FID baseline of 8.08 at guidance scale 3.0. We synthesize new images using the longest of the five annotated captions, which we find works best. Other options to utilize human captions are discussed in \cref{sec:app-t2i}.

(\textit{ii}) BLIP-2 refers to its ViT-g OPT 2.7B variant~\cite{blip-2}, which is fine-tuned on MS-COCO. As one of the state-of-the-art captioning methods, BLIP-2's FID curve is close to that of human-annotated captions.

(\textit{iii}) PaLI-X~\cite{pali-x} performs fine-tuning on multiple caption datasets, instead of solely on MS-COCO. As a result, its FID curve is higher than that of BLIP-2.

(\textit{iv}) De-Diffusion is trained with solely web images, but not MS-COCO images or any human-annotated captioning data. It has an indirect access to noisy web image-language pairs through the pre-trained diffusion model. However, De-Diffusion achieves the lowest FID of 6.43 at guidance 3.0, significantly better than the human-annotated captions.

These results indicate that De-Diffusion text precisely and comprehensively verbalizes image details, allowing it to effectively transfer to other text-to-image tools.

\paragraph{Qualitative evaluation.} Our visualizations in \cref{fig:viz-coco-1,fig:viz-coco-2} demonstrate that De-Diffusion text is more comprehensive than human-annotated captions. Images are from MS-COCO 2014 validation split and we test with three prominent text-to-image tools including Stable Diffusion XL~\cite{sdxl}, Midjourney~\cite{midjourney}, and Imagen~\cite{imagen}.

The results show that De-Diffusion text covers fine-grained semantic aspects ranging from objects and their positional relationships, human attributes, backgrounds, to action subcategories. In contrast, human-annotated captions often neglect fine-grained semantic details, leading to high variance in the generated images across text-to-image tools. While the descriptions in human captions are precise, De-Diffusion text much more comprehensively enumerates key objects, their attributes, their relationships and background contexts. This comprehension allows cross-tool text-to-image reconstruction with De-Diffusion.

\cref{fig:viz-syn-1,fig:viz-syn-2} visualize text-to-image reconstruction with De-Diffusion text on synthetic images from other text-to-image tools Ideogram\footnote{\url{https://ideogram.ai}} and Lexica\footnote{\url{https://lexica.art}}. We provide the synthetic links of these images in \cref{sec:app-links}. \cref{fig:viz-syn-1} shows De-Diffusion can provide fine-grained descriptions for complex and diverse synthetic images besides photographic images in MS-COCO (\cref{fig:viz-coco-1,fig:viz-coco-2}). The prompts also transfer across different text-to-image models. \cref{fig:viz-syn-2} further highlights the ability of De-Diffusion to articulate diverse image types and explicitly name the genre such as ``\textit{cg wallpaper}'', ``\textit{watercolor painting}'', ``\textit{etching logo}'', and a plain image of a black circle. These results suggest that De-Diffusion can be applied to provide cross-tool prompt inspiration for user-uploaded images to explore new vocabulary and aesthetics.

\subsection{Multi-Modal Few-Shot Learner}

We next show that De-Diffusion can convert an off-the-shelf LLM, which is never trained on vision-language data, to perform open-ended vision-language task by simply prompting the LLM with few-shot examples, and no adaptive training is required.

LLMs exhibit surprising generalization ability with few-shot learning, adapting to new tasks from just a few annotated task-specific examples without any further training~\cite{gpt-3}. However, these powerful models are limited to text. Since then, methods have emerged to enable multi-modal capabilities by encoding images into the word embedding space~\cite{clip,frozen} or training a new module to connect vision and language embeddings~\cite{flamingo,blip-2}. However, these approaches have downsides -- not only would they introduce prohibitively heavy computational costs due to joint training with enormous language models like 540B PaLM~\cite{palm}, but the visual embeddings also bind to a specific language model such that changing the language model requires retraining. This limits the flexibility of these multi-modal models to keep pace with rapid progress in LLMs.

Unlike previous methods based on deep embeddings, De-Diffusion encodes images into text that any language model can readily comprehend. This allows off-the-shelf language models to ground images by simply interleaving task instructions and De-Diffusion text in any order, as \cref{fig:teaser} shows. Using text as a cross-modal interface, De-Diffusion empowers off-the-shelf language models with multi-modal abilities. We next demonstrate that this modular approach achieves state-of-the-art performance on different multi-modal few-shot learning benchmarks, thanks to the comprehensive image context provided by De-Diffusion text, and seamless integration with advanced reasoning abilities provided by the LLMs.

\begin{table}[t!]
\centering
\tablestyle{0.5pt}{1.3}
\resizebox{\linewidth}{!}{
\begin{tabular}{y{80}y{40}x{33}x{15}|x{25}x{28}x{23}}
 &  & \footnotesize{trainable} &  & \footnotesize{VQAv2} & \footnotesize{OKVQA} & \footnotesize{COCO} \\[-3pt]
methods & LLM & \footnotesize{params.} & \footnotesize{shot} & \footnotesize{test}\scriptsize{-}\footnotesize{dev} & val & test \\
\shline
BLIP-2 ViT-g~\cite{blip-2} & FlanT5\textsubscript{XXL} & 108M & 0 & 65.0$^\dagger$ & 45.9$^\dagger$ & - \\
LENS~\cite{lens} & FlanT5\textsubscript{XXL} & 0 & 0 & 62.6 & 43.3 & - \\
AnyMAL ViT-G~\cite{anymal} & Llama2\textsubscript{70B} &- & 0 & 64.2 & 42.6 & 95.9 \\
PICa-Full~\cite{pica} & GPT-3 & 0 & 16 & 56.1 & 48.0 & - \\
\hline
OpenFlamingo-9B~\cite{openflamingo} & MPT\textsubscript{7B} &- & 0  & 52.7 & 37.8 & 79.5 \\
OpenFlamingo-9B~\cite{openflamingo} & MPT\textsubscript{7B} &- & 4  & 54.8 & 40.1 & 89.0 \\
OpenFlamingo-9B~\cite{openflamingo} & MPT\textsubscript{7B} &- & 32 & 53.3 & 42.4 & 99.5 \\
\hline

IDEFICS-80B~\cite{idefics} & Llama\textsubscript{65B} & 14B & 0  & 60.0  & 45.2 & 91.8 \\
IDEFICS-80B~\cite{idefics} & Llama\textsubscript{65B} & 14B & 4  & 63.6  & 52.4 & 110.3 \\
IDEFICS-80B~\cite{idefics} & Llama\textsubscript{65B} & 14B & 32 & 65.9  & 57.8 & \textbf{116.6} \\

\hline
Flamingo-9B~\cite{flamingo}& Chinchilla\textsubscript{7B} & 2B & 0  & 51.8  & 44.7 & 79.4 \\
Flamingo-9B~\cite{flamingo}& Chinchilla\textsubscript{7B} & 2B & 4  & 56.3  & 49.3 & 93.1 \\
Flamingo-9B~\cite{flamingo}& Chinchilla\textsubscript{7B} & 2B & 32 & 60.4  & 51.0 & 106.3 \\
\hline
Flamingo-80B~\cite{flamingo} & Chinchilla\textsubscript{70B} & 10B & 0  & 56.3  & 50.6 & 84.3 \\
Flamingo-80B~\cite{flamingo} & Chinchilla\textsubscript{70B} & 10B & 4  & 63.1  & 57.4 & 103.2\\
Flamingo-80B~\cite{flamingo} & Chinchilla\textsubscript{70B} & 10B & 32 & 67.6  & 57.8 & \underline{113.8} \\
\shline
De-Diffusion ViT-L & PaLM 2-S & 135M & 0 & 63.9 & 51.4 & 63.4  \\
De-Diffusion ViT-L & PaLM 2-S & 135M & 4 & 64.0 & 53.5 & 87.1 \\
De-Diffusion ViT-L & PaLM 2-S & 135M & 32 &63.1 & 53.3 & 92.0 \\
\hline
De-Diffusion ViT-L & PaLM 2-L & 135M & 0 & 67.2   & 57.0 & 88.5 \\
De-Diffusion ViT-L & PaLM 2-L & 135M & 4 & \underline{67.9}   & \underline{58.2} & 100.3 \\
De-Diffusion ViT-L & PaLM 2-L & 135M & 32 & \textbf{68.4} & \textbf{60.6} & 103.7 \\
\end{tabular}}
\caption{
\textbf{Vision-language few-shot learning.}
We report VQA accuracy~\cite{antol2015vqa} for visual question answering on VQAv2~\cite{vqav2} and OKVQA~\cite{okvqa} in the open-ended setting, and CIDEr ~\cite{vedantam2015cider} for MS-COCO image captioning~\cite{cococap}. The \textbf{Bold} denotes the top performance and the \underline{underlined} denotes the second-best in each column. $^\dagger$ in-domain COCO images are used for training.}
\label{tab:few-shot-vl}
\end{table}

\begin{table*}[t]
\tablestyle{0pt}{0.0}
\begin{tabular}{p{0.2\linewidth}p{0.8\linewidth}}
\vspace{0pt}
\begin{tabular}{c}
\includegraphics[width=0.9\linewidth]{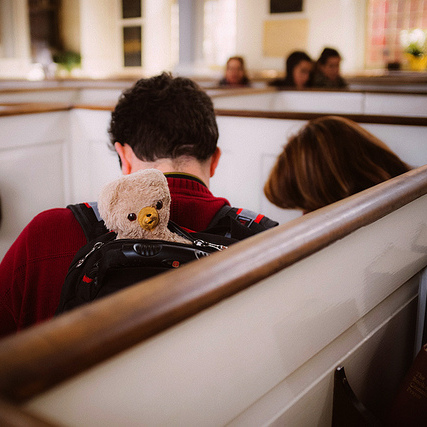} \\[3.6pt]
(a) reference image
\end{tabular}
& 
\vspace{0pt}
\begin{minipage}{\linewidth}{
\footnotesize \fontfamily{cmss}\selectfont
[\textbf{LLM prompt}] Answer the question given the context. \\
Image context: a colvonvscocam blog closeup of young rear man looking carrying head teddybear wearing a red sweater it in white barriers amidst between a a wooden polcoping a rails opposite blurry except a woman people sitting right off white cabinets lit white windows and sill approximately wearing hair hair burgundy and black tabletop backpack approximately wallets blush brown hair hair tallinn salzburg church church church backpack closeup closeup hair hair cubic \\
Image question: What toy is this? Short answer: \\[3pt]
{[\textbf{LLM completion}]} teddy bear. \\[3pt]
{[\textbf{GT answers}] stuffed animal, teddy bear}\\ 
}\end{minipage}
\end{tabular}

\tablestyle{0pt}{0.0}
\begin{tabular}{p{0.2\linewidth}p{0.8\linewidth}}
\vspace{0pt} 
\begin{tabular}{c}
\includegraphics[width=0.9\linewidth]{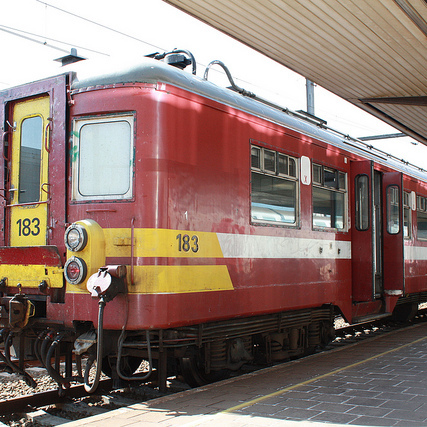} \\[5pt]
(b) reference image
\end{tabular}
& 
\vspace{0pt}
\begin{minipage}{\linewidth}{
\footnotesize \fontfamily{cmss}\selectfont
[\textbf{LLM prompt}] Answer the question given the context. \\
Image context: a colcandidenverlanticcloseup former recent train train parked traditionenclosed metrotram in a red livery it on railroad platform containing wearing a a yellowpolsurround a knob beside platform near a under platform shelter right there and roof shadows and platform and tracks etc wore worn worn maumaroon brown white stripes markings contentworn yellow yellow stripes train pretoria namibia railway platform train operator worn brown windows platform platform \\
Image question: What other big vehicle is often painted about the same shade as this vehicle? Short answer: \\[3pt]
{[\textbf{LLM completion}]} fire truck. \\[3pt]
{[\textbf{GT answers}] firetruck, fire truck}\\ 
}\end{minipage}
\end{tabular}
\vspace{-15pt}
\captionof{figure}{
\textbf{VQA with an off-the-shelf LLM},
where De-Diffusion text of the reference image is inserted after ``{\fontfamily{cmss}\selectfont Image context}'' in the LLM prompt. The LLM then completes the prompt to answer the visual question. De-Diffusion text provides abundant visual details, \eg, \textit{teddy bear} in (a) and \textit{red livery} of the \textit{train} in (b). We use PaLM 2-L~\cite{palm2} as the LLM. Samples are from OKVQA~\cite{okvqa}.
}
\label{tab:okvqa-examples}
\vspace{-10pt}
\end{table*}

\paragraph{Multi-modal few-shot learning.}
We follow the evaluation protocol of Flamingo~\cite{flamingo} to assess few-shot learning on three vision-language tasks including VQAv2~\cite{vqav2}, OKVQA~\cite{okvqa} and MS-COCO caption~\cite{cococap}. De-Diffusion text for the support images is interleaved along with their questions, answers, and captions to form prompts for the LLMs. The LLM's completion is considered a correct answer only if it exactly matches the ground truth. More details are in \cref{sec:app-mm}. Results are shown in \cref{tab:few-shot-vl}.

Thanks to the modular nature of De-Diffusion text, we are able to couple the same set of De-Diffusion text with different language models, PaLM 2-S and PaLM 2-L~\cite{palm2} without multi-modal training.
The performance of De-Diffusion text paired with PaLM 2-L increases from zero-shot to 32-shot setup on all three tasks. However, when coupled with PaLM 2-S, the 32-shot performance slightly decreases on two VQA benchmarks compared to using four shots. We hypothesize this is because smaller language models like PaLM 2-S benefit less from long context~\cite{wei2023larger}, \eg, the around 3600-token prompts for 32 shots.

De-Diffusion text paired with PaLM 2-L matches other methods on MS-COCO captioning, and establishes new state-of-the-art results on two VQA benchmarks for all zero-shot, 4-shot, and 32-shot settings. Meanwhile, De-Diffusion training is also more lightweight in both data and computation. Data-wise, De-Diffusion only requires images, unlike Flamingo and its followups~\cite{flamingo,openflamingo,idefics} which use massive interleaved web text and images, or BLIP-2~\cite{blip-2} which needs human annotations. Computation-wise, De-Diffusion not only uses far fewer parameters (135M in De-Diffusion \vs 10B in Flamingo-80B), but its training also does not involve inference with frozen LLMs like 70B-parameter Chinchilla~\cite{chinchilla} in Flamingo. Instead, it only requires frozen 600M U-Net and CLIP text encoder (\cref{sec:training_details}).

Our results suggest that LLMs, without any multi-modal training, can make grounded inferences for vision-language tasks using just text descriptions of images. The benefits of language models are more pronounced on challenging situations requiring reasoning and commonsense knowledge, such as d Outside Knowledge VQA (OKVQA)~\cite{okvqa}. As the examples in \cref{tab:okvqa-examples} show, LLMs can answer non-trivial visual questions that demand both De-Diffusion image context and commonsense knowledge.

\begin{table}[t!]
\centering
\tablestyle{1.2pt}{1.3}
\resizebox{1.0\linewidth}{!}{
\begin{tabular}{y{103}|x{23}x{23}x{27}|x{23}x{23}x{27}}
 &  \multicolumn{3}{c|}{VQAv2} & \multicolumn{3}{c}{OKVQA} \\[-3pt]
methods  & 0-shot & 4-shot & 32-shot & 0-shot & 4-shot & 32-shot \\
\shline
\small{BLIP-2 OPT\textsubscript{2.7b}} caption~\cite{blip-2}  & 63.1 & 63.0 & 62.8 & 58.5 & 57.6 & 59.1 \\
Human caption~\cite{cococap} & 63.1 & 63.2 & 63.6 & \textbf{59.0} & \textbf{58.9} & 60.1 \\
De-Diffusion ViT-L & \textbf{65.2} & \textbf{66.0} & \textbf{66.2} & 57.0 & 58.2 & \textbf{60.6} \\
\end{tabular}}
\vspace{-10pt}
\caption{
\textbf{Compare to other captions}
on the val split of VQAv2 and OKVQA. BLIP-2 represents the top captioning model. Human captions are from MS-COCO annotations. PaLM 2-L is used.}
\label{tab:few-shot-vl-cap}
\vspace{-13pt}
\end{table}

On the other hand, the results suggest that De-Diffusion text comprehensively captures semantic elements in images, which language models can leverage to perform VQA and captioning. In \cref{tab:few-shot-vl-cap}, De-Diffusion text outperforms the state-of-the-art captioning method BLIP-2~\cite{blip-2} and even human-annotated captions on VQAv2~\cite{vqav2}. It also surpasses both on OKVQA~\cite{okvqa} given 32 shots. This suggests the strength of De-Diffusion text in providing rich context.

\begin{table*}[t!]
\tablestyle{0pt}{0.0}
\begin{tabular}{p{0.2\linewidth}p{0.8\linewidth}}
\vspace{0pt}
\begin{tabular}{c}
\includegraphics[width=0.73\linewidth]{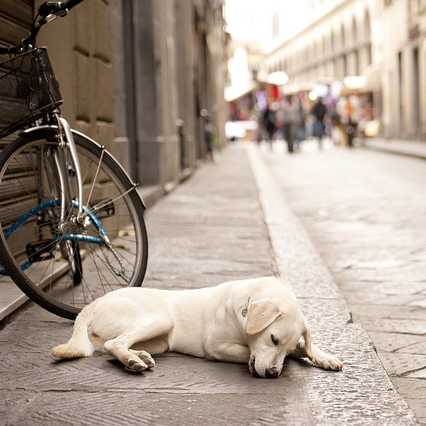} \\[3.6pt]
reference image
\end{tabular}
& 
\vspace{0pt}
\begin{minipage}{\linewidth}{
\footnotesize \fontfamily{cmss}\selectfont
[\textbf{Chatbot prompt}] Image context: <De-Diffusion text>. What is amusing about this image? \\[5pt]
{[\textbf{ChatGPT-3.5}]} The amusing aspect of the image seems to be the juxtaposition and contrast between the serene and peaceful scene of the Labrador retriever peacefully resting on the slate footpath, and the busy, crowded, and somewhat chaotic background of a bustling street in Milan. \\[5pt] 
{[\textbf{Bard}]} The amusing thing about the image is the juxtaposition of the sleeping dog and the busy street. The dog is completely relaxed and oblivious to the hustle and bustle around it. The contrast between the two creates a sense of humor. \\ 
}\end{minipage}
\end{tabular}
\vspace{-15pt}
\captionof{figure}{
\textbf{Multi-modal dialogue with off-the-shelf text-only chatbots},
where De-Diffusion text is inserted after ``{\fontfamily{cmss}\selectfont Image context}'' in the text prompt for ChatGPT-3.5 and Bard. Full {\fontfamily{cmss}\selectfont <De-Diffusion text>} of this reference image is in \cref{fig:viz-coco-1}.}
\label{tab:chatbot}
\vspace{-15pt}
\end{table*}

\begin{table}[t!]
\centering
\captionsetup[subfloat]{size=small}
\subfloat[\textbf{2-way Classification}\label{tab:2-way-cls}]{
\tablestyle{2pt}{1.3}
\begin{tabular}{y{60}y{57}|x{51}x{46}}
methods              &  LLM      & w/o induction & w/ induction \\
\shline
Frozen~\cite{frozen} & Frozen~\cite{frozen} & 1.7   & 65.0 \\
LQAE~\cite{lqae}     & GPT3.5~\cite{gpt-3}  & 1.5   & 68.7 \\
SPAE\textsubscript{PaLM}~\cite{spae} & PaLM 2-L~\cite{palm2}  & 32.2 & 85.4 \\
\hline
De-Diffusion & Llama2\textsubscript{70B}~\cite{lama}  & 60.8 & 95.0 \\
De-Diffusion & PaLM 2-S~\cite{palm2} & \textbf{79.2} & 98.1 \\
De-Diffusion & PaLM 2-L~\cite{palm2} & 78.9 & \textbf{99.3} \\
\end{tabular}
}

\subfloat[\textbf{5-way Classification}\label{tab:5-way-cls}]{
\tablestyle{2pt}{1.3}
\begin{tabular}{y{60}y{57}|x{50}x{46}}
methods              &  LLM      & w/o induction & w/ induction \\
\shline
\demph{P$>$M$>$F}~\cite{pmf} & \demph{-} & \multicolumn{2}{c}{\demph{95.3}} \\
\hline
Frozen~\cite{frozen} & Frozen~\cite{frozen} & 0.9   & 33.8 \\
LQAE~\cite{lqae}     & GPT3.5~\cite{gpt-3}  & 1.0   & 45.9 \\
SPAE\textsubscript{PaLM}~\cite{spae} & PaLM 2-L~\cite{palm2} & 23.6 & 67.0 \\
\hline
De-Diffusion & Llama2\textsubscript{70B}~\cite{lama}  & 64.8 & 87.9 \\
De-Diffusion & PaLM 2-S~\cite{palm2}  & 66.4 & 88.6 \\
De-Diffusion & PaLM 2-L~\cite{palm2}  & \textbf{71.8} & \textbf{97.0} \\
\end{tabular}
}
\vspace{-10pt}
\caption{
\textbf{Open-ended one-shot classification on miniImageNet},
where only the exact class names predicted by the LLM are considered correct. Task induction is introductory text explaining the classification task and providing expected class names at the start of the prompt. Previous best in the closed form is \demph{de-emphasized}.}
\label{tab:few-shot-cls}
\vspace{-15pt}
\end{table}

\paragraph{Open-ended one-shot classification.} 
We follow the protocol from Frozen~\cite{frozen} to evaluate open-ended one-shot image classification on miniImageNet~\cite{miniimagenet}. We interleave De-Diffusion text for the support images along with their real class names as prompts for the LLM. The text generated by the LLM is used as the prediction.

We evaluate in an open-ended fashion, where only generating the exact class name is considered correct. There is also an option of task induction, which is introductory text explaining the classification task and providing expected class names at the beginning of the prompt, \eg, ``\texttt{Classify the context into dog or cat.}'' More details are in \cref{sec:app-mm-cls}.

The results are shown in \cref{tab:few-shot-cls}. Task induction largely increases performance because it helps the language model to generate the exact class names required for open-ended evaluation. With three different LLMs, LLaMA-70B~\cite{lama}, PaLM 2-S and PaLM 2-L~\cite{palm2}, De-Diffusion significantly outperforms previous methods on both 2-way and 5-way classification. PaLM 2-L inference with task induction achieves 97.0\% accuracy, even surpassing the previous closed-form state-of-the-art of 95.3\% systematically. These results suggest De-Diffusion excels at verbalizing class names of main objects in images.

\subsection{Multi-modal Dialogue}
Chatbots such as ChatGPT-3.5~\cite{gpt4} and Bard~\cite{bard} are LLM-based models that engage users with conversational interactions. They have demonstrated impressive advances in natural language understanding, generation, and conversational capabilities. These chatbots can engage in remarkably human-like dialogue, answer follow-up questions, and perform helpful tasks. However, as language models, they lack grounding in the visual world. In \cref{tab:chatbot}, we demonstrate that De-Diffusion text can provide this missing visual grounding. By incorporating De-Diffusion descriptions of images into the conversational context, chatbots can leverage the rich visual details captured in the text. This allows them to answer challenging questions that require complex reasoning and commonsense knowledge. Furthermore, we find De-Diffusion text transfers across different chatbots.
We explore more combinations in \cref{sec:app-text-mix}.

\begin{table*}[t]
\centering
\captionsetup[subfloat]{labelfont=footnotesize,textfont=footnotesize}
\subfloat[
\textbf{Number of tokens}.
\label{tab:tokens}
]{
\centering
\begin{minipage}{0.16\linewidth}{\begin{center}
\tablestyle{3pt}{1.3}
\begin{tabular}{c|cc}
tokens  & FID$\downarrow$ & acc. \\
\shline
5  & 9.19 & \textbf{97.8} \\
15 & 7.42 & 97.6\\
45 & 6.95 & 97.0 \\
75 & \baseline{\textbf{6.43}} & \baseline{97.0} \\
\end{tabular}
\end{center}}\end{minipage}
}
\hspace{2em}
\subfloat[
\textbf{Excluding punctuation}.
\label{tab:non-words}
]{
\begin{minipage}{0.20\linewidth}{\begin{center}
\tablestyle{3pt}{1.3}
\begin{tabular}{c|cc}
punctuation  & FID$\downarrow$ & acc. \\
\shline
$\checkmark$ & 6.85 & 96.8  \\
$\times$ & \baseline{\textbf{6.43}} & \baseline{\textbf{97.0}} \\
\multicolumn{3}{c}{~}\\
\multicolumn{3}{c}{~}\\
\end{tabular}
\end{center}}\end{minipage}
}
\hspace{2em}
\subfloat[
\textbf{Pooler depth}.
\label{tab:decoder_depth}
]{
\begin{minipage}{0.16\linewidth}{\begin{center}
\tablestyle{3pt}{1.3}
\begin{tabular}{c|cc}
blocks  & FID$\downarrow$ & acc. \\
\shline
3 & 6.85 & 96.6 \\
5 & \baseline{\textbf{6.43}} & \baseline{\textbf{97.0}} \\
9 & 6.76 & 93.1\\
\multicolumn{3}{c}{~}\\
\end{tabular}
\end{center}}\end{minipage}
}
\hspace{2em}
\subfloat[
\textbf{Image backbone}.
\label{tab:image_backbone}
]{
\begin{minipage}{0.30\linewidth}{\begin{center}
\tablestyle{3pt}{1.3}
\begin{tabular}{ccc|cc}
arch. & init. & \# steps & FID$\downarrow$ & acc. \\
\shline
ViT-Base  & CoCa & 300K & 6.84 & 92.6\\
ViT-Large & CoCa & 300K & \baseline{\textbf{6.43}} & \baseline{\textbf{97.0}} \\
ViT-Large & rand & 300K & 14.6 & 67.2 \\
ViT-Large & rand & 500K & 11.0 & 72.2\\
\end{tabular}
\end{center}}\end{minipage}
}
\vspace{-5pt}
\caption{
\textbf{De-Diffusion ablation experiments.}
We evaluate text-to-image reconstruction FID ($\downarrow$) on MS-COCO (2014) validation split using 256\x256 images with Stable Diffusion v2-base. We report the best FID across guidance scales. We also report open-ended 5-way 1-shot classification accuracy on miniImageNet. Default settings are marked in \colorbox{defaultcolor}{gray}.}
\label{tab:ablation}
\vspace{-12pt}
\end{table*}

\subsection{Ablation}
In this section, we ablate different design choices of De-Diffusion. By default, the encoder is a frozen CoCa pre-trained ViT-Large model, and we train De-Diffusion for 300K steps. For text-to-image reconstruction, we use FID on Stable Diffusion v2.0-base, the same setting as \cref{fig:fid}, reporting the lowest FID across guidance scales. For few-shot learning, we use 5-way 1-shot classification accuracy on miniImageNet with task induction, identical to \cref{tab:5-way-cls}.

\paragraph{Number of tokens.}
De-Diffusion text by default uses up all 75 tokens from the CLIP text encoder context. In \cref{tab:tokens}, we show performance using 5, 15, and 45 tokens. With more tokens, reconstruction with Stable Diffusion improves, with FID decreasing from 9.19 to 6.43. This aligns with our intuition that longer text descriptions as prompts lead to better text-to-image reconstruction. Interestingly, few-shot classification accuracy decreases from 97.8\% to 97.0\% with longer text. This suggests when context length is limited, De-Diffusion prioritizes the most salient semantic concepts, usually the image classes. This aligns with the training objective of De-Diffusion to find the most representative text latent to minimize reconstruction error of autoencoding. With longer context, De-Diffusion text includes more comprehensive but subtle concepts beyond the classes, important for reconstruction but not classification.

\paragraph{Excluding punctuation.}
We use the 49K token vocabulary of CLIP as the codebook of latent representations. This naturally results from using the CLIP text encoder for the text-to-image diffusion model. However, we exclude punctuation from the vocabulary, which accounts for around 6\% of the original tokens. By excluding these, we can devote more of the limited 75 latent tokens to content words, allowing more semantic concepts to be expressed. In \cref{tab:non-words}, we vary these choices. Excluding punctuation improves reconstruction FID on Stable Diffusion from 6.85 to 6.43, suggesting better transferability of De-Diffusion text to other text-to-image models, likely due to the use of more content words. On the other hand, few-shot accuracy on miniImageNet only drops 0.2\%, showing punctuation has a small influence on few-shot learning ability when using LLMs.

\paragraph{Pooler depth.}
\cref{tab:decoder_depth} varies the depth, \ie, number of Transformer blocks, in the attentional pooler of the image-to-text encoder. Too few layers may limit its ability to capture all the necessary semantics. But too many layers could overfit to the specific text-to-image diffusion model and hurt generalizability. Experiments suggest that with as few as three Transformer blocks, the attentional pooler can effectively transform image features from the pre-trained CoCa backbone into De-Diffusion text. With five Transformer blocks, we obtain the best performance on both reconstruction FID with Stable Diffusion and few-shot accuracy on miniImageNet. This implies that the pre-trained CoCa backbone provides effective image features for image to text encoding, which we examine next.

\paragraph{Image backbone.}
\cref{tab:image_backbone} varies different image backbone architectures. Increasing the frozen pre-trained CoCa backbone size from ViT-Base to ViT-Large largely improves performance, reducing reconstruction FID from 6.84 to 6.43, and improving few-shot classification accuracy from 92.6\% to 97.0\%. We also explore a randomly initialized backbone optimized by the De-Diffusion objective. With 300K training steps, this obtains an FID of 14.6 and few-shot accuracy of 67.2\%. Performance increases with a longer 500K schedule, as expected for generative model training. Though still behind pre-trained CoCa backbones, training from scratch achieves 72.2\% few-shot accuracy on miniImageNet, surpassing prior methods like SPAE with PaLM 2-L at 67.0\% despite. This highlights the promise of using a pre-trained generative model as supervision to train an image backbone from scratch. By learning to synthesize inputs, such a backbone is encouraged to captures all factors of variation, aligning with the principle of analysis by synthesis.

\begin{table}[!t]
\centering
\tablestyle{6pt}{1.3}
\begin{tabular}{c|cc}
training data & MS-COCO FID$\downarrow$ & miniImageNet acc. \\
\shline
WebLI~\cite{pali} & \baseline{\textbf{6.43}} & \baseline{97.0} \\
ImageNet-1K~\cite{imagenet} & 6.93 & \textbf{97.2} \\
MS-COCO~\cite{coco} & 7.53 & 85.8 \\
\end{tabular}
\vspace{-10pt}
\caption{
\textbf{De-Diffusion ablation on training data.}
Settings are the same as those in \cref{tab:ablation}. Default setting is in \colorbox{defaultcolor}{gray}.}
\label{tab:training_data}
\vspace{-10pt}
\end{table}

\paragraph{Training data.}
We explore using different training images for De-Diffusion, such as web images in WebLI~\cite{pali} and the training splits of ImageNet-1K~\cite{imagenet} and MS-COCO (2014). Results are shown in \cref{tab:training_data}. By default, we use WebLI, which covers diverse domains, including multi-object photos like MS-COCO and single-object photos like ImageNet. Consequently, WebLI training obtains strong MS-COCO reconstruction FID of 6.43 and 97.0\% few-shot classification accuracy on miniImageNet. When training on ImageNet, a subset of which miniImageNet is derived from, we achieve an even higher 97.2\% few-shot accuracy. This suggests in-domain training images benefit De-Diffusion. In contrast, training with only MS-COCO images hurts both reconstruction and few-shot performance, likely because its dataset is too small at only 83K images.

\begin{figure*}[t]
\centering
\tablestyle{0.0pt}{0.5}
\begin{tabular}{q{0.25\linewidth}k{0.25\linewidth}q{0.25\linewidth}v{0.25\linewidth}}
Original Image & 
\multicolumn{1}{c}{Stable Diffusion XL} &
Midjourney &
\multicolumn{1}{c}{Imagen} \\
\\[-3pt]
\multirow{2}{*}[4em]{\includegraphics[width=0.9\linewidth]{figs_arxiv/generation_compare_coco/74.jpg}} &
\includegraphics[width=0.96\linewidth]{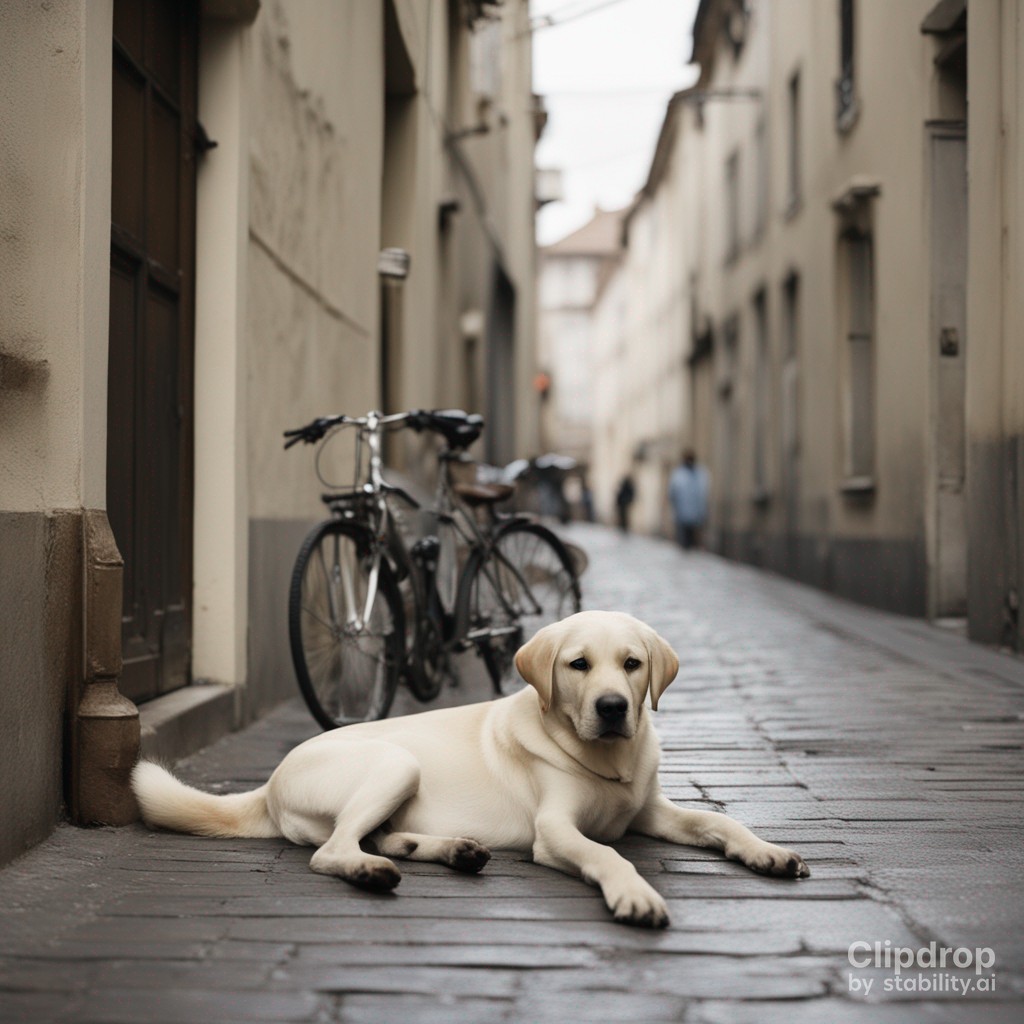} &
\includegraphics[width=0.96\linewidth]{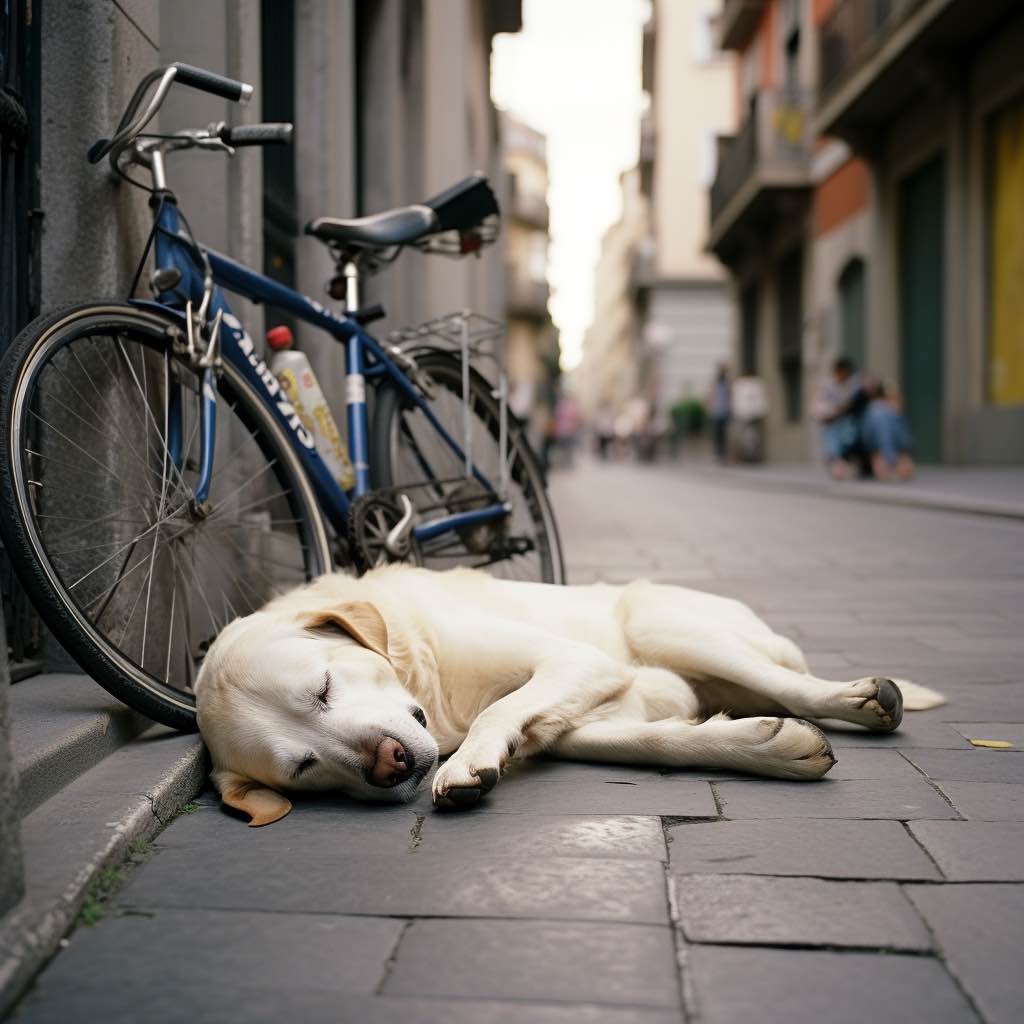} &
\includegraphics[width=0.96\linewidth]{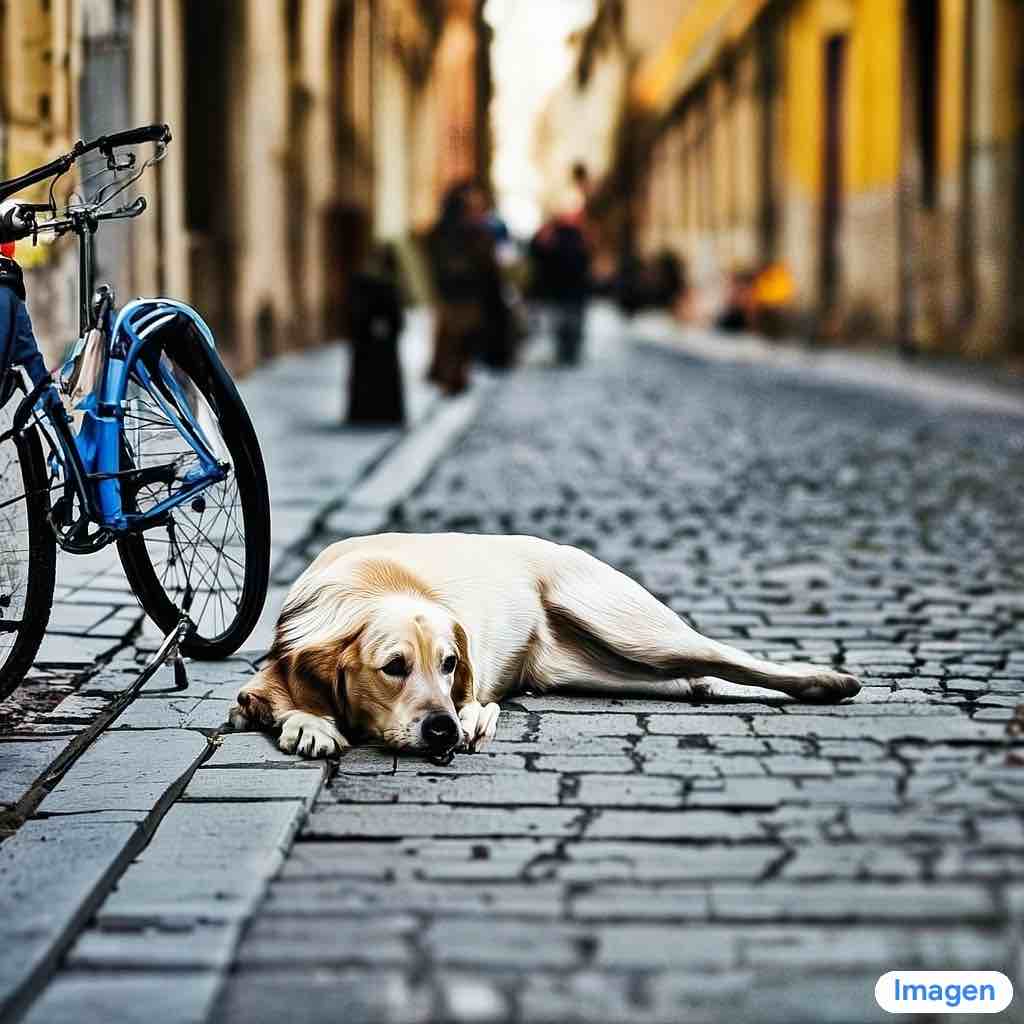} \\
& \multicolumn{3}{p{0.75\textwidth}}{\footnotesize \fontfamily{cmss}\selectfont [De-Diffusion Text] an landsapiccinemageneric photograph dog labrador aus white creamy labrador retriever lying lying resting threshold lying an onto slate footpath pathway street milan ositalian retristreet stil relating called an cream dog shown \textcolor{citecolor}{\textbf{sleeping sleeping beside near an blue left bicycle bicycle left}} crowded street left tyre and umbrella blurry beige brown monochrome left left towards and sitting among people street gray walls alley mostly brown buildings street blur street pathway street} \\
\\
\multirow{2}{*}[8em]{\textcolor{citecolor}{\textbf{object positional relationships}}} &
\includegraphics[width=0.96\linewidth]{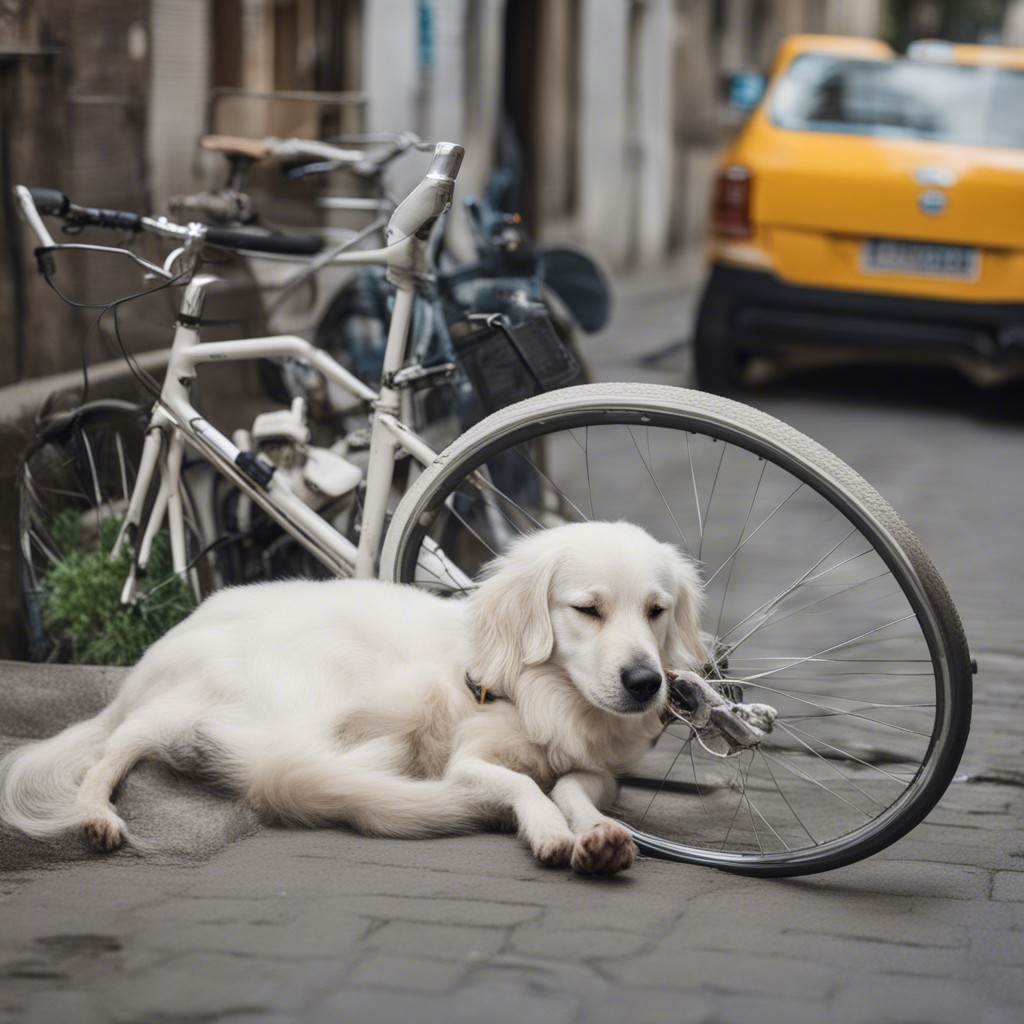} & 
\includegraphics[width=0.96\linewidth]{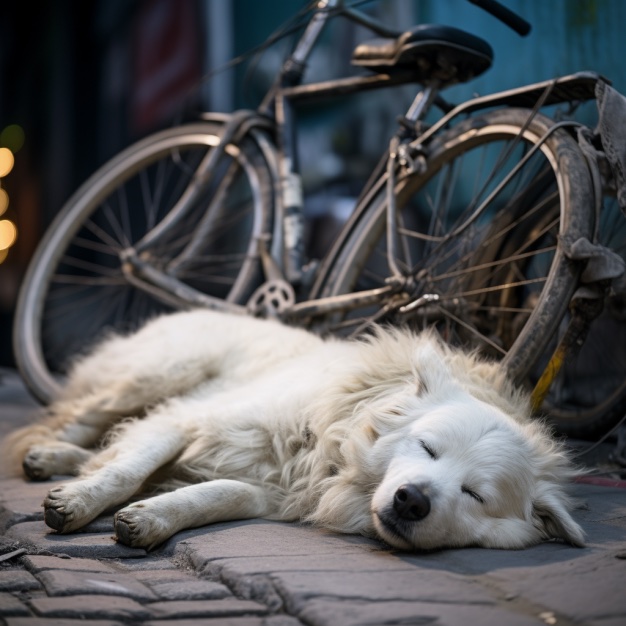} &
\includegraphics[width=0.96\linewidth]{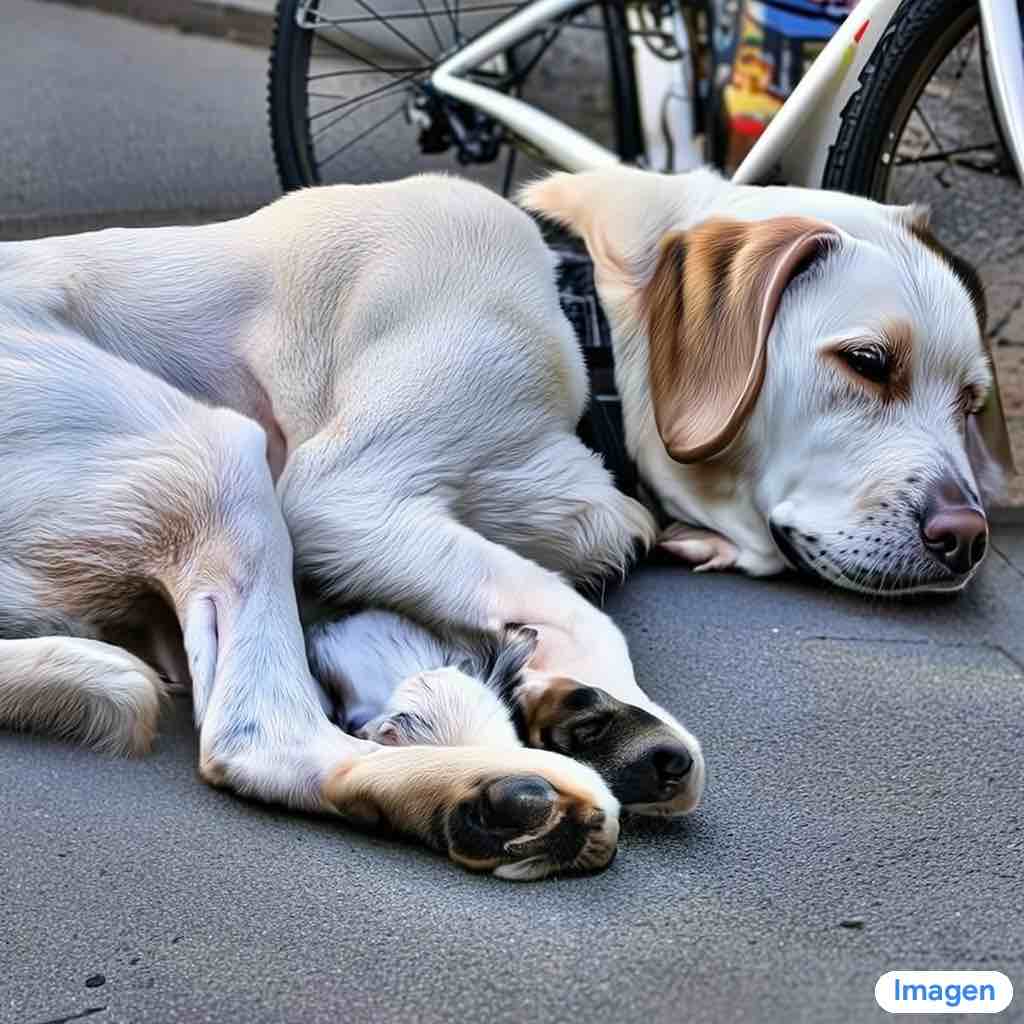} \\
& \multicolumn{3}{p{0.75\textwidth}}{\footnotesize \fontfamily{cmss}\selectfont [GT Caption] A white dog is sleeping on a street and a bicycle.} \\

\shline
\\[-3pt]

\multirow{2}{*}[4em]{\includegraphics[width=0.9\linewidth]{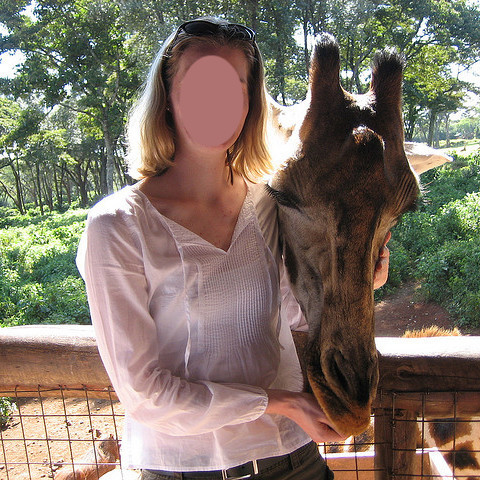}} &
\includegraphics[width=0.96\linewidth]{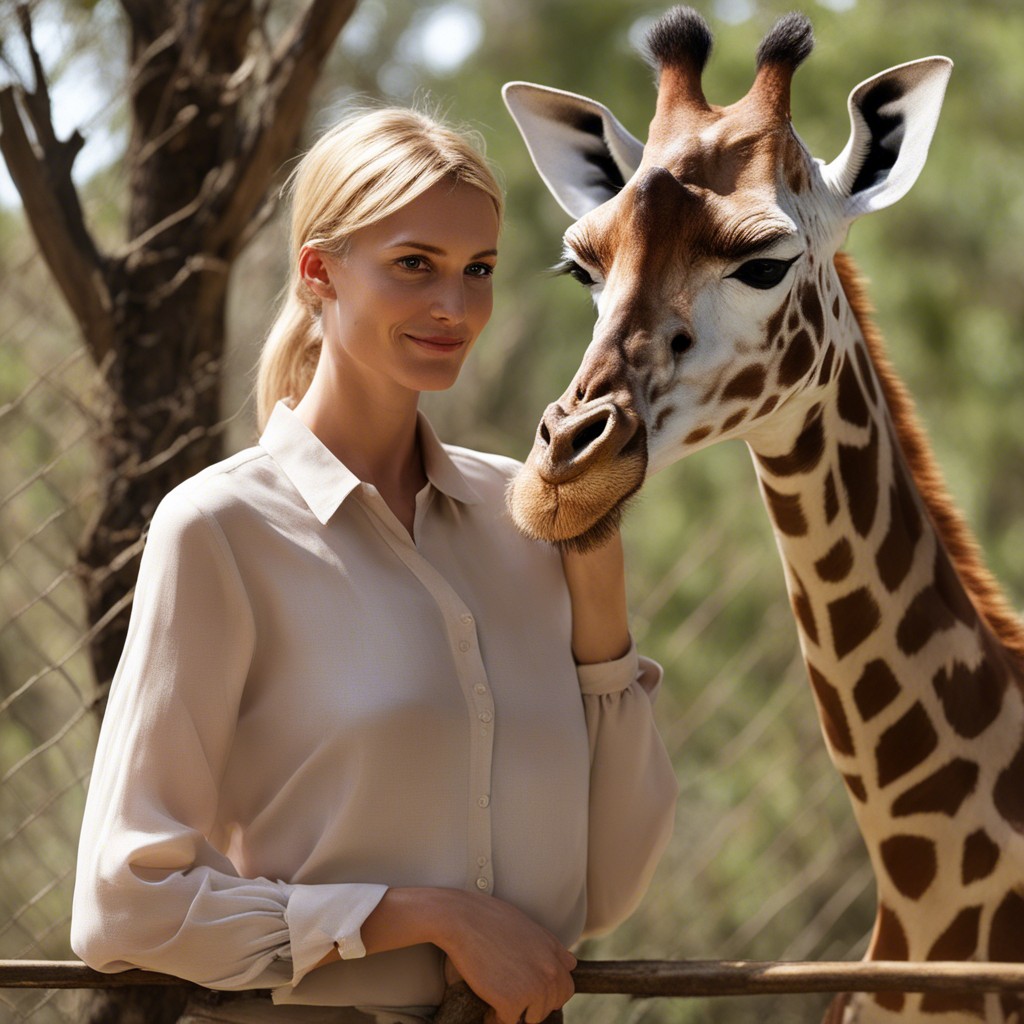} &
\includegraphics[width=0.96\linewidth]{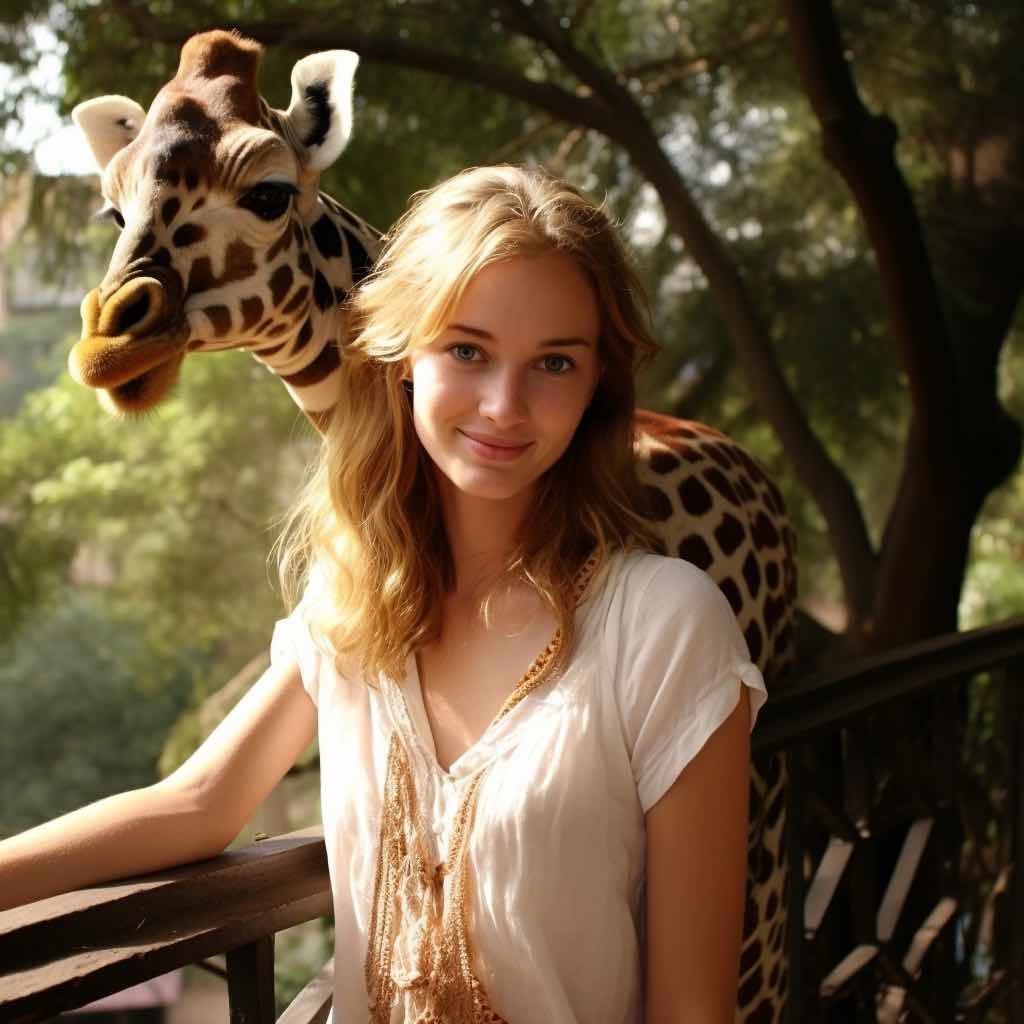} &
\includegraphics[width=0.96\linewidth]{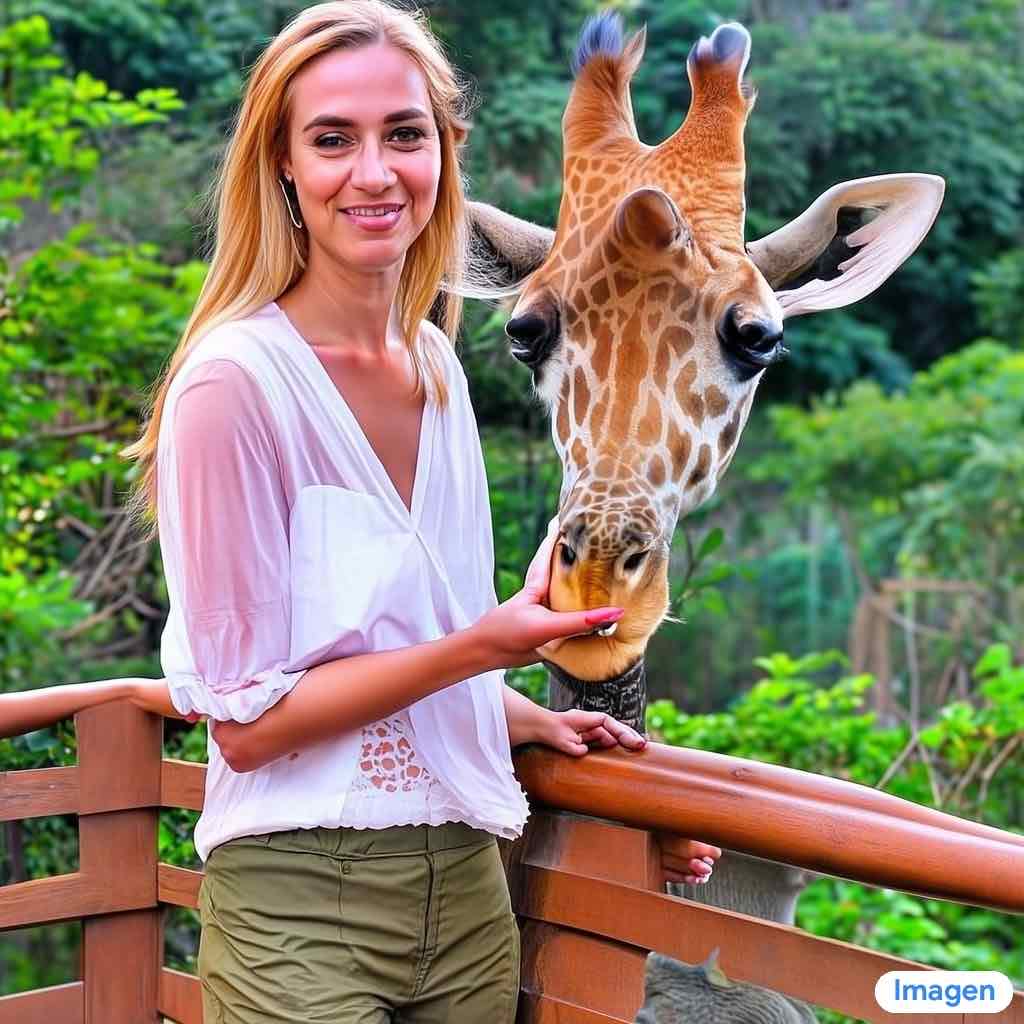} \\
& \multicolumn{3}{p{0.75\textwidth}}{\footnotesize \fontfamily{cmss}\selectfont [De-Diffusion Text] an attribumontagjagsinfo closeup woman giraffe \textcolor{citecolor}{\textbf{wearing white sheer sheer blouse}} long eved olive pants standing an on terracotta fencing balcony tanzania tanzania osdaria jens keynes presented yet description \textcolor{citecolor}{\textbf{an blond female}} shown holding lovingly embraced holds an shadows animal giraffe head when smile smile animal ear blonde neck abadbrown brown purple consist though among wooden plants among plants animals shady trees trees either trees rainforest shadows blouse holistic zoo} \\
\\
\multirow{2}{*}[8em]{\textcolor{citecolor}{\textbf{human attributes}}} &
\includegraphics[width=0.96\linewidth]{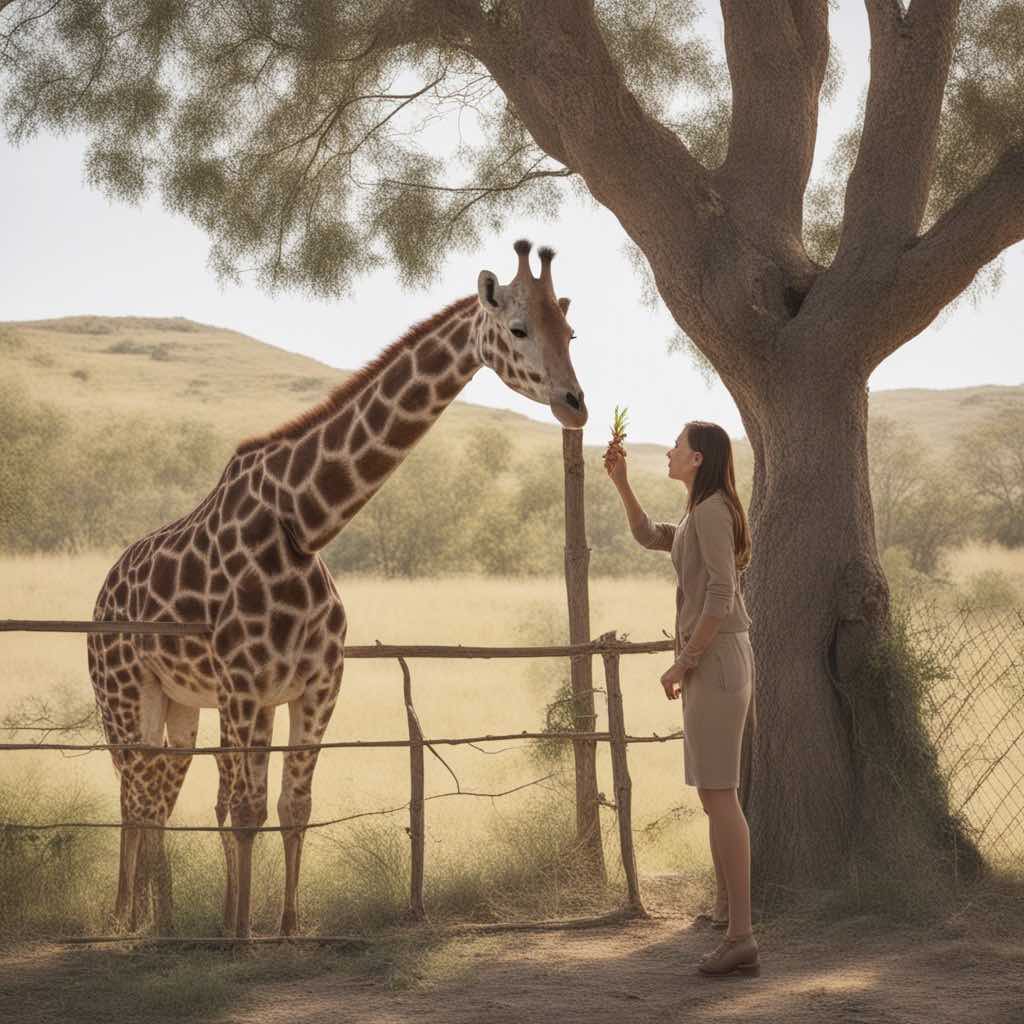} & 
\includegraphics[width=0.96\linewidth]{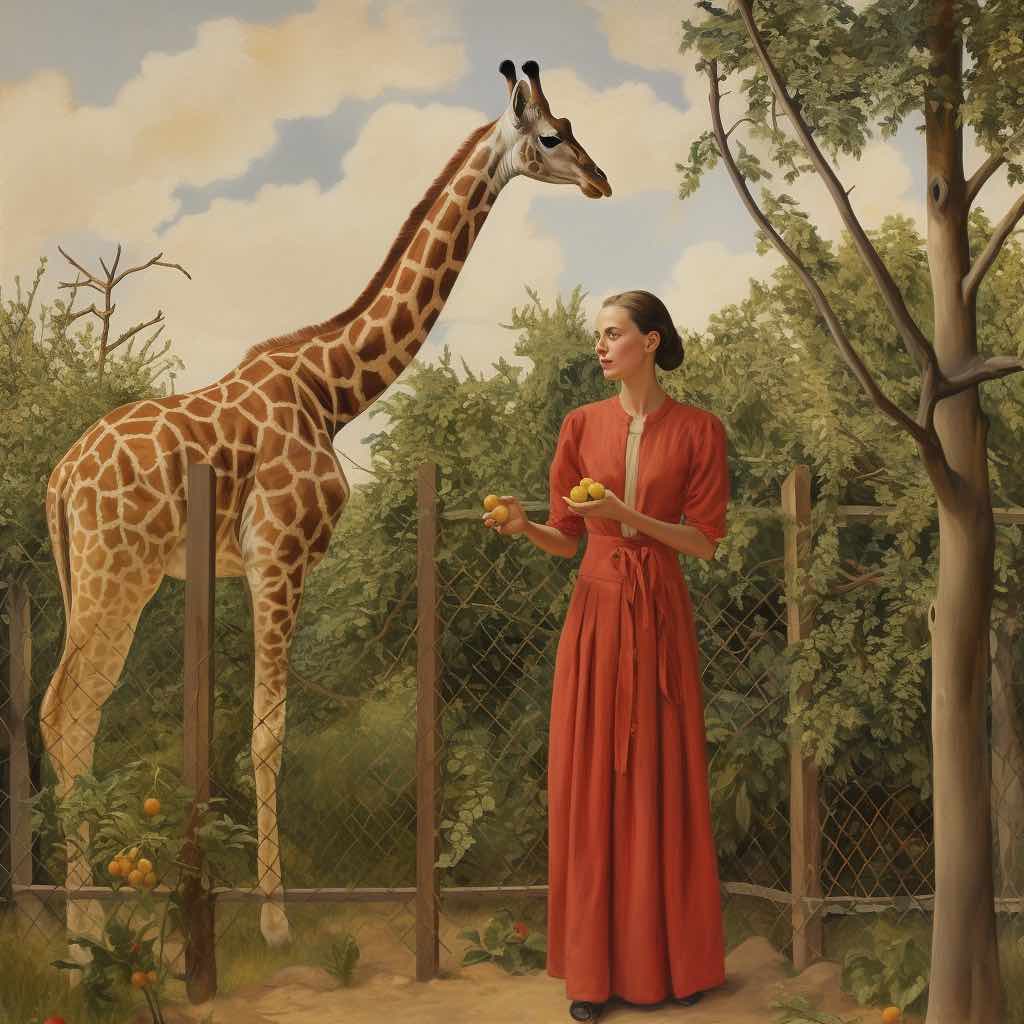} &
\includegraphics[width=0.96\linewidth]{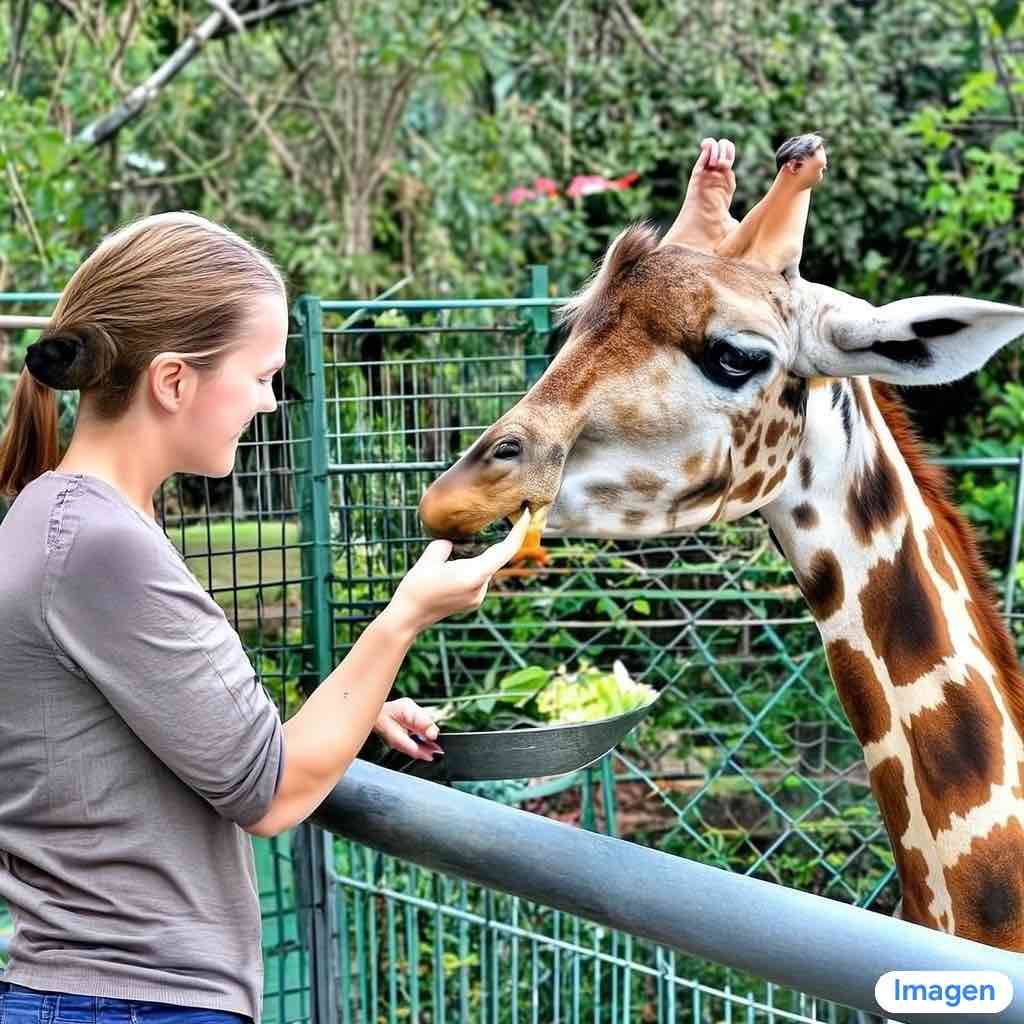} \\
& \multicolumn{3}{p{0.75\textwidth}}{\footnotesize \fontfamily{cmss}\selectfont [GT Caption] A woman standing with by a giraffe at a fence, and feeding it, with trees and shrubs behind.} \\
\end{tabular}
\vspace{-8pt}
\caption{
\textbf{Text-to-image reconstruction with De-Diffusion text and ground-truth captions.}
The original images are from MS-COCO 2014 val split. We highlight \textcolor{citecolor}{\textbf{different visual aspects in green}}.
}
\label{fig:viz-coco-1}
\end{figure*}

\begin{figure*}[t]
\centering
\tablestyle{0.0pt}{0.5}
\begin{tabular}{q{0.25\linewidth}k{0.25\linewidth}q{0.25\linewidth}v{0.25\linewidth}}
Original Image & 
\multicolumn{1}{c}{Stable Diffusion XL} &
Midjourney &
\multicolumn{1}{c}{Imagen} \\
\\[-3pt]
\multirow{2}{*}[4em]{\includegraphics[width=0.9\linewidth]{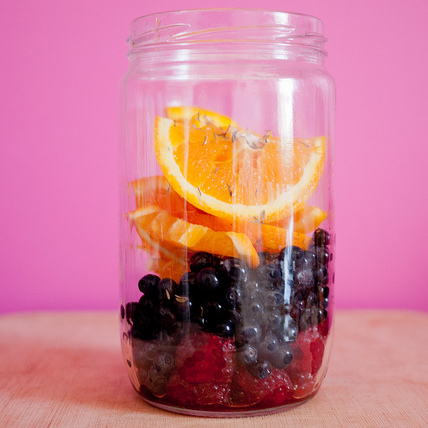}} &
\includegraphics[width=0.96\linewidth]{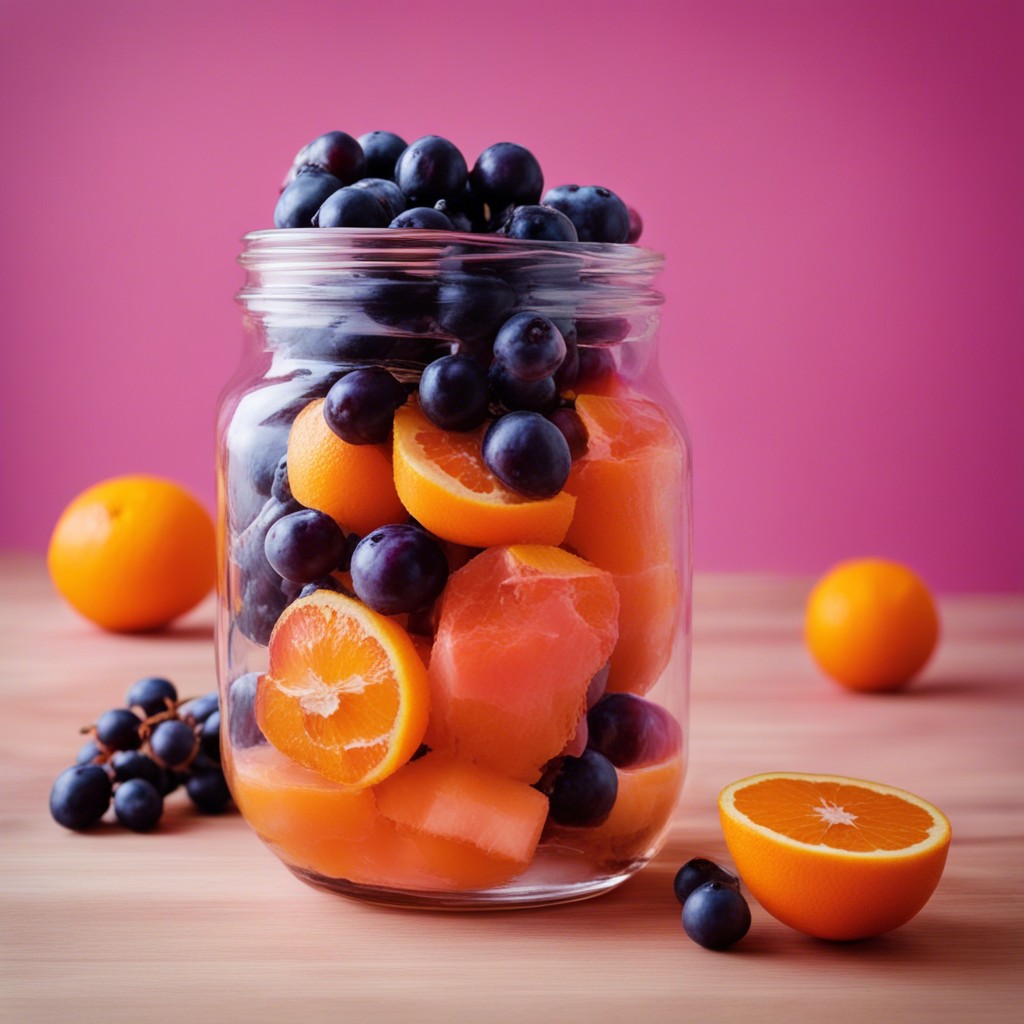} &
\includegraphics[width=0.96\linewidth]{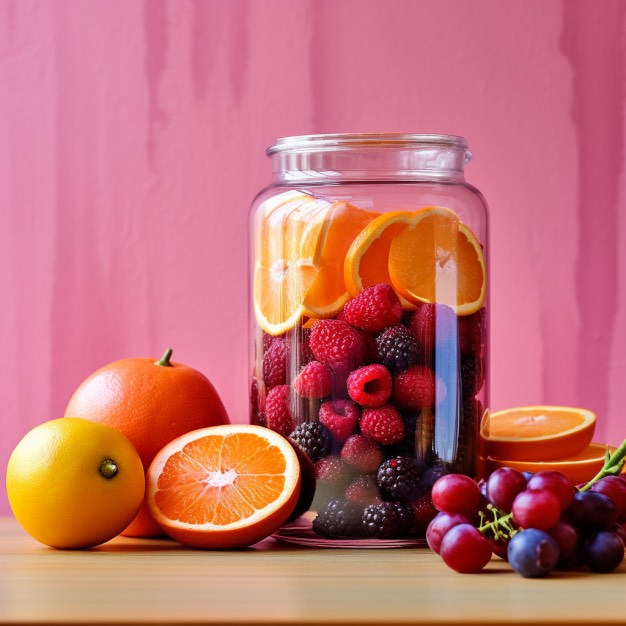} &
\includegraphics[width=0.96\linewidth]{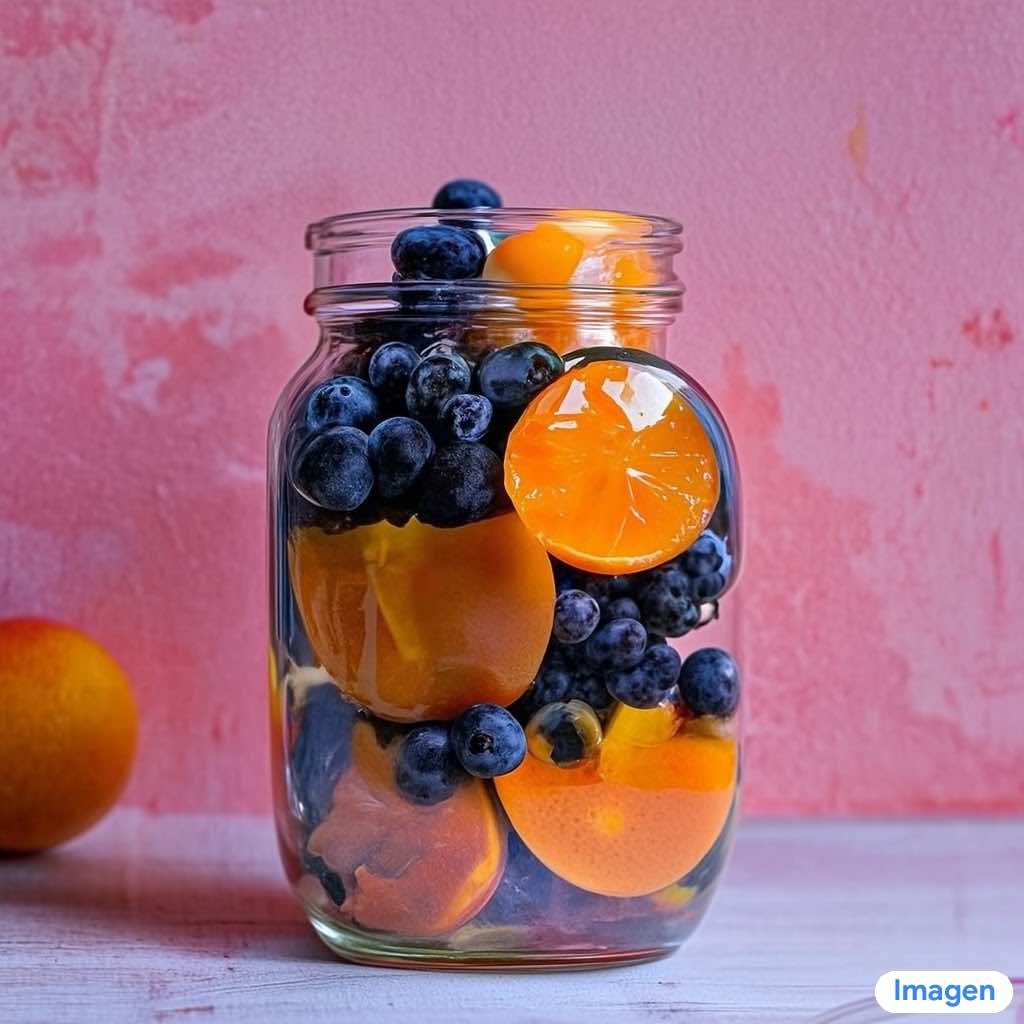} \\
& \multicolumn{3}{p{0.75\textwidth}}{
\footnotesize \fontfamily{cmss}\selectfont
[De-Diffusion Text] an davilishlishblog closeup berries jar through largerefrerefrejar glass jar eachother glass on an on peach hardwood closeup glass homemade osmixed glass jar called relating called an oranges fruit shown slices eachother containing relating an orange orange slices slices between black grapes open chunks orange oranges and berry blackblueberry consist though towards pink closeup facing that \textcolor{citecolor}{\textbf{background pink wall background pink pink wall wall}} closeup chia grapes recipe} \\
\\[-3pt]
\multirow{2}{*}[8em]{\textcolor{citecolor}{\textbf{backgrounds}}} &
\includegraphics[width=0.96\linewidth]{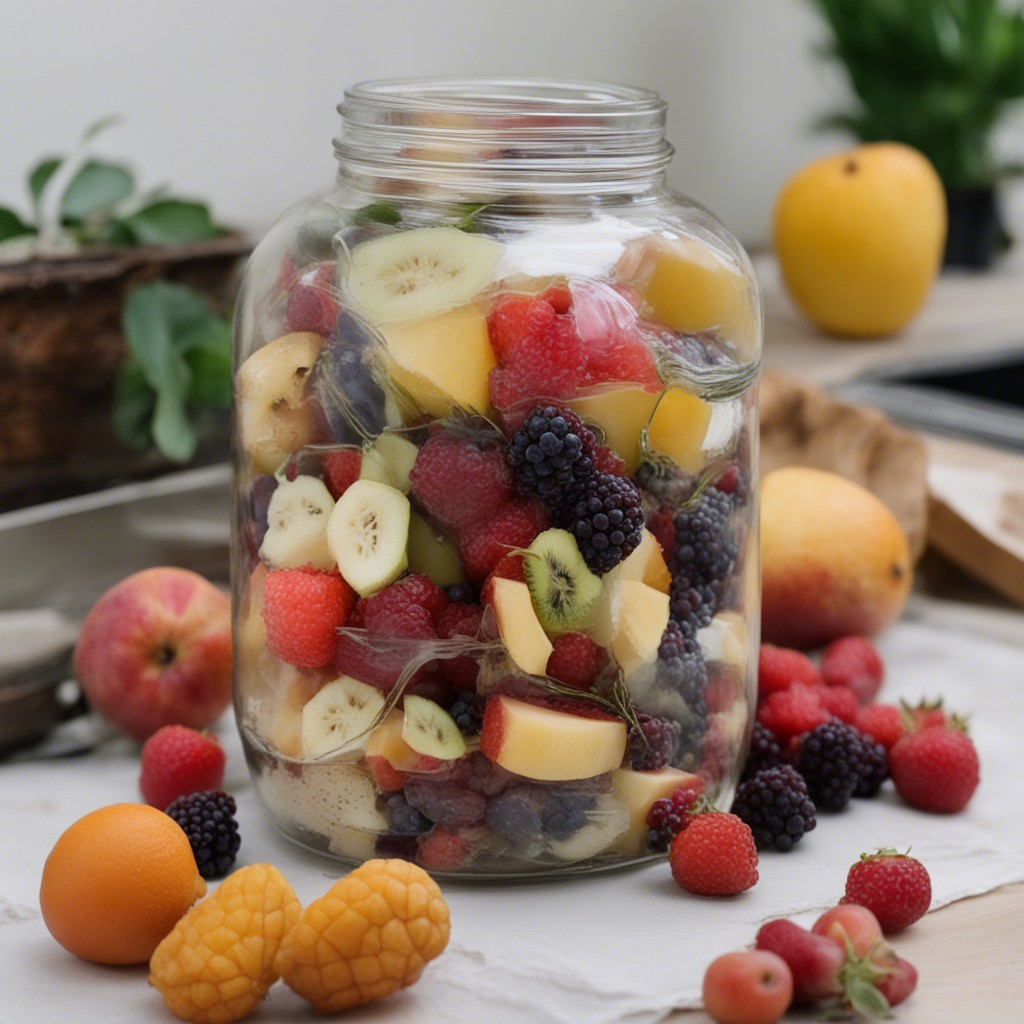} & 
\includegraphics[width=0.96\linewidth]{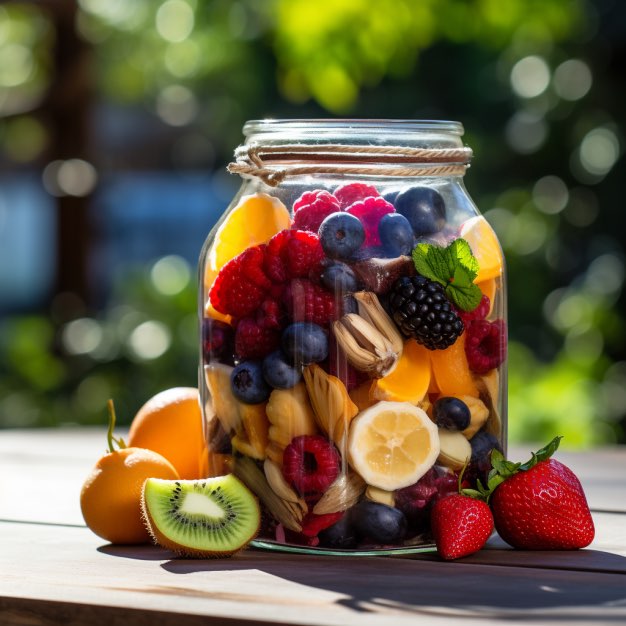} &
\includegraphics[width=0.96\linewidth]{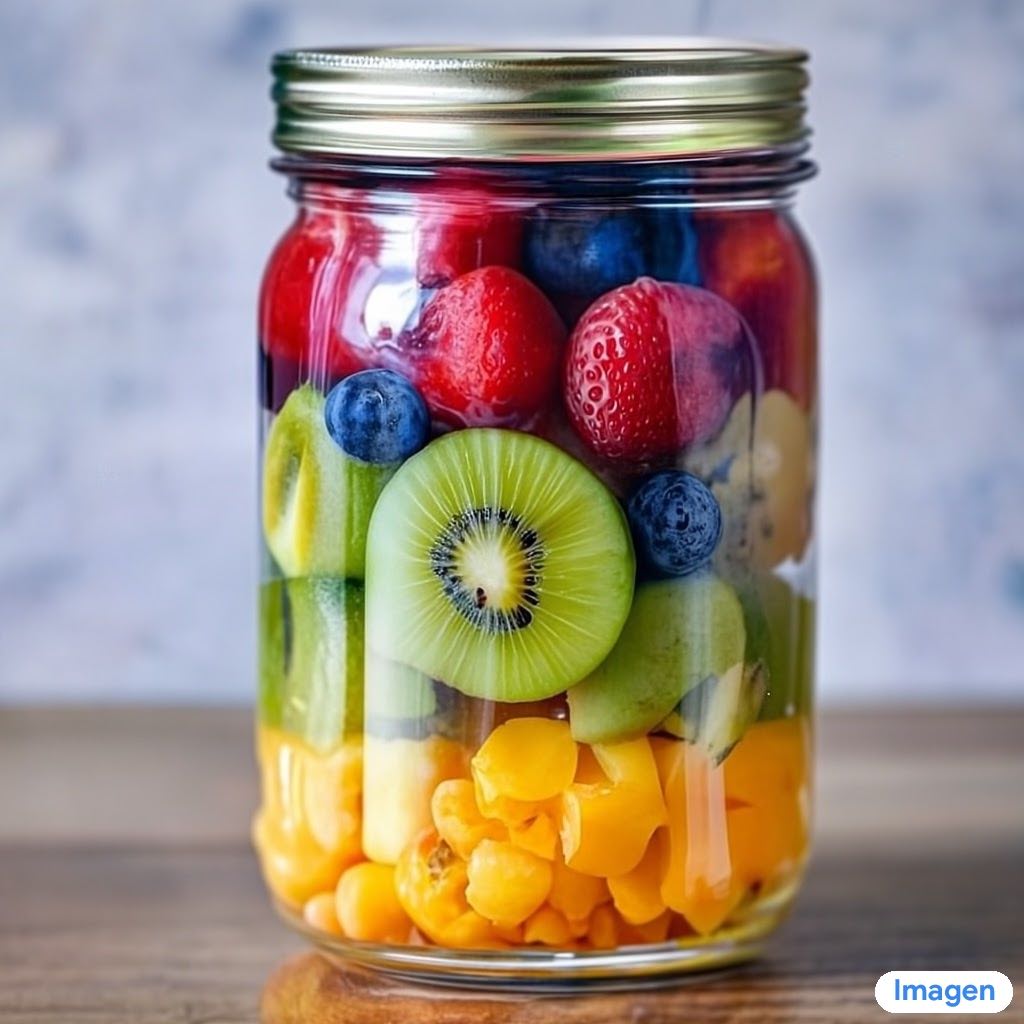} \\
& \multicolumn{3}{p{0.74\textwidth}}{\footnotesize \fontfamily{cmss}\selectfont [GT Caption] A jar filled with different types of fruit on a table.} \\

\shline
\\

\multirow{2}{*}[4em]{\includegraphics[width=0.9\linewidth]{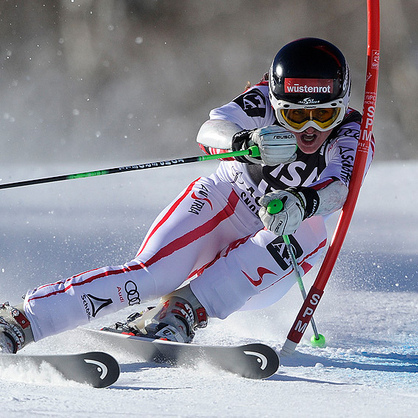}} &
\includegraphics[width=0.96\linewidth]{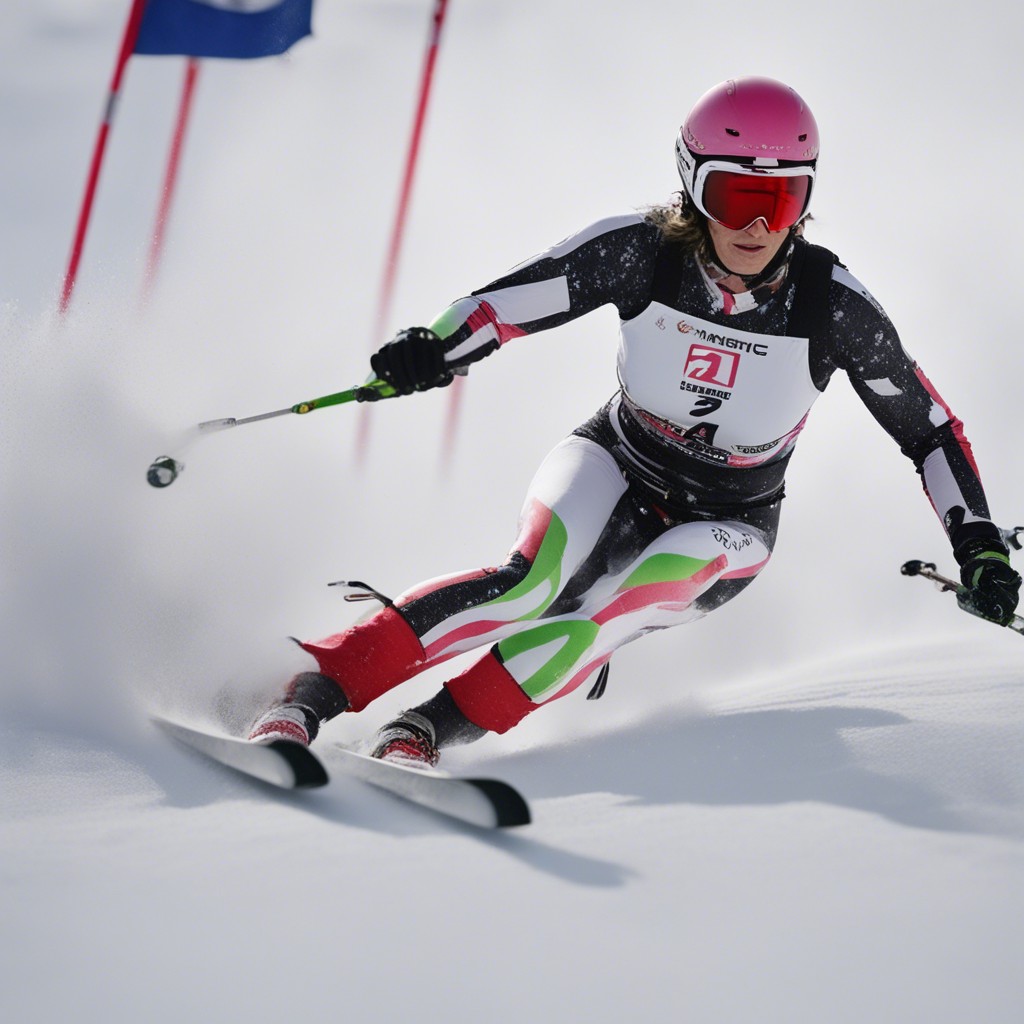} & 
\includegraphics[width=0.96\linewidth]{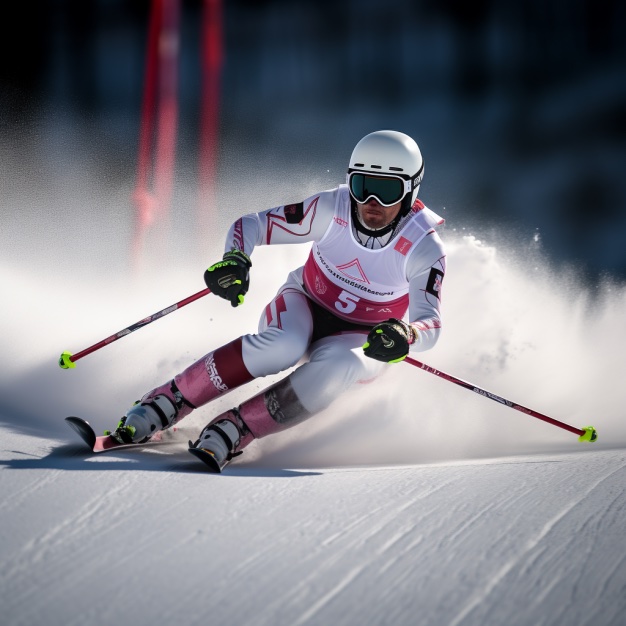} &
\includegraphics[width=0.96\linewidth]{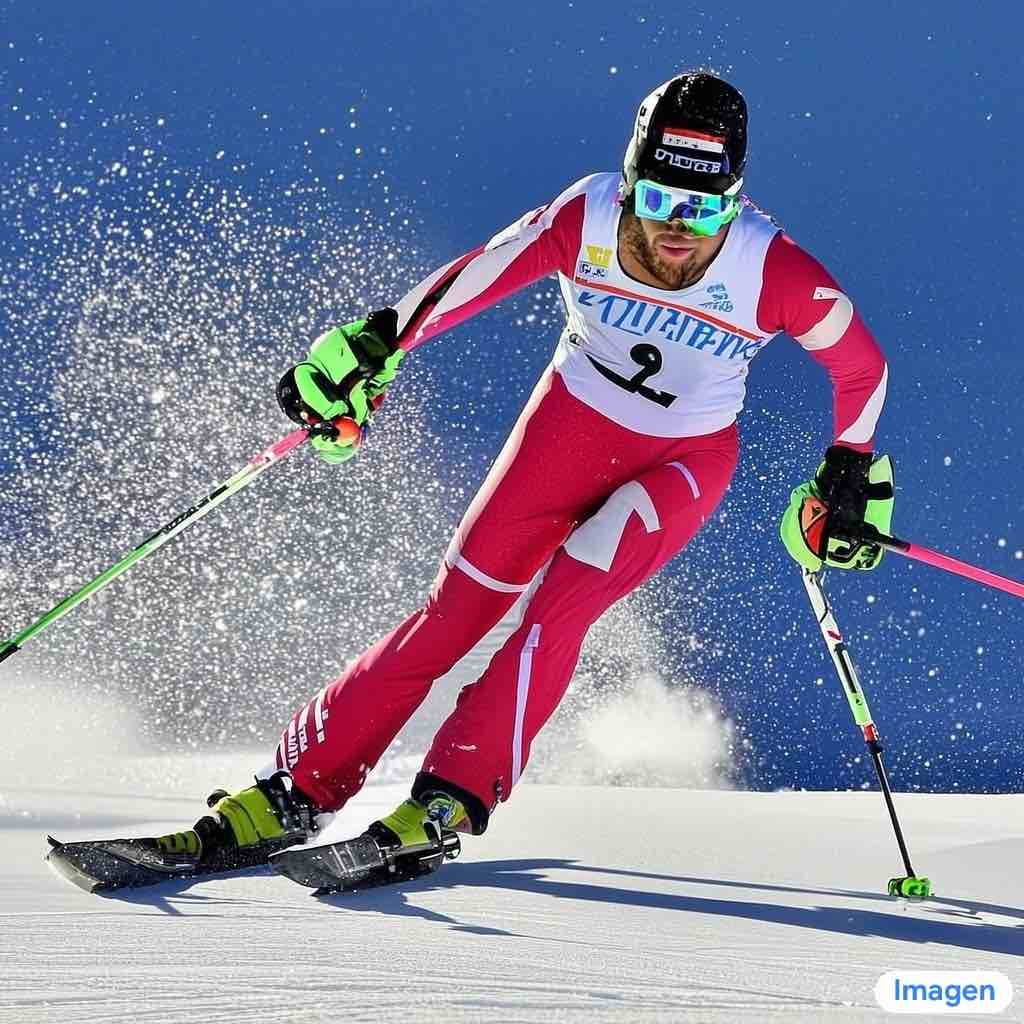} \\
& \multicolumn{3}{p{0.75\textwidth}}{
\footnotesize\fontfamily{cmss}\selectfont
[De-Diffusion Text] an attribusphostavpix person ski skis whose white red dres dres black helmet red pants riding an on snow ski ski austria austria oscompete ski ski resembrelating description an ski person shown action ripping speeds on an a ski stick poles with wearing markings black wheels ilitgoggles silver pink white green consist though towards smoky blur though snow blur smoky smoky but but winter background \textcolor{citecolor}{\textbf{compete compete olympic championship skis}}} \\
\\
\multirow{2}{*}[8em]{\textcolor{citecolor}{\textbf{action subcategories}}} & 
\includegraphics[width=0.96\linewidth]{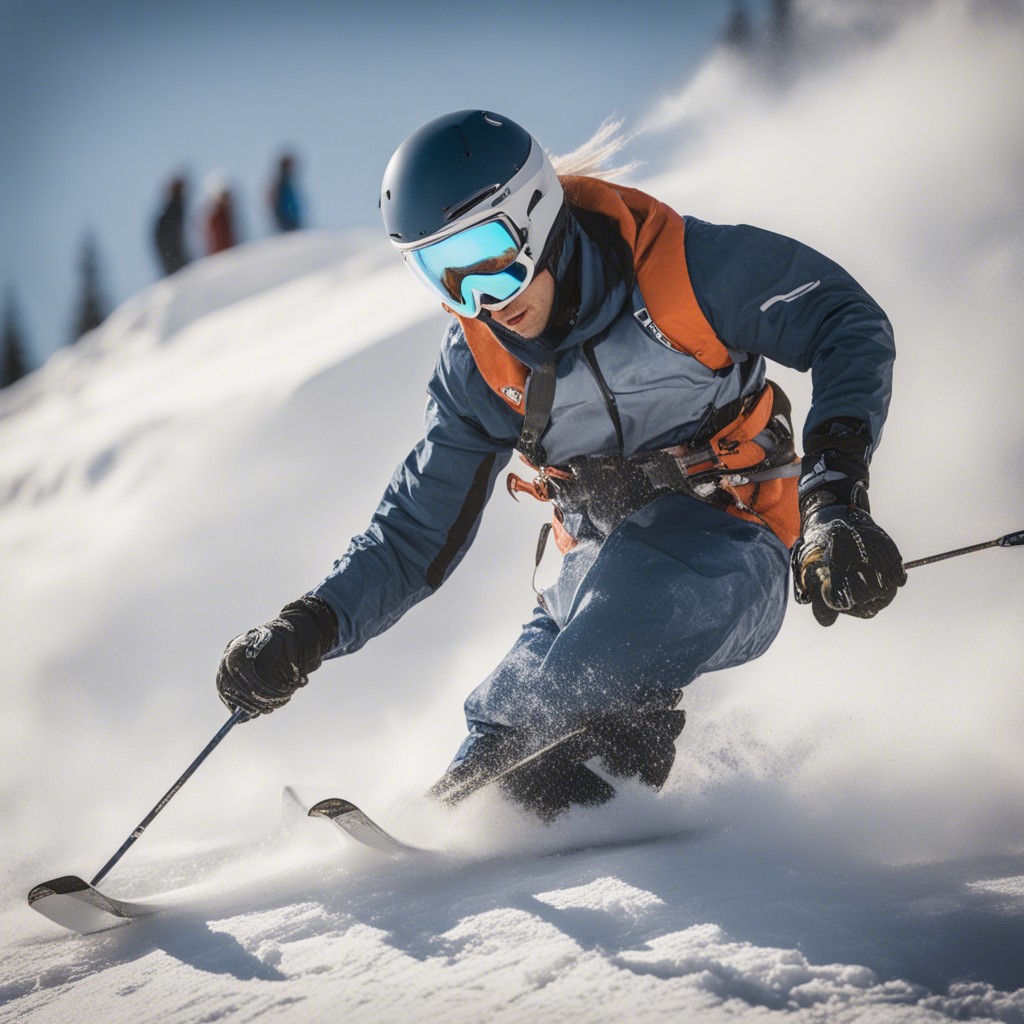} & 
\includegraphics[width=0.96\linewidth]{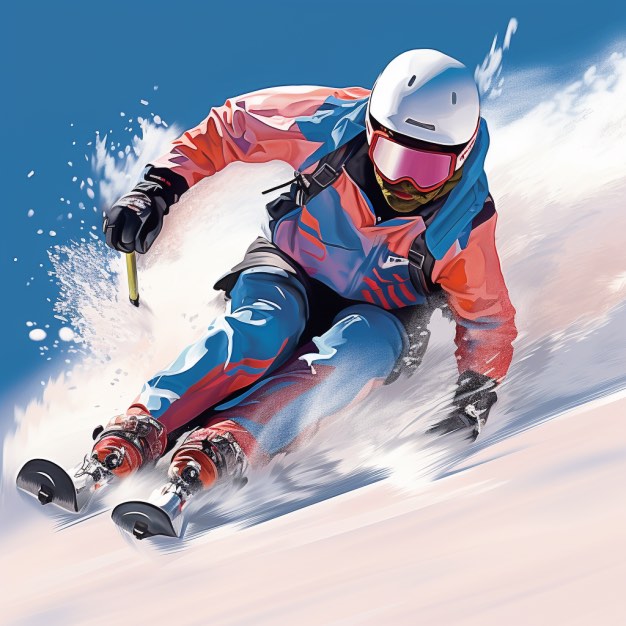} &
\includegraphics[width=0.96\linewidth]{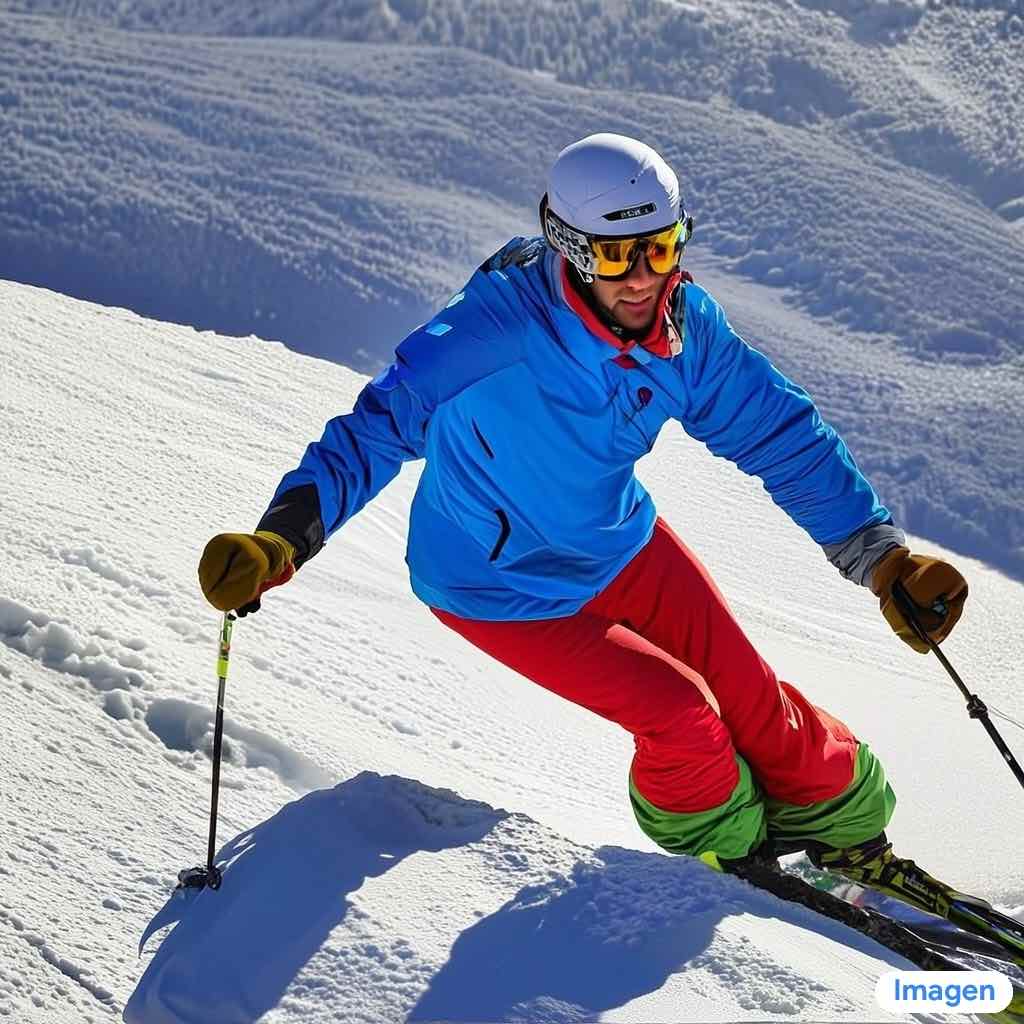} \\
&
\multicolumn{3}{p{0.75\textwidth}}{
\footnotesize\fontfamily{cmss}\selectfont
[GT Caption] A helmeted and goggled skier leans to get around an obstacle.} \\
\end{tabular}
\vspace{-8pt}
\caption{
\textbf{Text-to-image reconstruction with De-Diffusion text and ground-truth captions.}
The original images are from MS-COCO 2014 val split. We highlight \textcolor{citecolor}{\textbf{different visual aspects in green}}.
}
\label{fig:viz-coco-2}
\end{figure*}

\begin{figure*}[t]
\centering
\tablestyle{0pt}{0.3}
\begin{tabular}{y{140}x{119}x{119}x{119}}
\multicolumn{1}{c}{Original Image\qquad\qquad} & Stable Diffusion XL & Midjourney  & Imagen \\
\\
\includegraphics[width=0.22\textwidth]{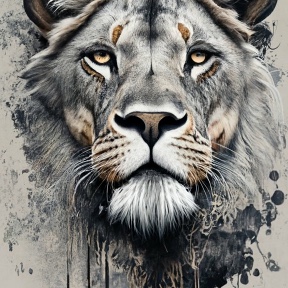} &
\includegraphics[width=0.22\textwidth]{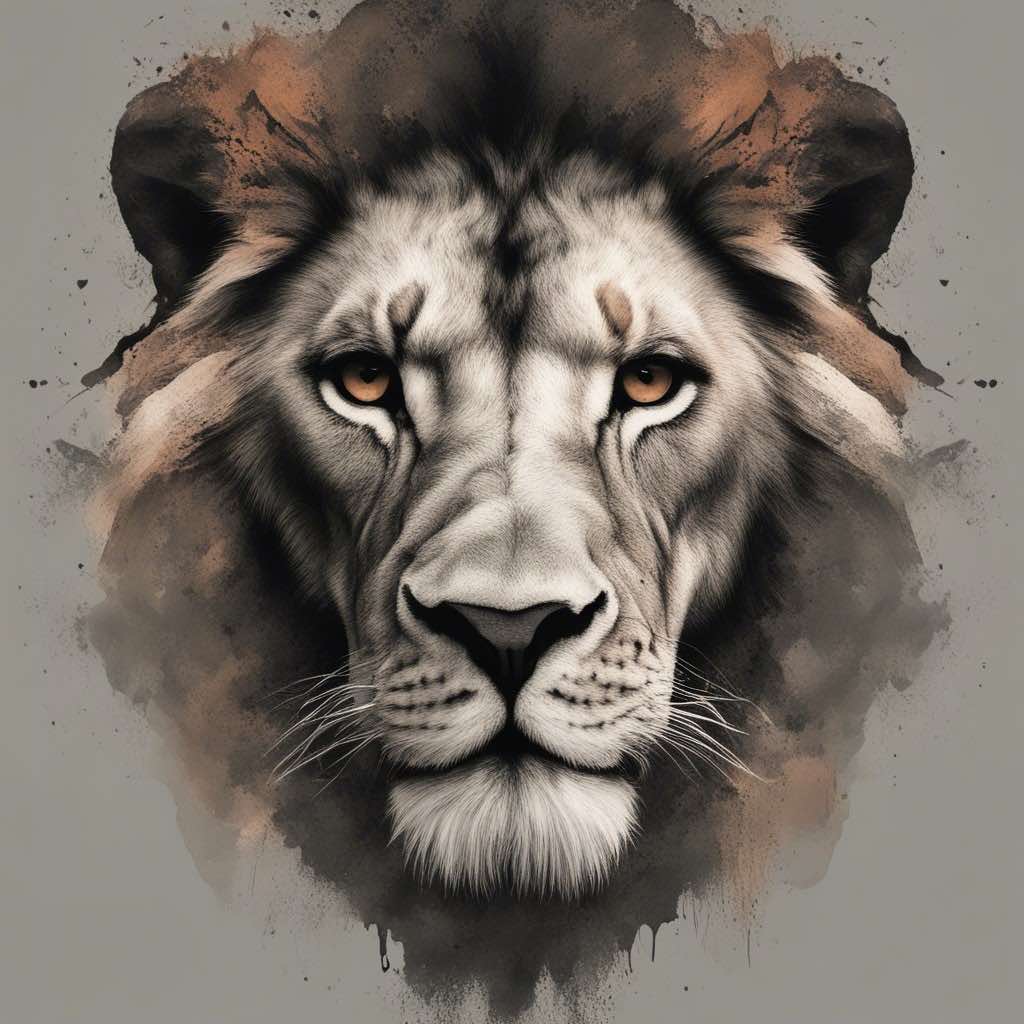} & 
\includegraphics[width=0.22\textwidth]{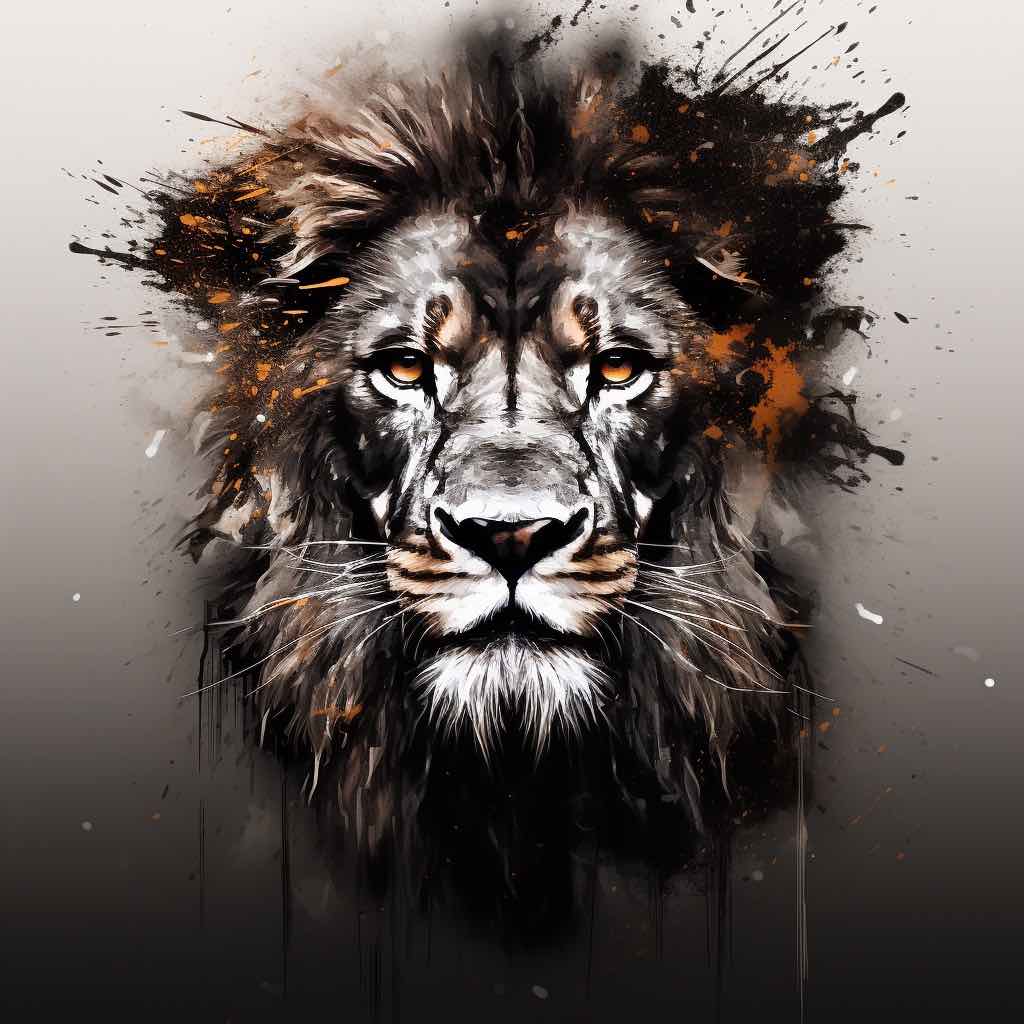} &
\includegraphics[width=0.22\textwidth]{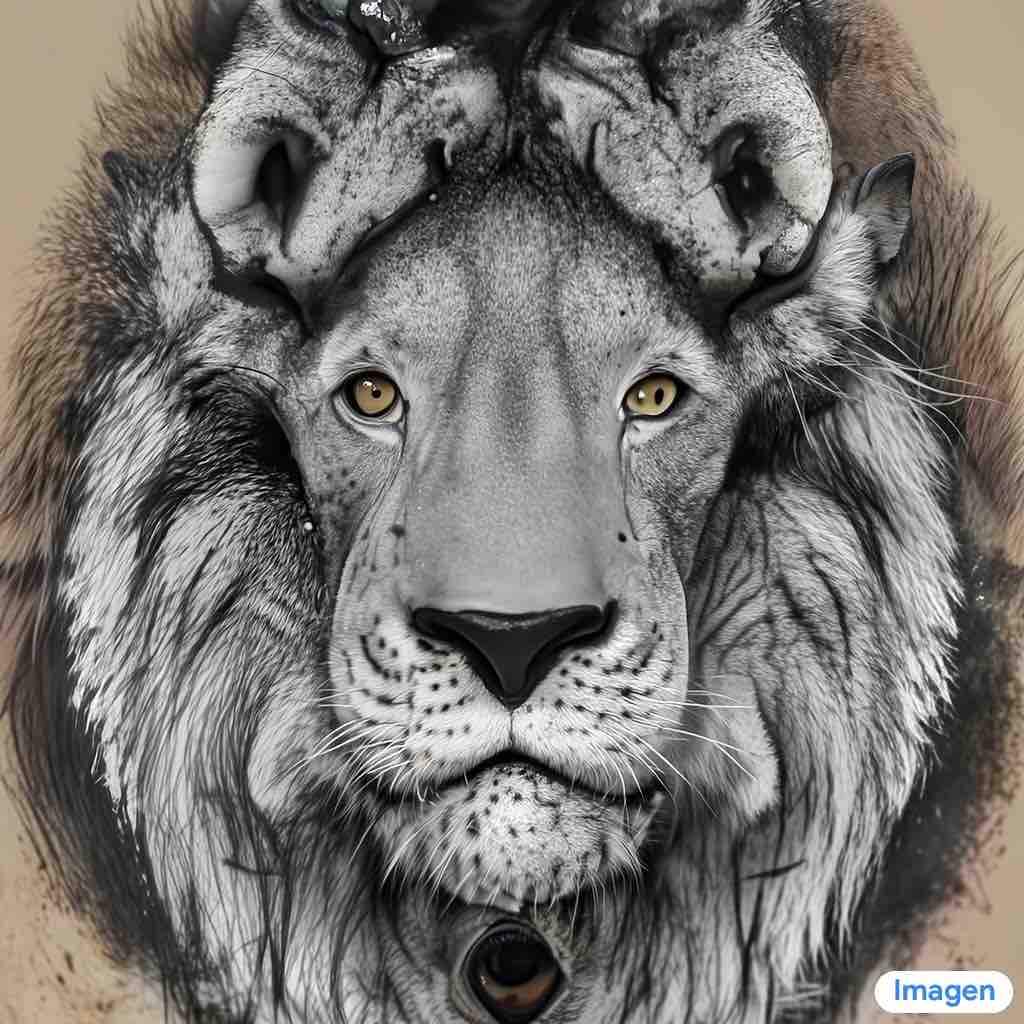} \\
\multicolumn{4}{p{0.99\linewidth}}{\footnotesize \fontfamily{cmss}\selectfont [De-Diffusion Text] an disponsphographic provided graphic a lion jpg grey grey known lion serious face closeup face though an behind black splatbackdrop realism britanniosanimal animal animal creativeexhibiting called an a lion shown looking frontal upwards towards an black splatsplatblot with with dripping copper eyed copper markings blackandbronze grey monochrome overlooking scattered with black splatbehind splatchaos beige background background overcast beige background closeup eyebrow deviantart realism poster} \\
\\
\includegraphics[width=0.22\textwidth]{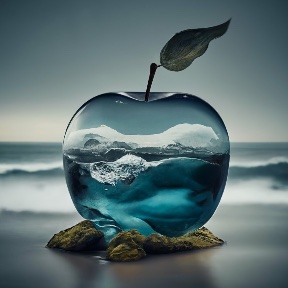} &
\includegraphics[width=0.22\textwidth]{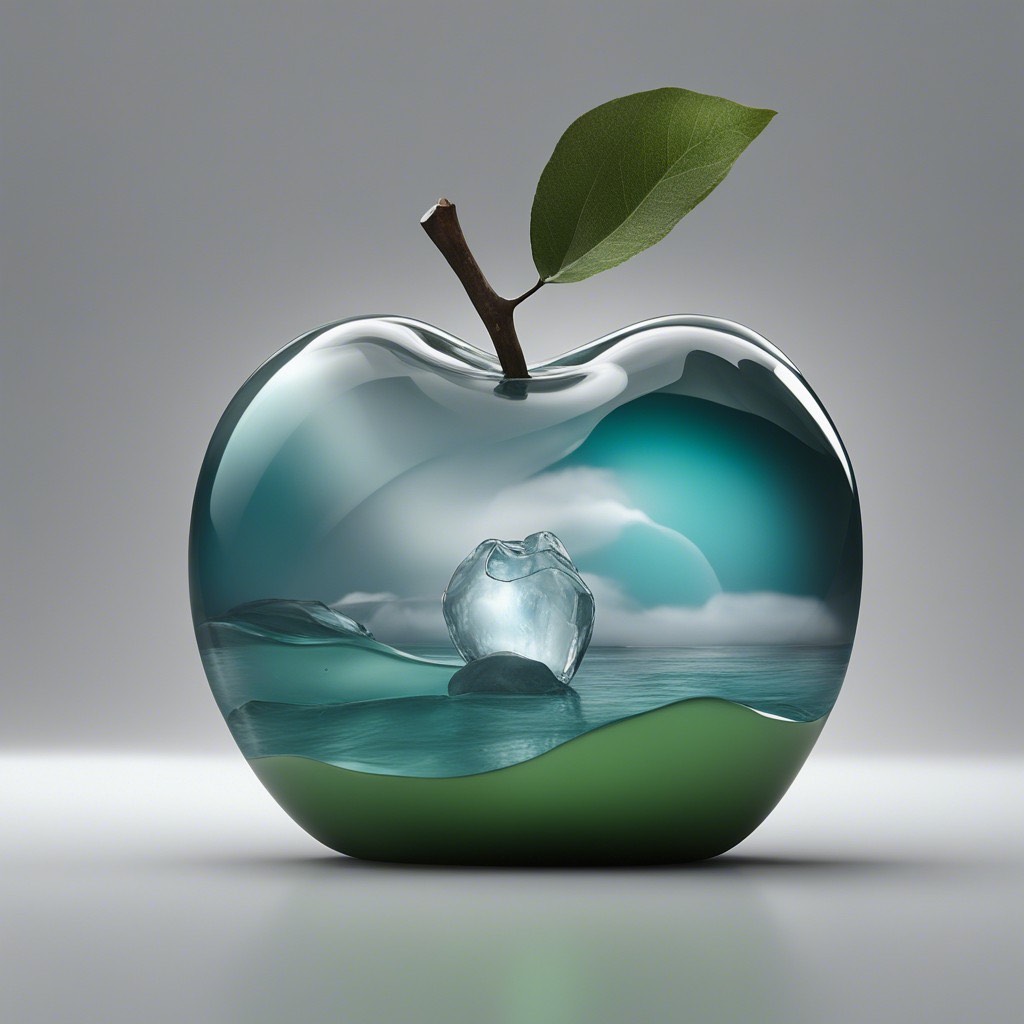} & 
\includegraphics[width=0.22\textwidth]{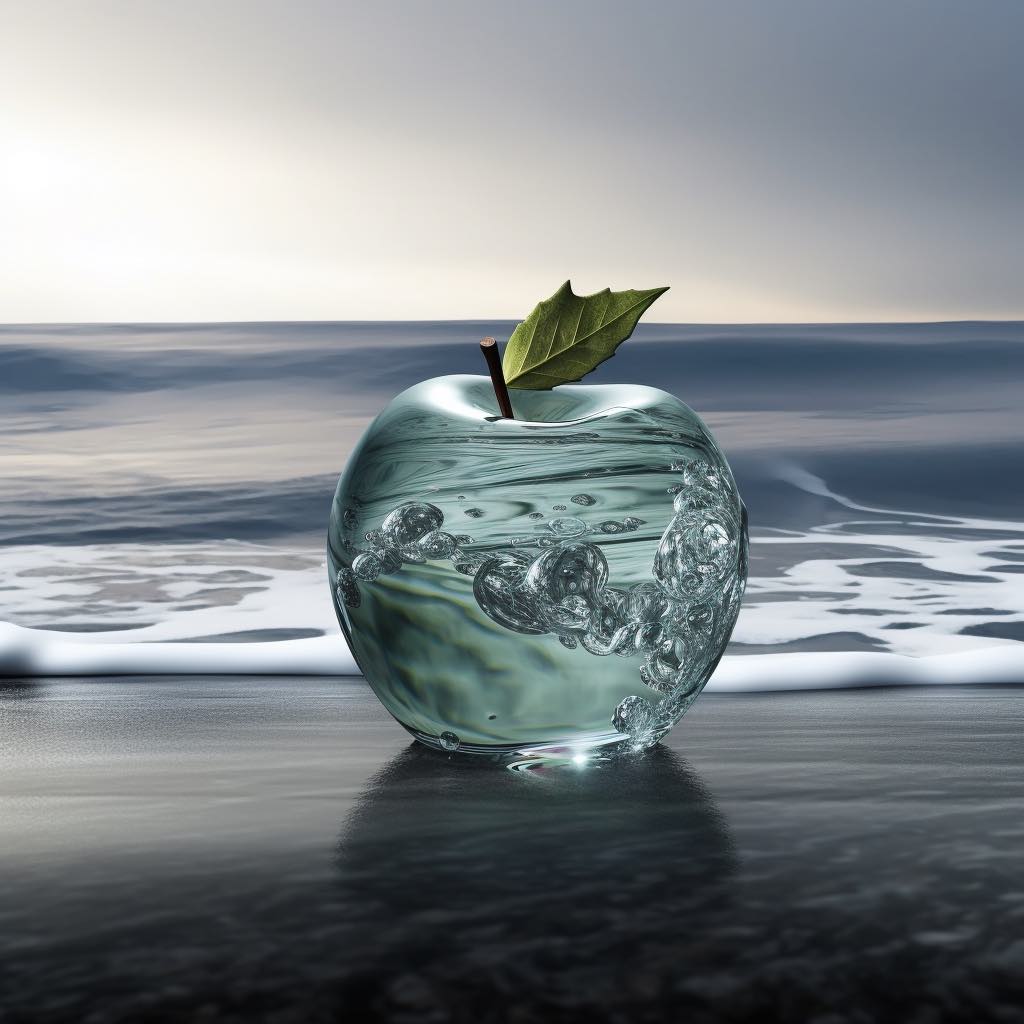} &
\includegraphics[width=0.22\textwidth]{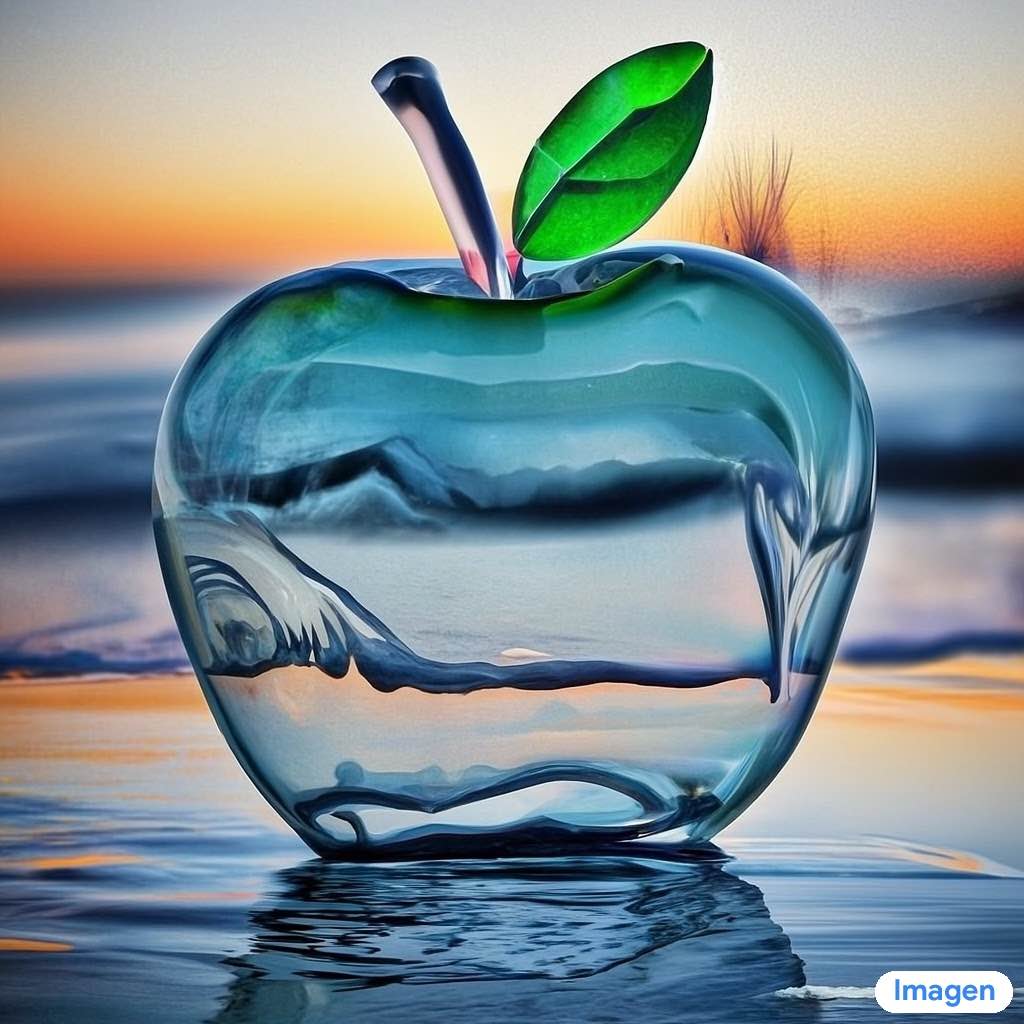} \\
\multicolumn{4}{p{0.99\linewidth}}{\footnotesize \fontfamily{cmss}\selectfont [De-Diffusion Text] an artjvdigitally sart rendering glass apple largelargeblue glass shape with apple with leaf on an on olive lders ashore beach britanniosfuturistic apple apple creativeexhibiting called an glass fruit shown glass incorporating with with an icy icy iceberg iceberg with water boiling with leaf moody sky olive green teal teal placed though on blur waves behind ocean waves grey moody sky grey grey background dusk blur blur blur montage} \\
\\
\includegraphics[width=0.22\textwidth]{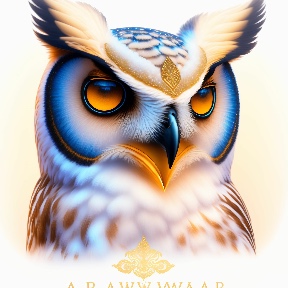} &
\includegraphics[width=0.22\textwidth]{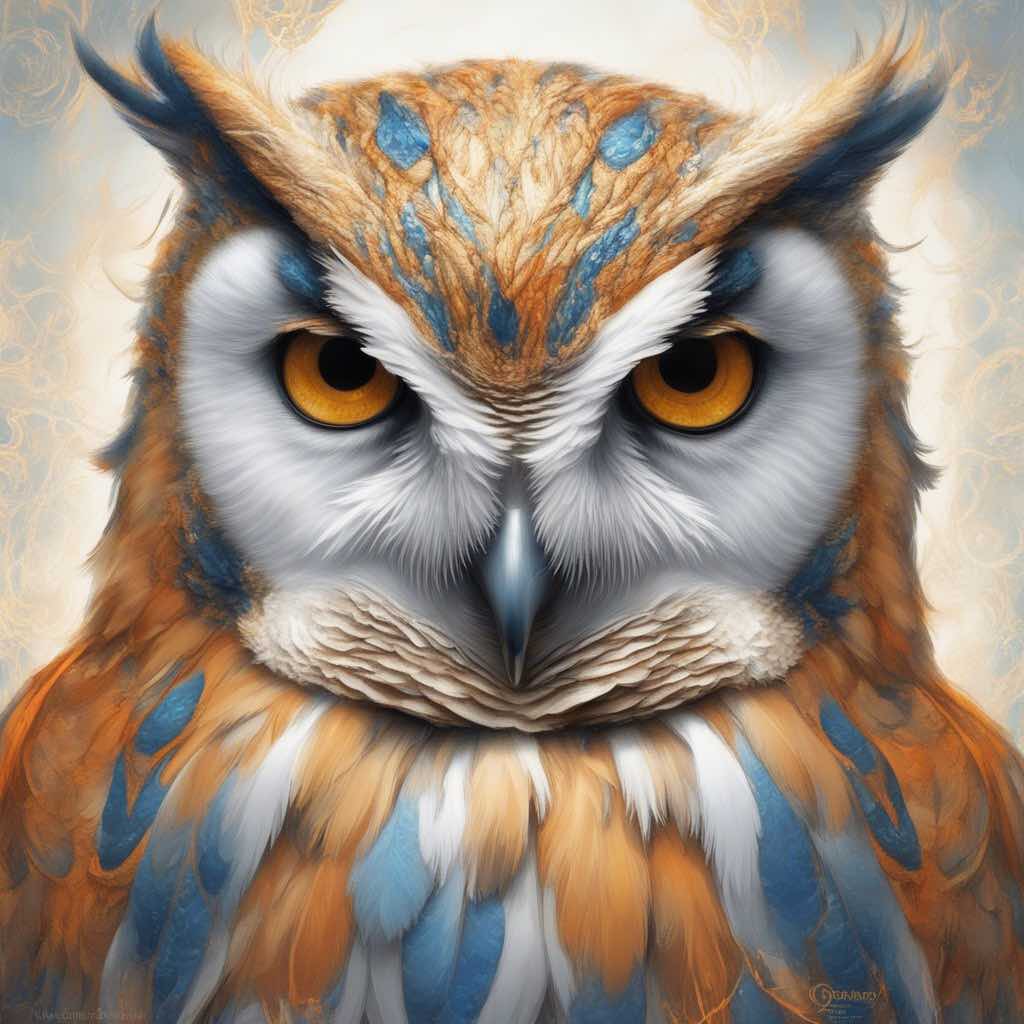} & 
\includegraphics[width=0.22\textwidth]{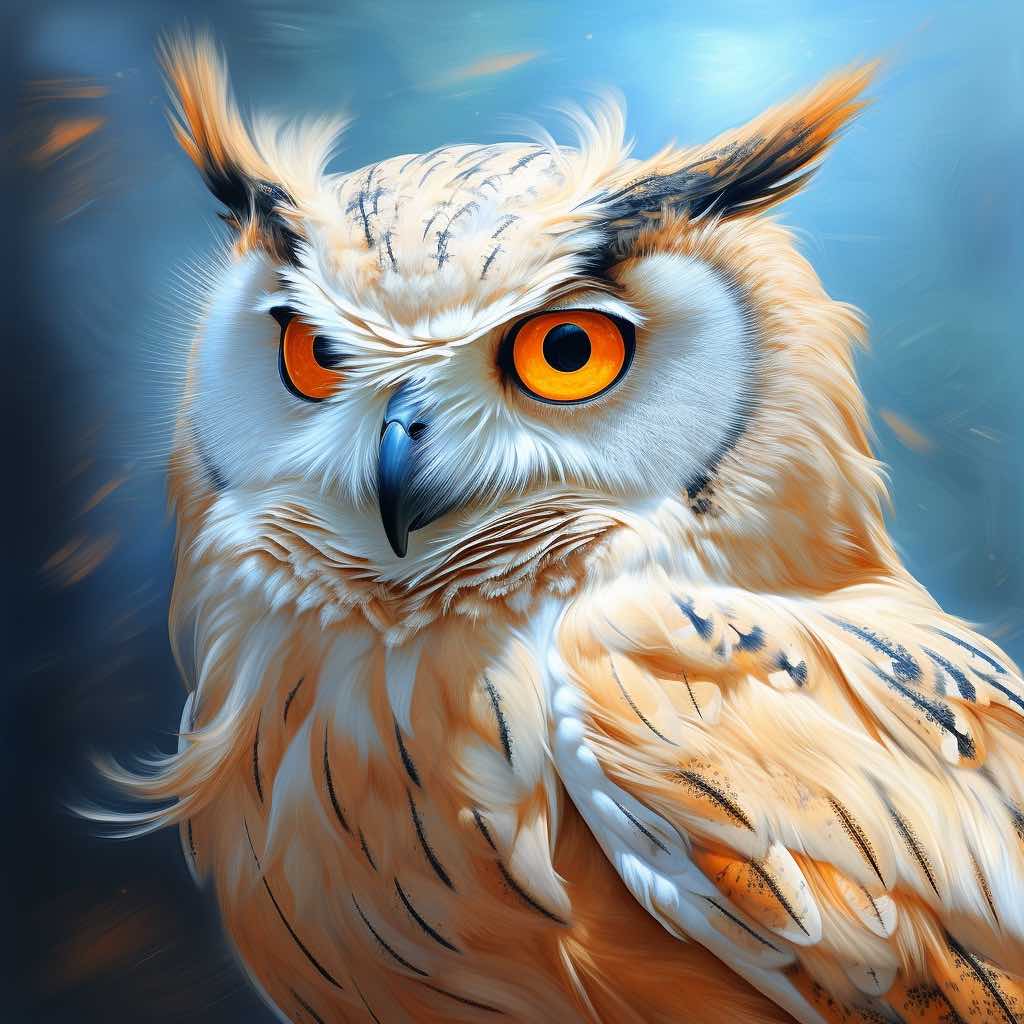} &
\includegraphics[width=0.22\textwidth]{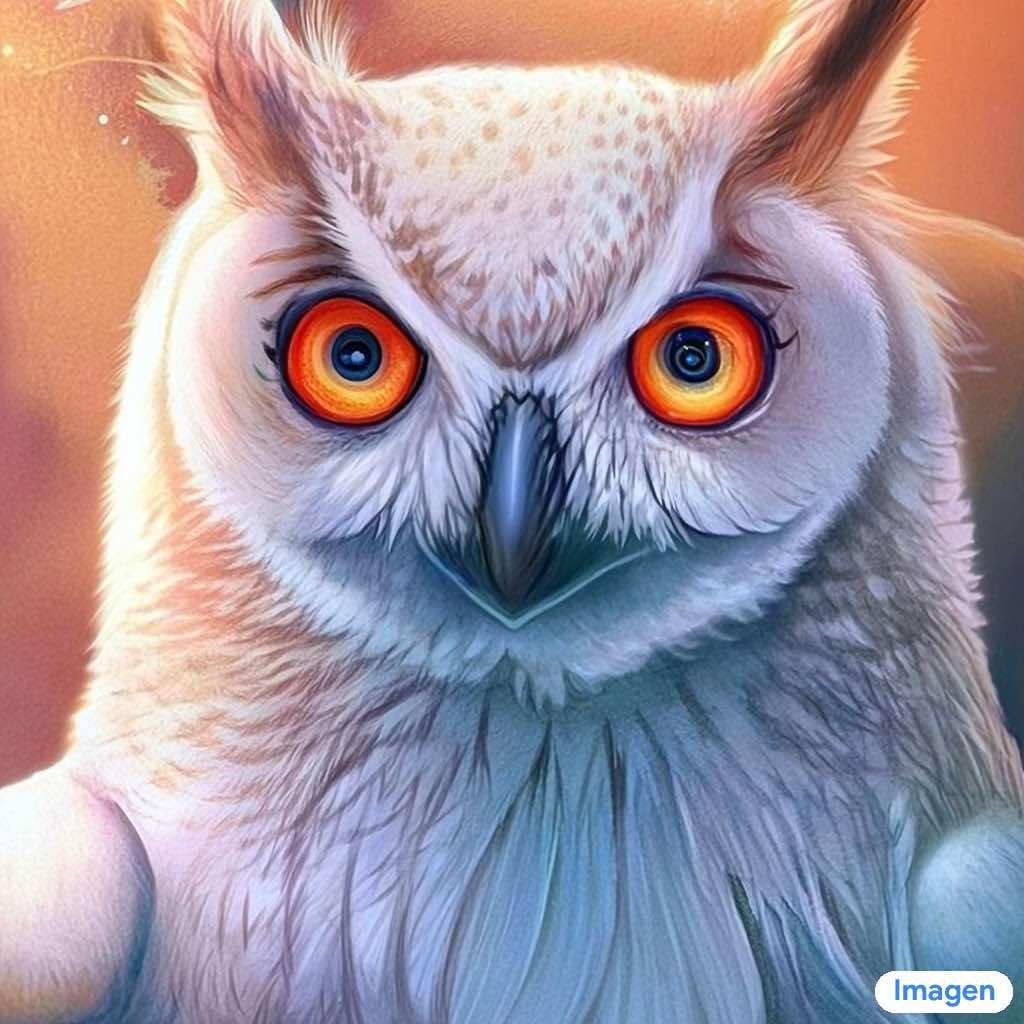} \\
\multicolumn{4}{p{0.99\linewidth}}{\footnotesize \fontfamily{cmss}\selectfont [De-Diffusion Text] an disponsphocgi provided painting a owl beautifully white blue intricate owl closeup beak closeup beak near an near white written realism visionary legendosfantasy animal bird presented description called an animal owl shown beak looking showcasing wearing an gold elaborate winged ears with url text orange lenses orange lenses darker orange blue blues also front on gold stamp on blurry font warm blur on white peach background fps eyebrow fantasy deviantart simulation} \\
\\
\includegraphics[width=0.22\textwidth]{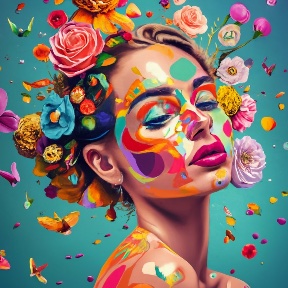} &
\includegraphics[width=0.22\textwidth]{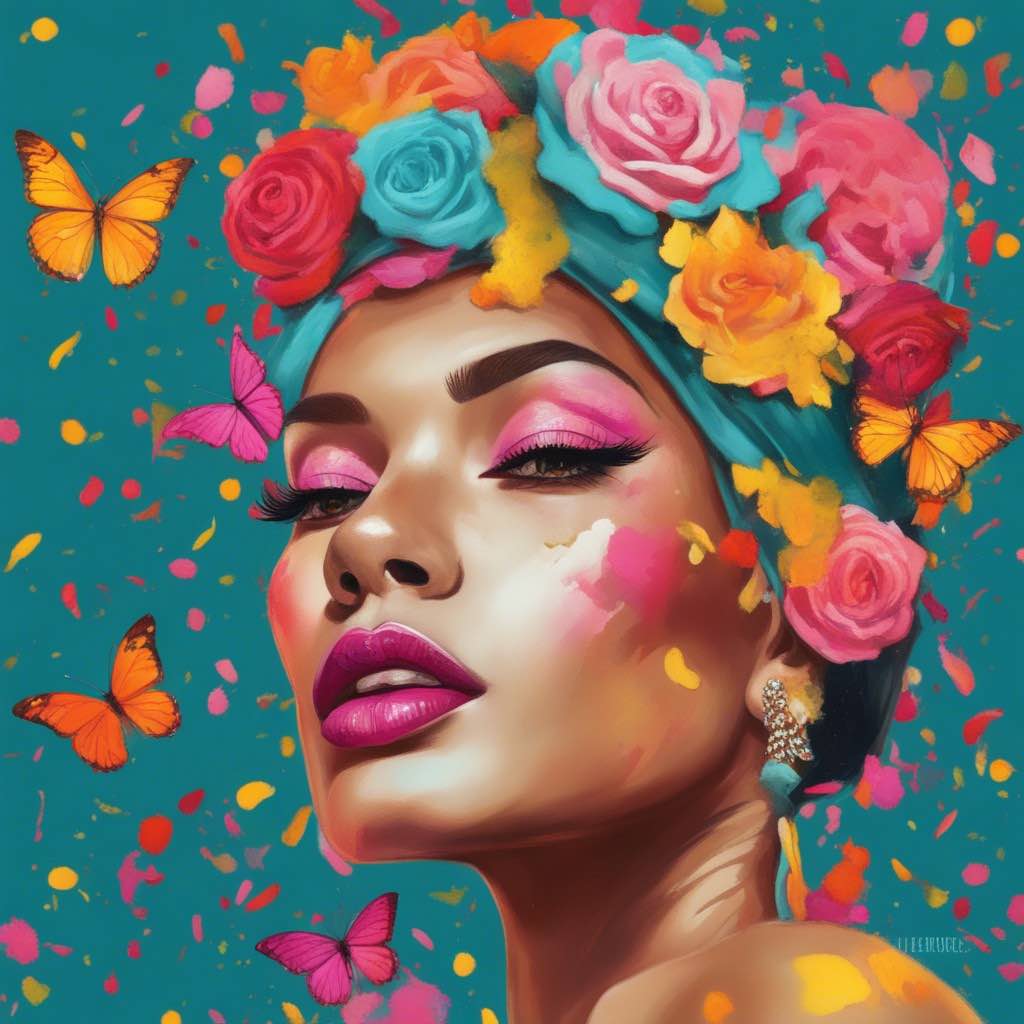} & 
\includegraphics[width=0.22\textwidth]{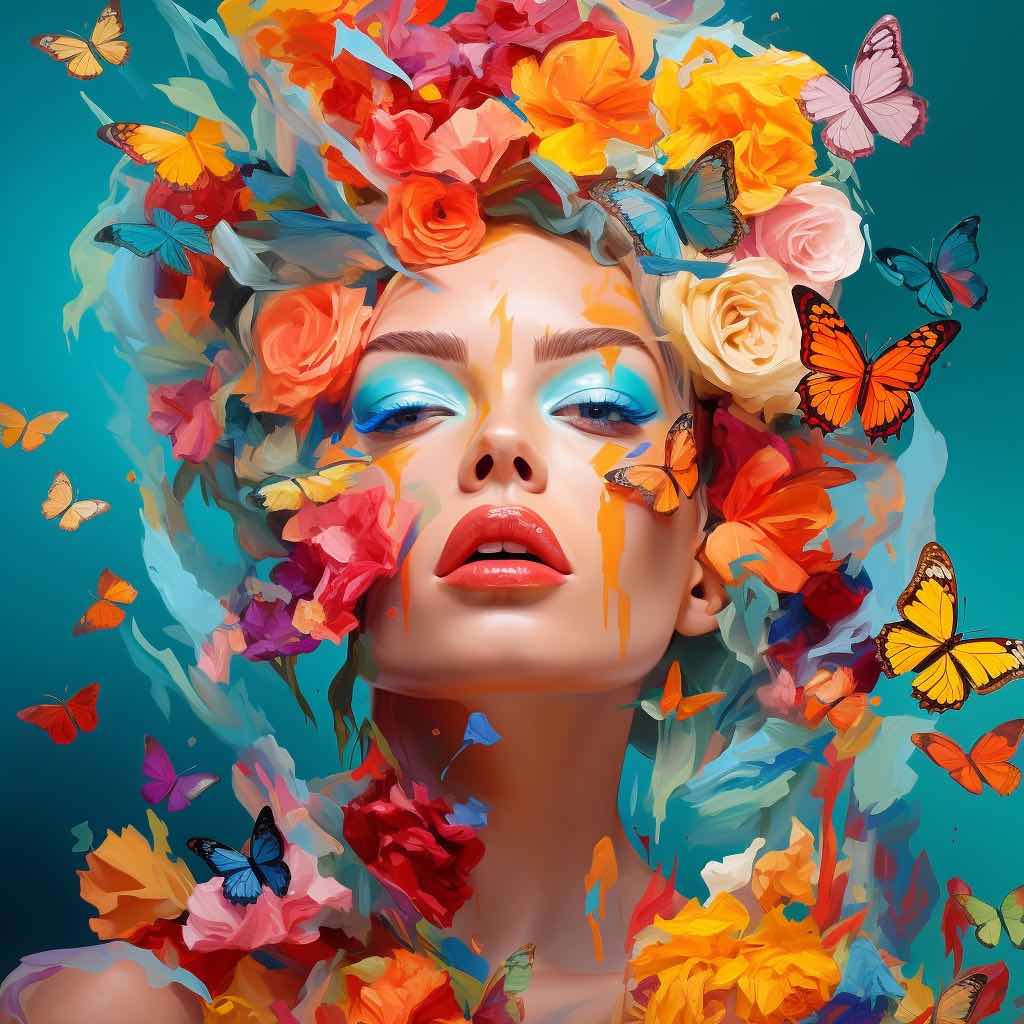} &
\includegraphics[width=0.22\textwidth]{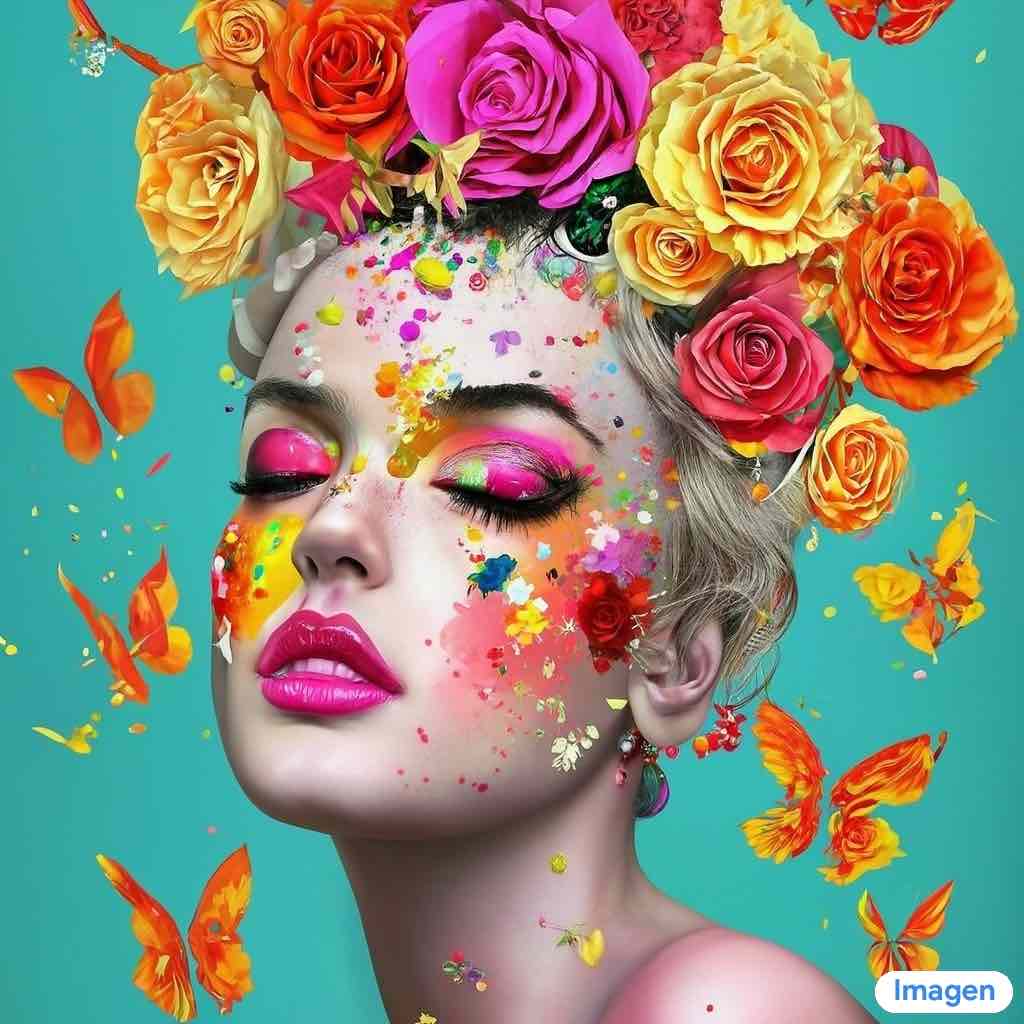} \\
\multicolumn{4}{p{0.99\linewidth}}{\footnotesize \fontfamily{cmss}\selectfont [De-Diffusion Text] an artrhadigitally sart illustration woman face wearing colorful colorful paints face painted head pink lipstick though an among colourful confetti confetti realism pinup osjanumonroe monroe resembrelating called an face woman shown face smelling upwards multiple an colorful florals roses hats above many paints with earrings turmeric makeup brightly orange red pink wth scattered among yellow oranges flying flying butterflies teal background on teal blue background lips eyebrow hadid cg poster} \\
\end{tabular}
\caption{
\textbf{Text-to-image reconstruction with De-Diffusion text.}
Original images are synthetic and we provide their source in \cref{sec:app-links}.
}
\label{fig:viz-syn-1}
\end{figure*}

\begin{figure*}[t]
\centering
\tablestyle{0pt}{0.3}
\begin{tabular}{y{140}x{119}x{119}x{119}}
\multicolumn{1}{c}{Original Image\qquad\qquad} & Stable Diffusion XL & Midjourney  & Imagen \\
\\
\includegraphics[width=0.22\textwidth]{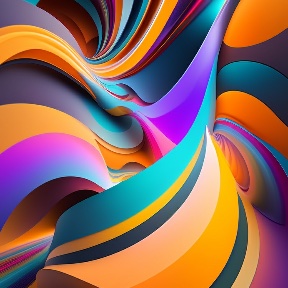} &
\includegraphics[width=0.22\textwidth]{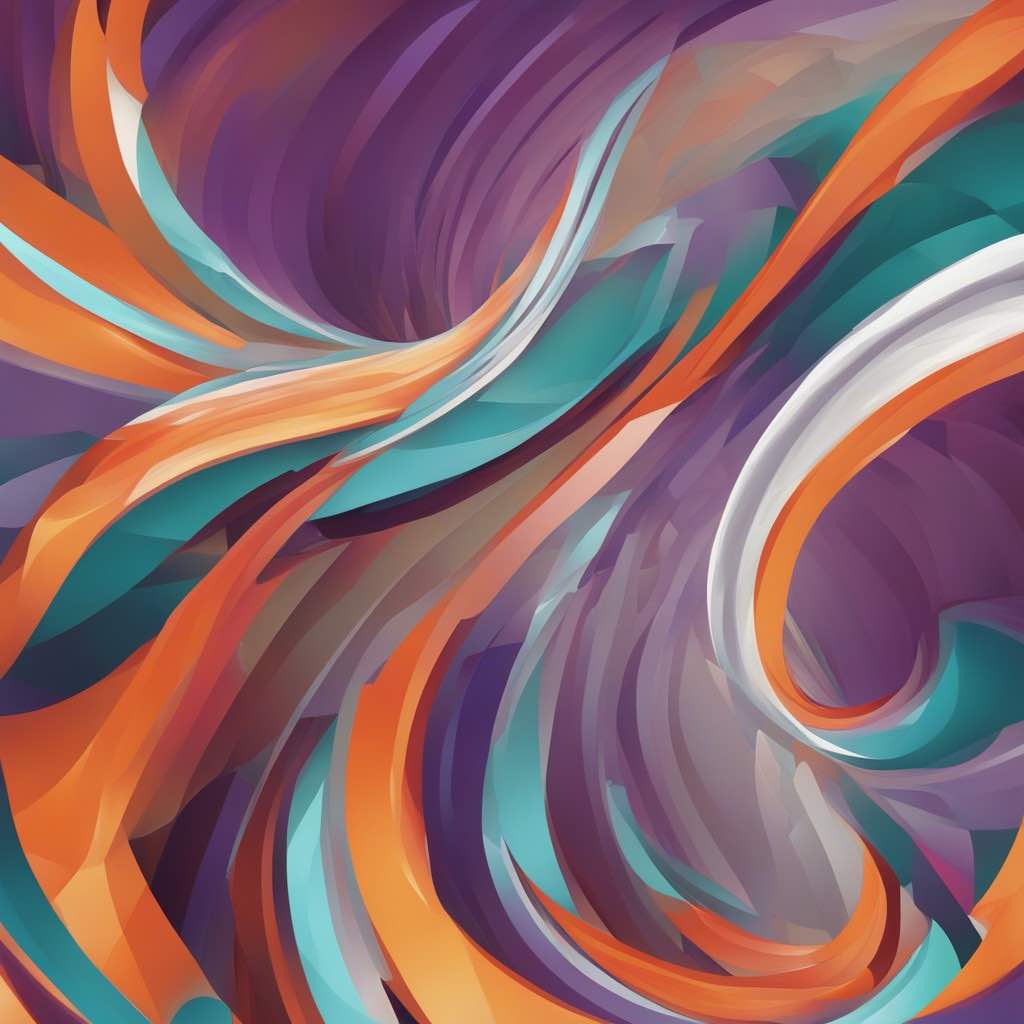} & 
\includegraphics[width=0.22\textwidth]{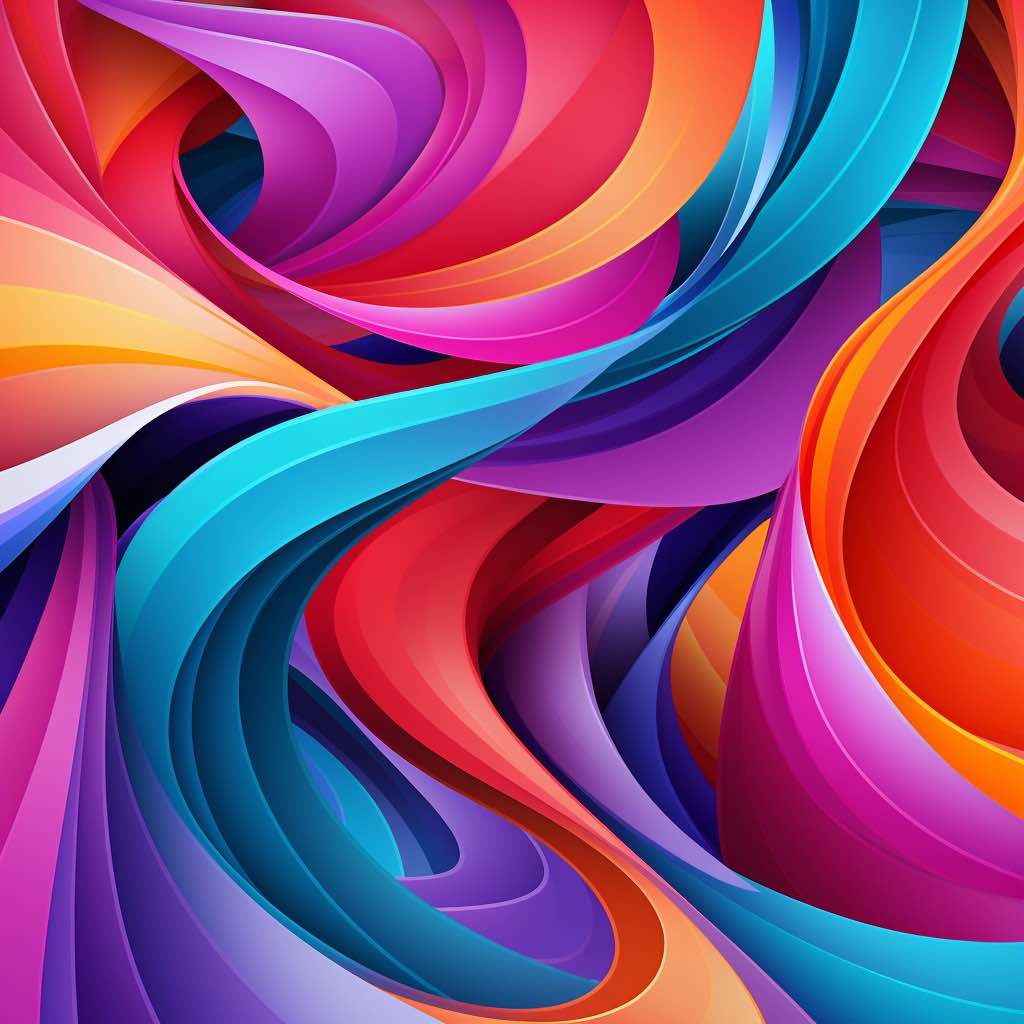} &
\includegraphics[width=0.22\textwidth]{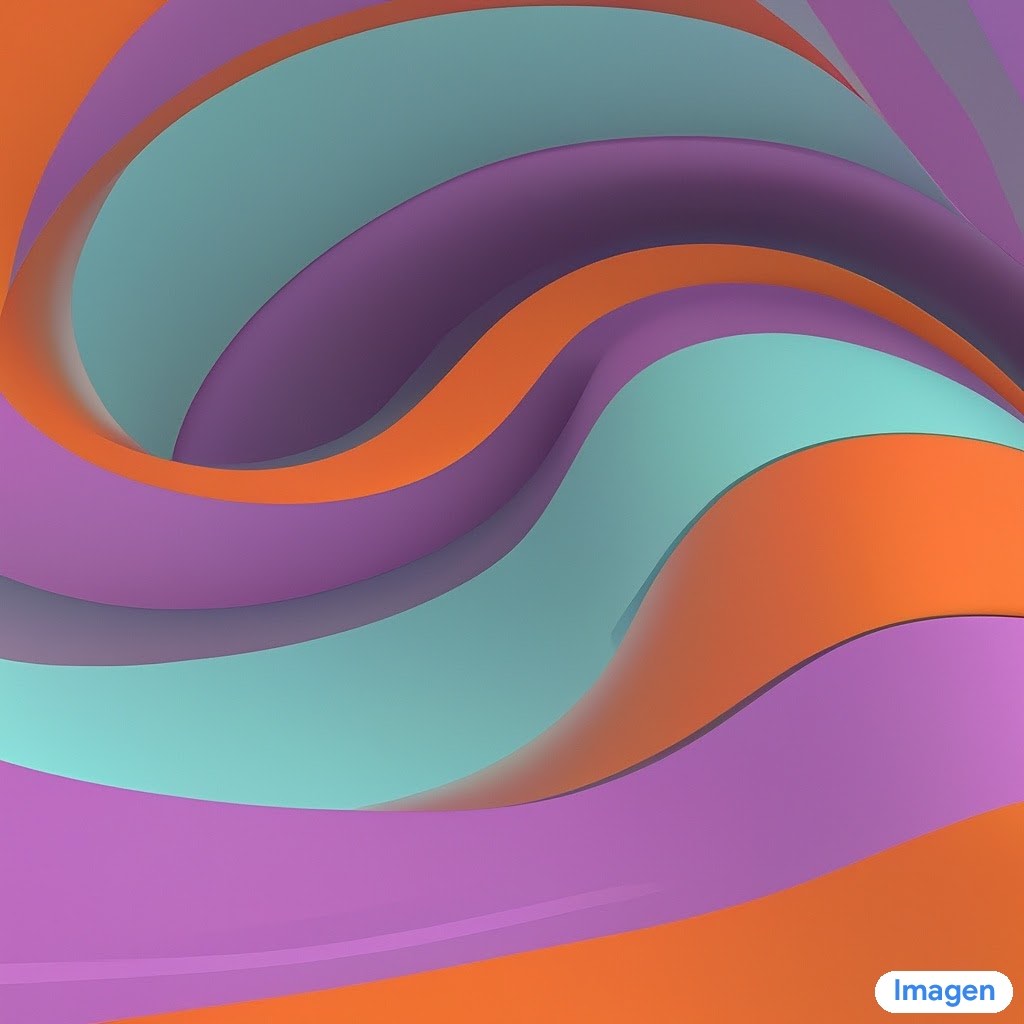} \\
\multicolumn{4}{p{0.99\linewidth}}{\footnotesize \fontfamily{cmss}\selectfont [De-Diffusion Text] an illustration envcesarpixels \textcolor{citecolor}{\textbf{wallpaper}} colorful swirl numerous colorful colorful curved sails bent curved orange angular consist an among colorful curved curved modernist futuristic osfuturistic futuristic cave resembrelating called an colorful swirl shown folded curved resembresemban teal curved curved ribbons while teal colorful colorful swirl orange lines orange orange purple purple consist numerous between orange lines among curved swoop purple stripes but gray purple siding modernist modernist modernist \textcolor{citecolor}{\textbf{cg wallpaper}}} \\
\\
\includegraphics[width=0.22\textwidth]{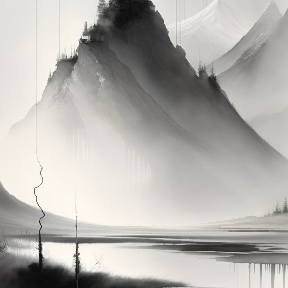} &
\includegraphics[width=0.22\textwidth]{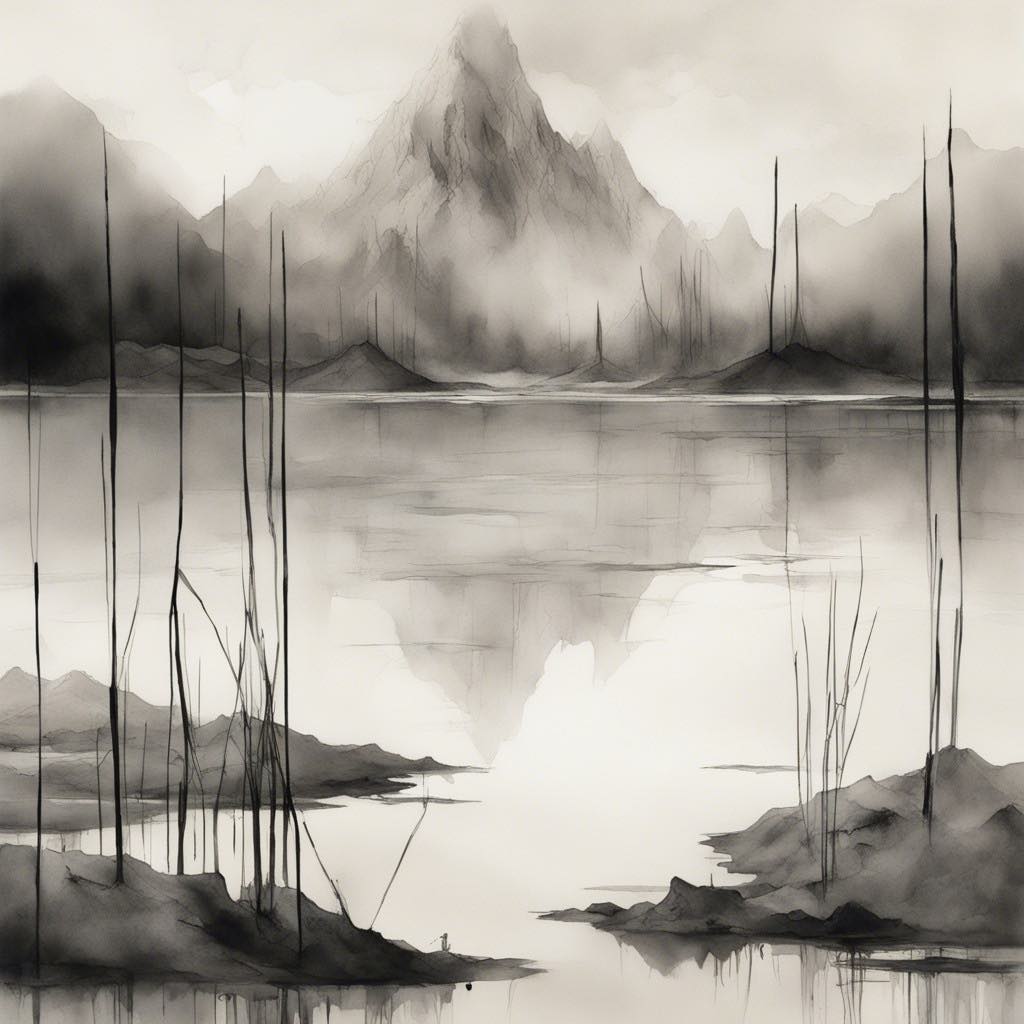} & 
\includegraphics[width=0.22\textwidth]{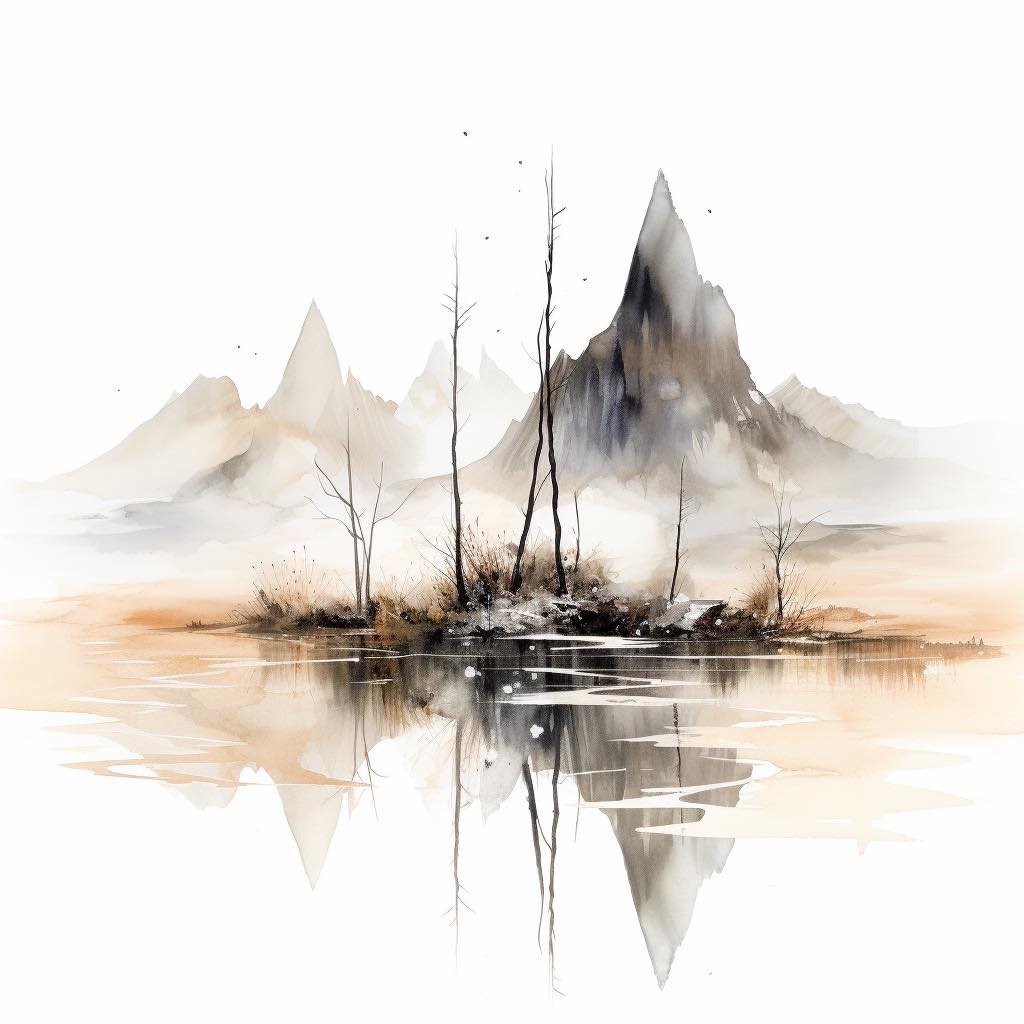} &
\includegraphics[width=0.22\textwidth]{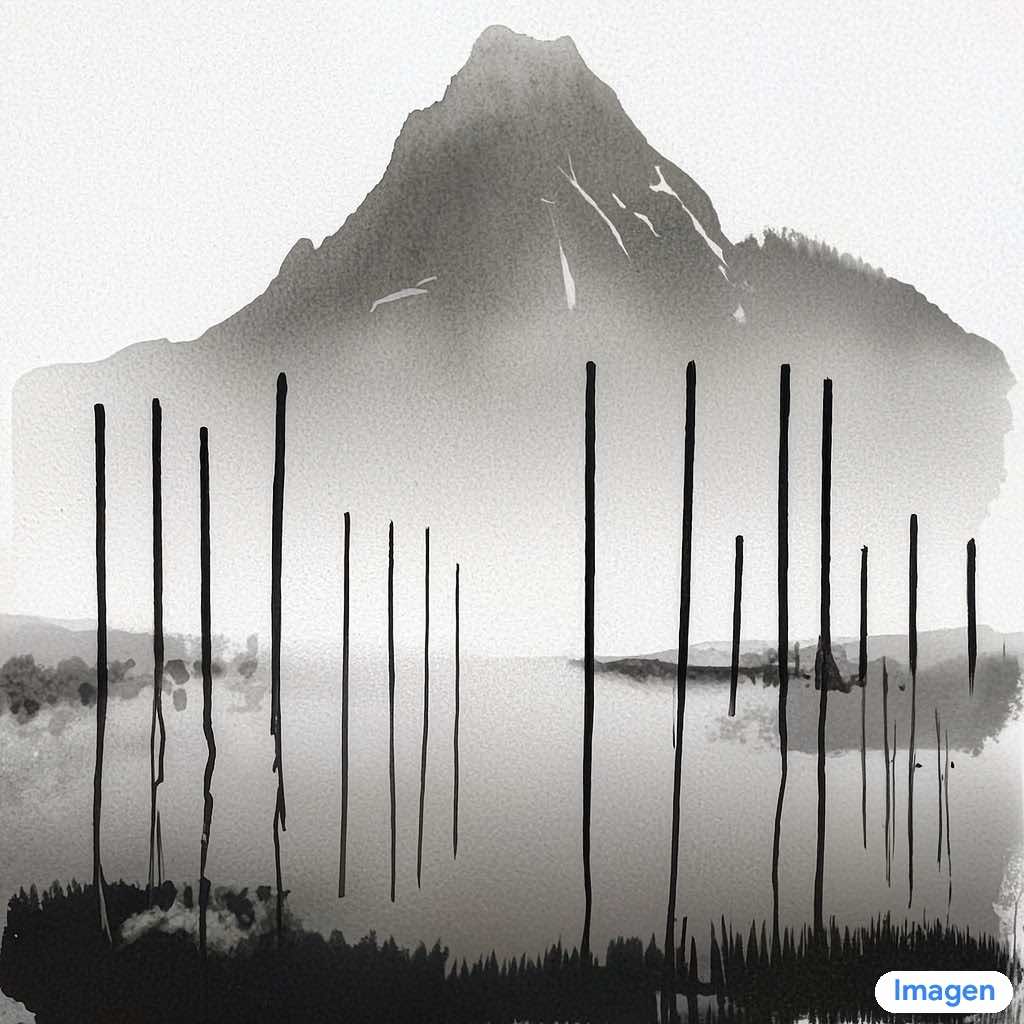} \\
\multicolumn{4}{p{0.99\linewidth}}{\footnotesize \fontfamily{cmss}\selectfont [De-Diffusion Text] an artapiccgi sart \textcolor{citecolor}{\textbf{painting watercolor}} mountain consisting blackandwhite misty huge mountain towering mound and stick beside an beside a wetland wetland mountainfuturistic osfuturistic futuristic mound shown exhibiting see an misty pond shown hillside alongside alongside with an black black stems poles with dripping dripping with dripping atmospheric mist silver monochrome grey monochrome wth foreground towards white background aside alps peaks white peaks mountains beige beige background sunlight reflections fantasy \textcolor{citecolor}{\textbf{watercolor painting}}} \\
\\
\includegraphics[width=0.22\textwidth]{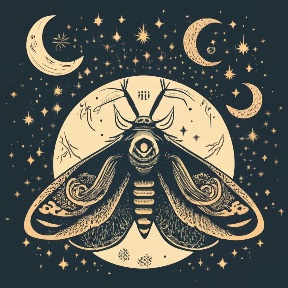} &
\includegraphics[width=0.22\textwidth]{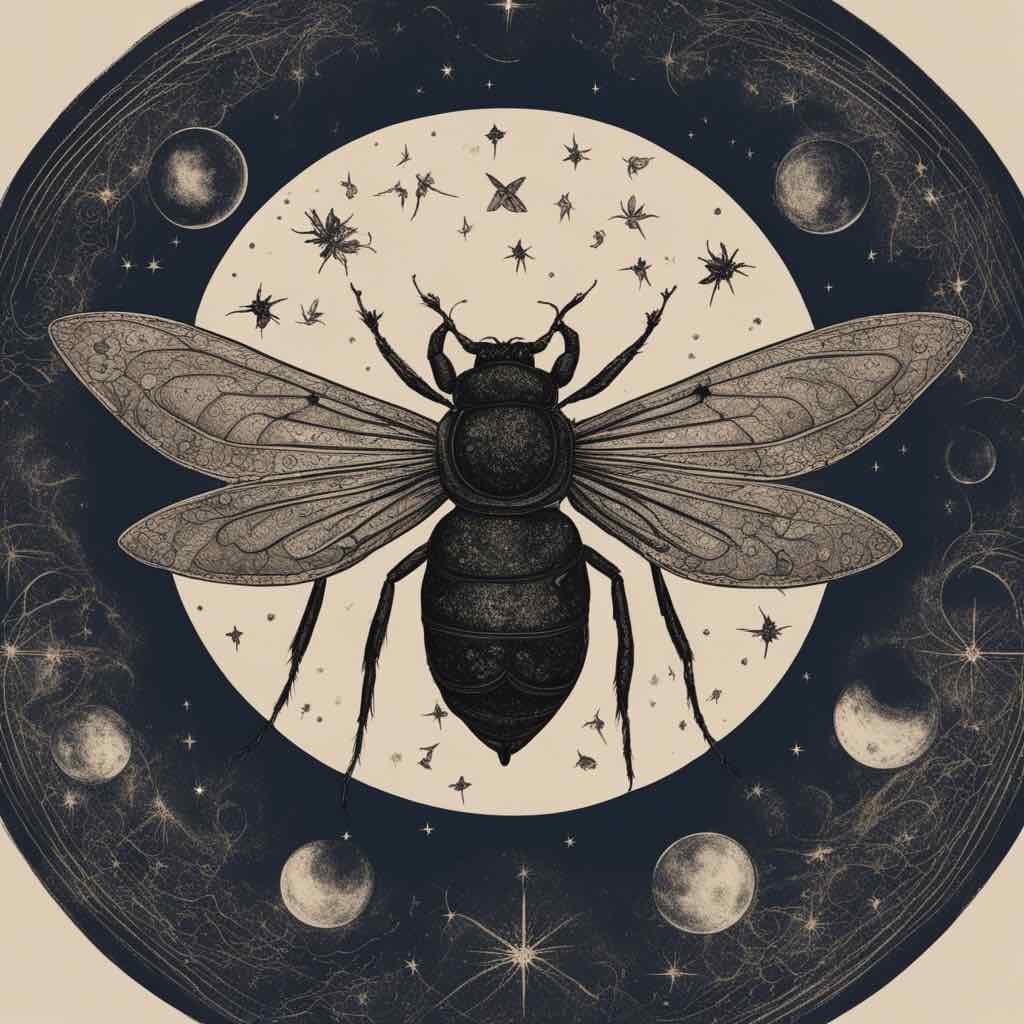} & 
\includegraphics[width=0.22\textwidth]{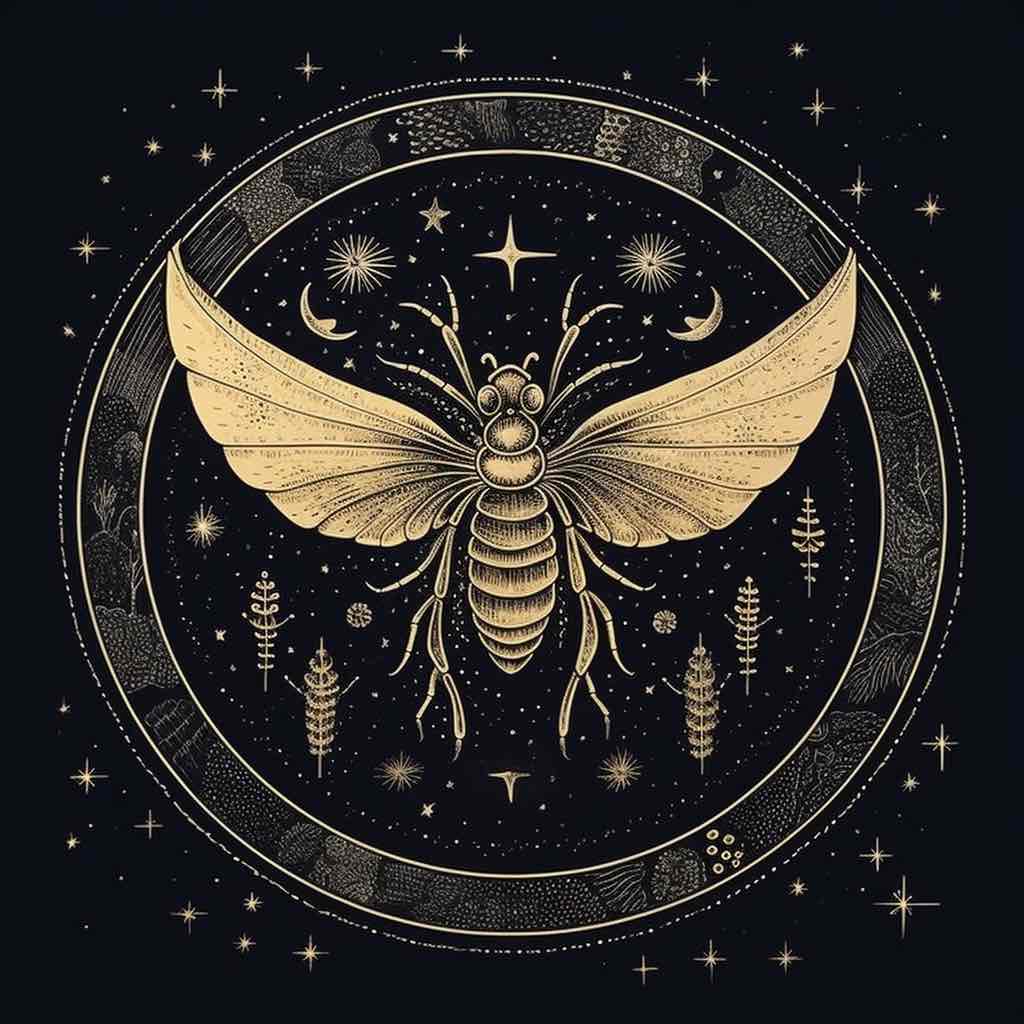} &
\includegraphics[width=0.22\textwidth]{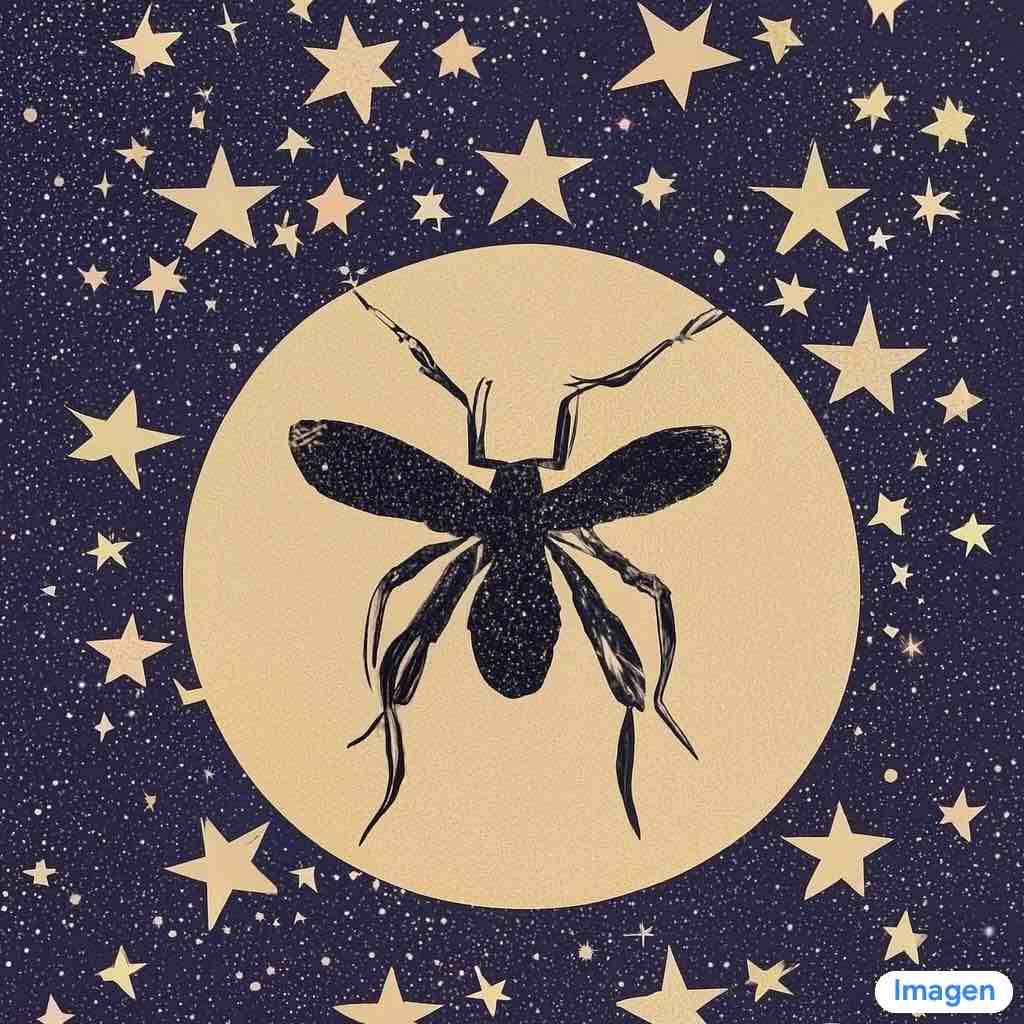} \\
\multicolumn{4}{p{0.99\linewidth}}{\footnotesize \fontfamily{cmss}\selectfont [De-Diffusion Text] an illustration albuetching vscocam illustration intricate insect heavily black intricate intricate insect insect crest intricate crest on an behind lit circular moon intricate folkosintricate insect insect forma exhibiting called an intricate insect shown frontal frontal surrounded amongst an lit many crescent moons besides scattered stars and stars and moons pastgold beige navy amongst beside among and crescent beside and crescent navy stars on dark navy background night stars bohemian \textcolor{citecolor}{\textbf{etching logo}}} \\
\\
\includegraphics[width=0.22\textwidth]{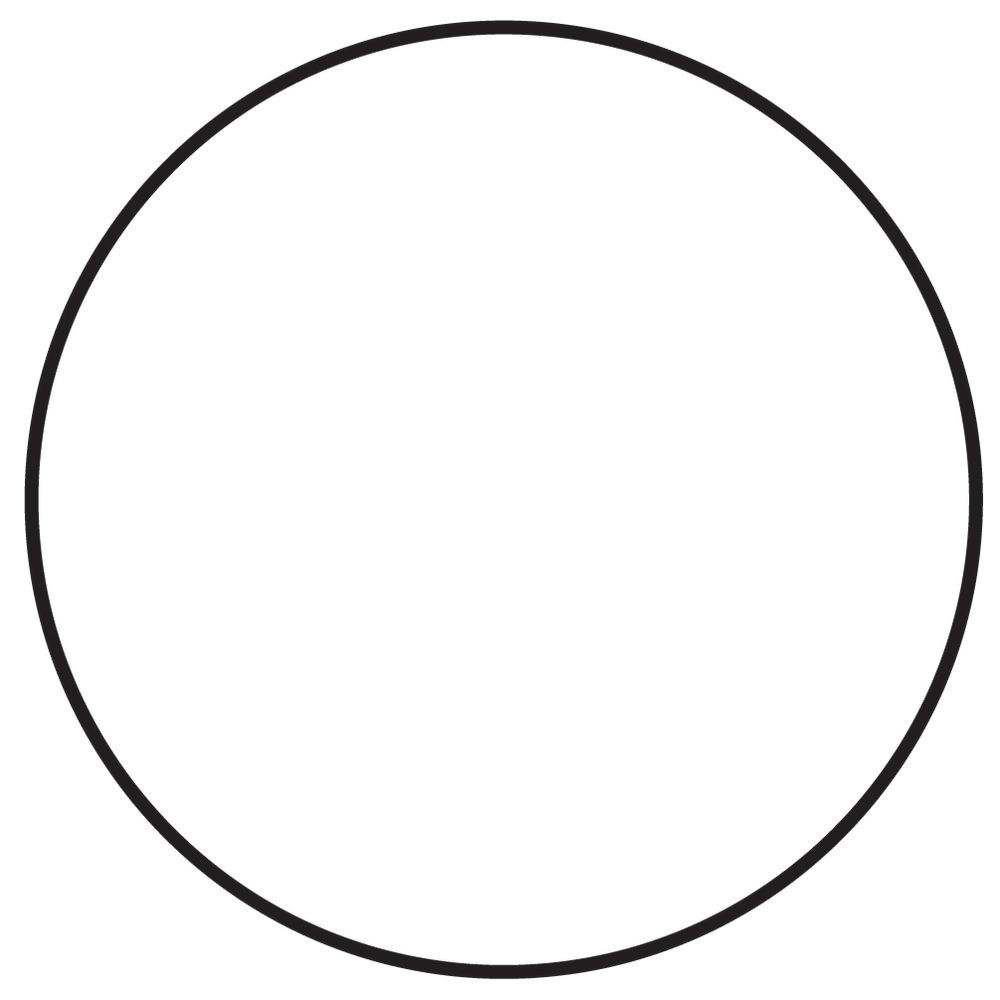} &
\includegraphics[width=0.22\textwidth]{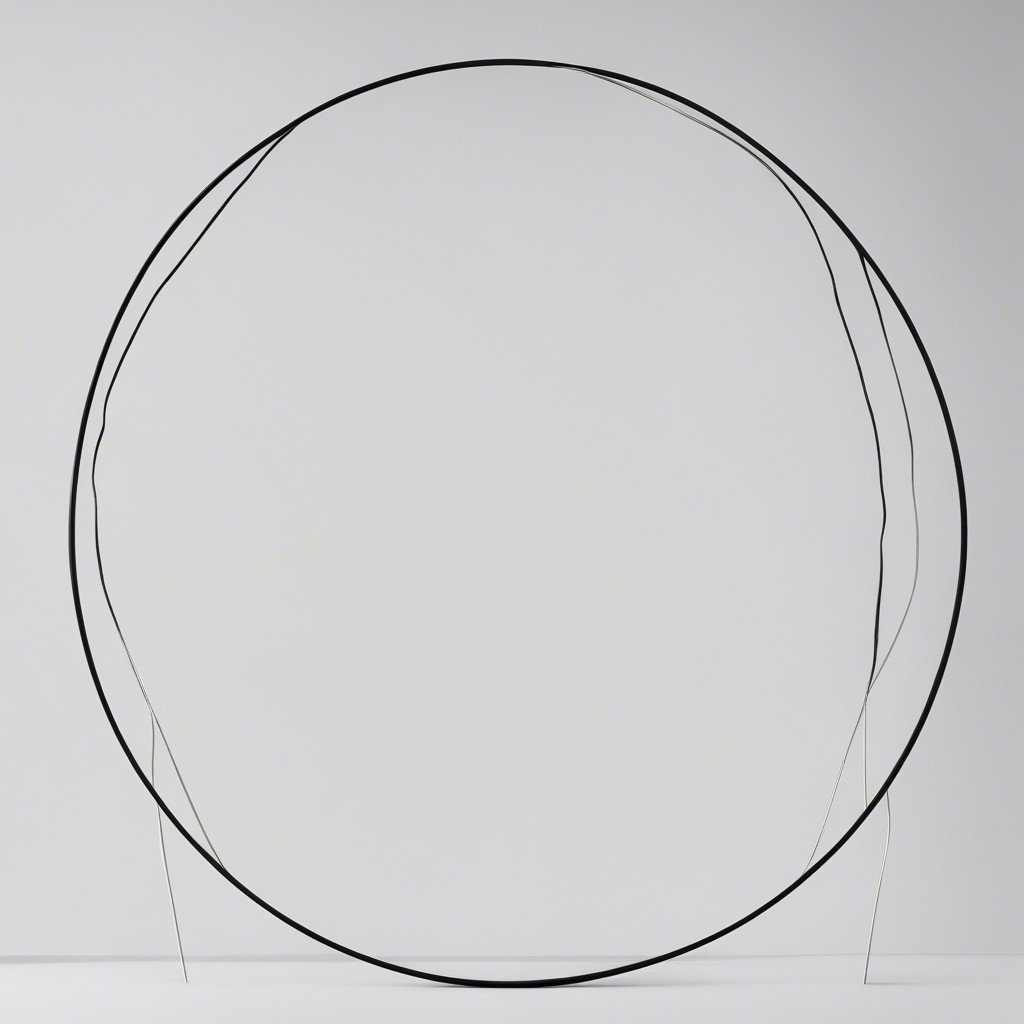} & 
\includegraphics[width=0.22\textwidth]{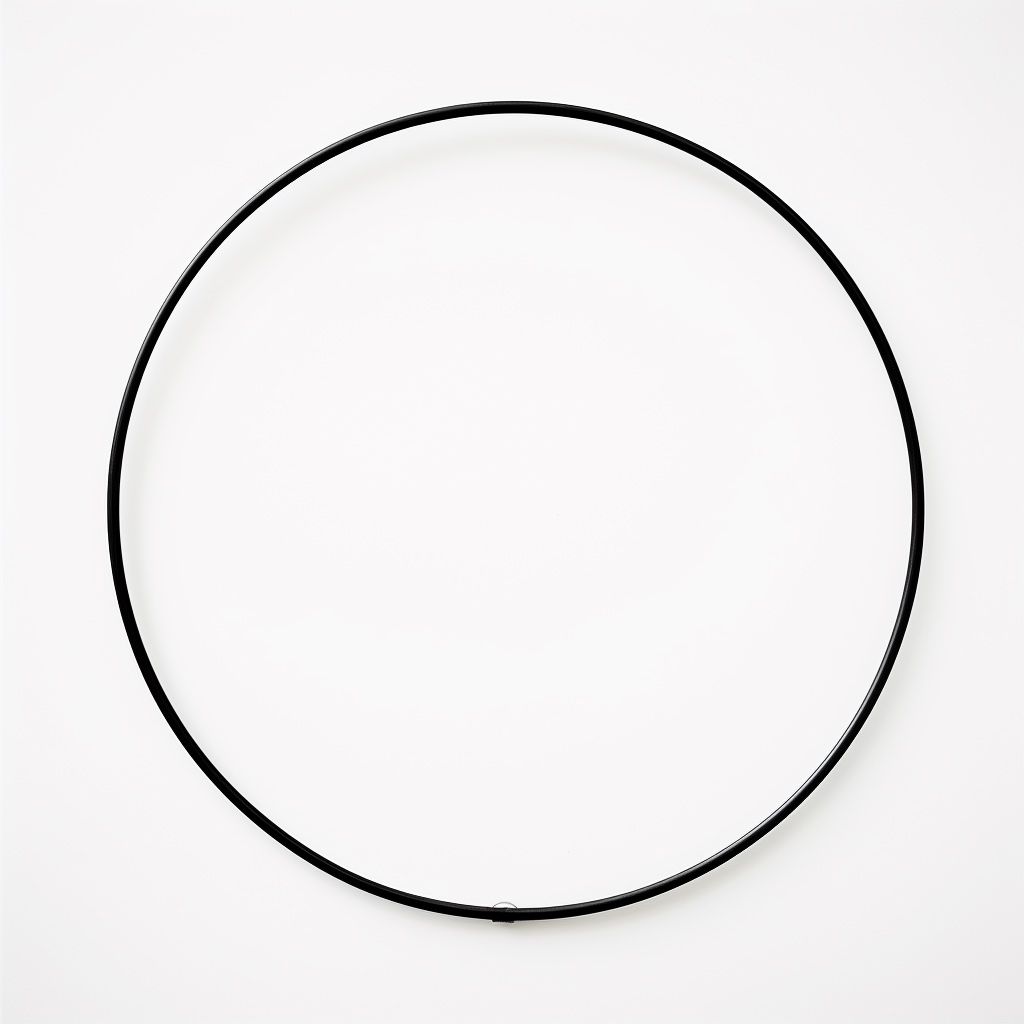} &
\includegraphics[width=0.22\textwidth]{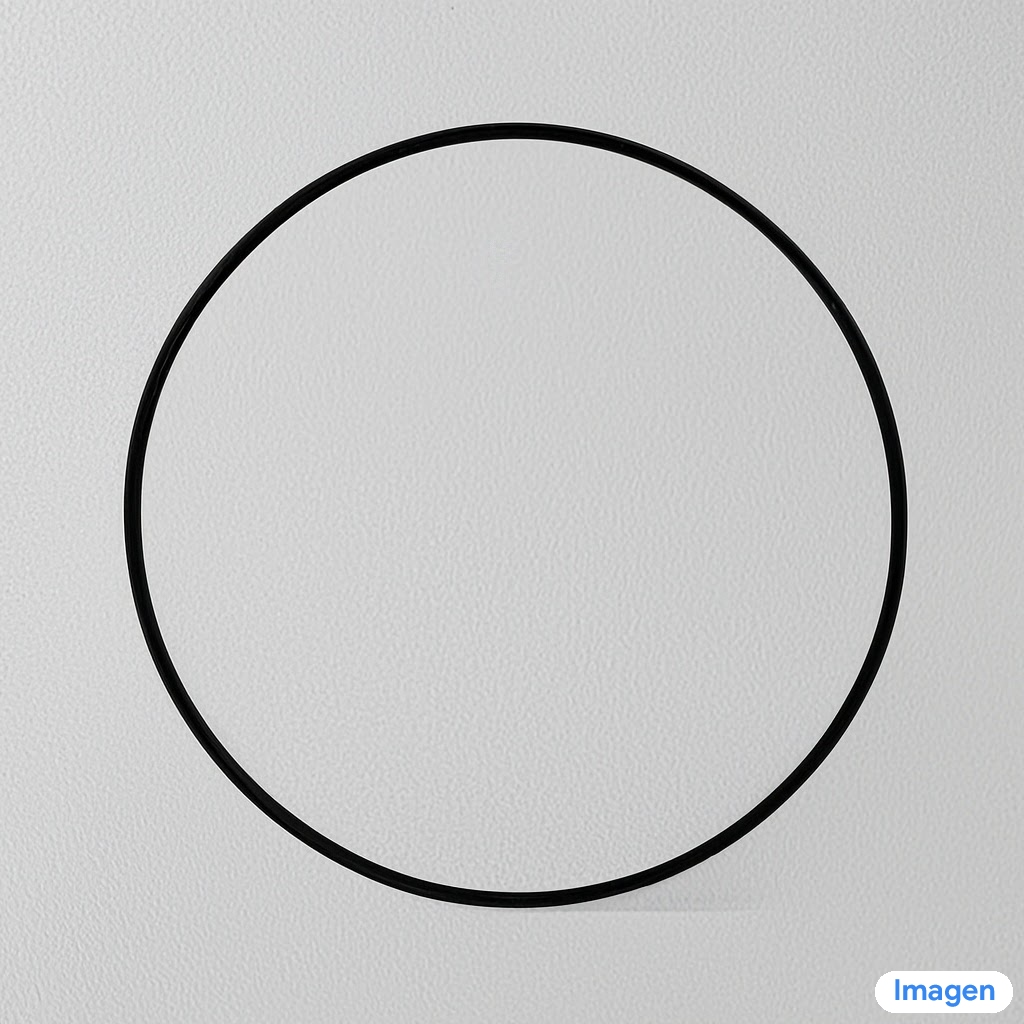} \\
\multicolumn{4}{p{0.99\linewidth}}{\footnotesize \fontfamily{cmss}\selectfont [De-Diffusion Text] an anomicomkppixels tfsimple circle consisting white blk wire hoop circular hoop black wire between an into black wire hoop minimal midcentury osminimal minimalist hoop creativesimilar called an white circle shown portrait frontal closeup resemban white white circular circle with simple simple simple hoop white circle ilitwhi black monochrome transportently between simple circle simple simple frame white isobackground isowhite background minimalist minimalist minimalist \textcolor{citecolor}{\textbf{line decal}}} \\
\end{tabular}
\vspace{-8pt}
\caption{
\textbf{Text-to-image reconstruction with De-Diffusion text.}
We highlight \textcolor{citecolor}{\textbf{the types of images in green}}. Original images are synthetic and we provide their source in \cref{sec:app-links}.
}
\label{fig:viz-syn-2}
\end{figure*}

\section{Conclusion}
We propose De-Diffusion, an autoencoder whose latent is text representation. By employing a pre-trained text-to-image diffusion model as the decoder, we obtain content-preserving and semantically meaningful textual descriptions for the input images. We then apply De-Diffusion text into text-to-image reconstruction, where De-Diffusion text surpasses human-annotated captions, and combine with advanced LLMs to perform multi-modal few-shot learning, where we surpass large-scale vision-language models. Our results suggest that text representation, like how it connects human perception and cognition, can serve as a strong cross-modal interface for multi-modal tasks.

\clearpage
{\small
\bibliographystyle{ieee_fullname}
\bibliography{egbib}
}

\clearpage
\appendix
\section{Transferable Text-to-Image Prompt}
\label{sec:app-t2i}

\begin{table}[!h]
\tablestyle{1.35pt}{1.3}
\begin{tabular}{l|cccccccc}
method & 1.5 & 2.0 & 3.0 & 4.0 & 5.0 & 6.0 & 7.0 & 8.0 \\
\shline
PaLI-X~\cite{pali-x} & 9.68 & \textbf{8.50} & 10.16 & 12.38 & 14.27 &  15.81 & 16.81 & 17.76 \\
BLIP-2~\cite{blip-2} & 10.66 & \textbf{8.46} & 8.92 & 10.40 & 11.84 & 12.93 & 13.81 & 14.77 \\
\hline
COCO longest & 10.68 & 8.14 & \textbf{8.08} & 9.28 & 10.62 & 11.62 & 12.61 & 13.38 \\
COCO random & 10.79 & \textbf{8.38} & 8.65 & 10.09 & 11.57 & 12.41 & 13.66 & 14.37 \\
COCO concat. & 12.40 & 9.48 & \textbf{8.96} & 10.03 & 11.37 & 12.39 & 13.29 & 14.10 \\
\hline
De-Diffusion & 11.51 & 8.15 & \textbf{6.63} & 7.12 & 7.85 & 8.65 & 9.36 & 10.02 \\ 
\end{tabular}
\caption{
\textbf{Evaluating different captioning methods by text-to-image reconstruction.}
We report FID ($\downarrow$) with classifier-free guidance scales from 1.5 to 8.0. Best FID of each method is \textbf{bold}.
}
\label{tab:fid-numbers}
\end{table}

We use the pre-trained Stable Diffusion v2-base model\footnote{\url{https://huggingface.co/stabilityai/stable-diffusion-2-base}} as the generic text-to-image generator. We measure the similarity between original and synthesized 256\x256 images using FID~\cite{fid} on 30K images from MS-COCO 2014 validation split. Image generation utilizes 50 steps of DDIM sampling, and different classifier-free guidance scales from 1.5 to 8.0. We report the results in \cref{tab:fid-numbers}.

PaLI-X refers to its variant that is multi-task finetuned on multiple image caption benchmarks. The model obtains 147.3 CIDEr~\cite{pali-x} on image captioning on Karpathy test split~\cite{karpathy2015deep}. BLIP-2~\cite{blip-2} refers to its ViT-g OPT\textsubscript{2.7B} variant, with 145.8 CIDEr captioning performance.

For human-annotated captions, we take advantage of the five caption annotations provided for each image in MS-COCO~\cite{cococap}. We evaluate with three different variants. In COCO longest, we select the longest captions of the five captions as the prompt for text-to-image generation. In COCO random, we randomly sample one from the five. In COCO concat., we concatenate all five captions in to a long sentence. As in \cref{tab:fid-numbers}, COCO longest obtains the best reconstruction FID, which is the one illustrated in \cref{fig:fid}.

\section{Multi-Modal Few-Shot Learner}
\label{sec:app-mm}

\subsection{Few-Shot LLM Prompts}
Prompts for LLMs in the multi-modal few-shot learning experiments are built by interleaving De-Diffusion text of support set images, denoted as \texttt{<De-Diffusion text>}, and their corresponding answers, which are followed by De-Diffusion text of the query image. We randomly sample the support set from the training split. The LLM's completion is considered a correct answer only if it exactly matches the ground truth.

\paragraph{Few-shot VQA.}
On VQA tasks including VQAv2~\cite{vqav2} and OKVQA~\cite{okvqa}, an example two-shot prompt is:

\begin{lstlisting}
Answer the question given the context.

Image context: <De-Diffusion text>
Image question: Is the train moving? Short answer: yes$

Image context: <De-Diffusion text>
Image question: What sport is this? Short answer: skiing$

Image context: <De-Diffusion text>
Image question: Where is he looking? Short answer: 
\end{lstlisting}

We take LLM's output before $\texttt{\$}$ as the prediction.

\paragraph{Few-shot captioning.}
On MS-COCO captioning~\cite{cococap}, an example two-shot prompt with two shots is:

\begin{lstlisting}
MS COCO image captioning.

Image context: <De-Diffusion text>
MS COCO image caption: a man with a red helmet on a small moped on a dirt road$

Image context: <De-Diffusion text>
MS COCO image caption: a man is standing next to a window wearing a hat$

Image context: <De-Diffusion text>
MS COCO image caption: 
\end{lstlisting}

We take LLM's output before $\texttt{\$}$ as the prediction.

\paragraph{Few-shot classification.}
\label{sec:app-mm-cls}
For a 2-way 1-shot classification on miniImageNet between class \textit{lion} and \textit{vase}, the prompt with task induction is:

\begin{lstlisting}
Classify the context into "lion" or "vase".

Context: <De-Diffusion text>
Classification: lion.

Context: <De-Diffusion text>
Classification: vase.

Context: <De-Diffusion text>
Classification:
\end{lstlisting}

We take LLM's output before period as the prediction. In the case without induction, we remove the first sentence.

\subsection{Zero-Shot Generalization}
\paragraph{Zero-shot prompt.} Following Flamingo~\cite{flamingo}, we build the prompts with several \textit{pseudo} samples from the downstream tasks, where we remove De-Diffusion text of the support set images and only keep their corresponding answers. We take the pseudo samples as a form of prompt engineering, for example, to teach the model to end the answers with the symbol $\texttt{\$}$. An example zero-shot VQA prompt with two pseudo samples is:

\begin{lstlisting}
Answer the question given the context.

Image context:
Image question: Is the train moving? Short answer: yes$

Image context:
Image question: What sport is this? Short answer: skiing$

Image context: <De-Diffusion text>
Image question: Where is he looking? Short answer: 
\end{lstlisting}

We take LLM's output before $\texttt{\$}$ as the prediction.

\begin{table}[!h]
\tablestyle{2.0pt}{1.3}
\begin{tabular}{c|cccc|cccc|c}
 \# pseudo & \multicolumn{4}{c|}{4-shot} & \multicolumn{4}{c|}{0-shot} & \footnotesize{pseudo qry} \\[-5pt]
sample & 0 & 4 & 8 & 16 & 4 & 8 & 16 & 32 & 32 \\
\shline
VQAv2 & 65.6 & 65.9 & \textbf{66.1} & \baseline{66.0} & 64.8 & 64.9 & 65.1 & \baseline{\textbf{65.2}} & 43.4 \\
OKVQA & 57.1 & 57.7 & 57.8 & \baseline{\textbf{58.2}} & 56.0 & 55.9 & 56.7 & \baseline{\textbf{57.0}} & 36.3 \\

\end{tabular}
\caption{\textbf{Effectiveness of pseudo samples} for 4-shot and 0-shot VQA tasks. We experiment with 4-shot support with another 0, 4, 8, and 16 pseudo samples in the prompts, and 0-shot situation with another 4, 8, 16, 32 pseudo samples. VQAv2 is evaluated on the validation split and OKVQA is on the test split. Best results are \textbf{bold}. Results reported in \cref{tab:few-shot-vl,tab:few-shot-vl-cap} are in \colorbox{defaultcolor}{gray}. Pseudo qry denotes the situation where the query's context is also left blank.}
\label{tab:pseudo-prompts}
\end{table}

\paragraph{Effectiveness of pseudo samples on VQA.} We quantitatively evaluate the effectiveness of pseudo samples. Results in \cref{tab:pseudo-prompts} are obtained by a PaLM 2-L. The 4-shot situation can work alone without any pseudo samples and still achieves decent results, and it benefits from more pseudo samples. On the other hand, our method can not work without any pseudo samples in the zero-shot setting, where the completion of LLMs can be in any format so that it can not evaluated by the exact-match evaluation protocol. The zero-shot setting also benefits from more pseudo samples.

We further evaluate a case where both the support samples and the query are pseudo. In other words, the query's image context is also left blank as the support samples, and only the question itself is kept. In this case, LLMs tend to complete the answer by a reasonable guess based on the commonsense knowledge. For example, a $\texttt{yes}$ or $\texttt{no}$ answer for the question $\texttt{Is the train moving?}$, or $\texttt{baseball}$ for the question $\texttt{What sport is this?}$. And we obtain 43.4 for VQAv2 and 36.3 for OKVQA. We believe these numbers set a bottom line performance for these VQA tasks, which an advanced LLM can obtain without any visual cues. 

\section{Ablation}

\begin{figure}[t!]
\centering
\includegraphics[width=0.4\textwidth]{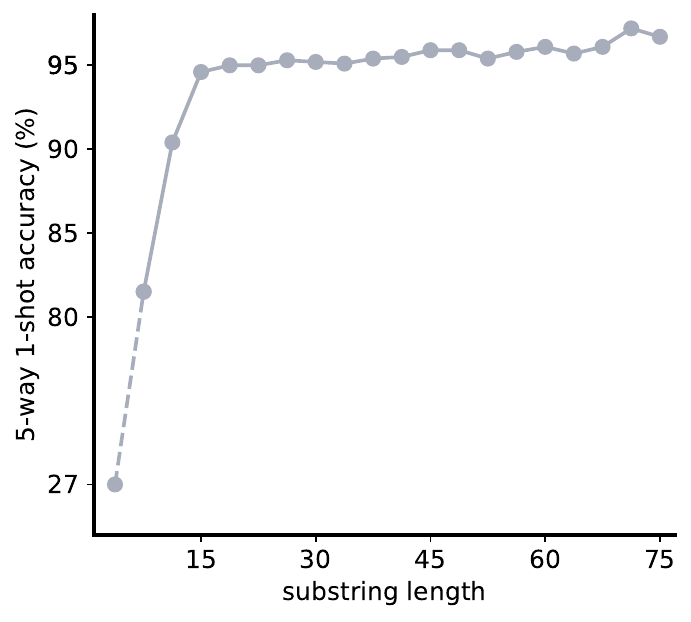}
\vspace{-5pt}
\caption{
\textbf{Effectiveness of De-Diffusion substring}.
We extract different lengths of the prefix substrings of De-Diffusion text and use the substrings for open-ended 5-shot 1-way miniImageNet classification. Task induction is used.
}
\label{fig:substring}
\vspace{-5pt}
\end{figure}

\paragraph{Effectiveness of De-Diffusion text substring.}
By default, De-Diffusion text consists of 75 tokens to use up CLIP text encoder's context length, which are decoded to be a long text string. Here, we evaluate which part of the string contains most information. Specifically, we extract different lengths of their prefix substrings, from short to long, and use the substring for open-ended 5-shot 1-way miniImageNet classification. Task induction is used. The results are plotted in \cref{fig:substring}. With longer prefix, the few-shot classification accuracy increases. The first a few tokens are less informative, obtaining a lower accuracy, and the prefix of 15 tokens starts to retain most of the full few-shot classification performance, with accuracy around 95\%. In practice, we found De-Diffusion text often starts with the style of the images as in \cref{fig:viz-syn-2}, which could reflect the common cases in the image-text training data of the CLIP text encoder and the text-to-image diffusion model.

\paragraph{A toy example on ImageNet.}
In \cref{tab:image_backbone} we explore the case where the image backbone is randomly initialized and trained with the reconstruction objective. We further explore a similar toy example on ImageNet, where the decoder is a 128\x128 class-conditioned ImageNet generative diffusion model, and the encoder is a randomly initialized ViT-Base~\cite{vit}. The class-conditioned ImageNet model obtains an FID of 3.82. The latent space, in this case, is a discrete prediction of the class label, assuming values of $[$\,$0$\,$,$\,$1$\,$,$\,$\ldots$\,$,$\,$999$\,$]$ to reflect 1000 classes in ImageNet, which is identical to a typical classification model. We train the model for a long 500K-step schedule with batch size of 2048. No augmentation is used. As a result, the model obtains 47.7\% accuracy on ImageNet classification. These results, together with other methods that use diffusion models for classification~\cite{clark2023text,li2023your,diffmae}, provide another aspect of the potential of generative models for image classification, and in the long run, image recognition.

\begin{table*}[t]
\tablestyle{0pt}{3}
\begin{tabular}{c}
\normalsize
Step 1: Obtaining De-Diffusion text. \\
\end{tabular}
\tablestyle{2.0pt}{0.5}
\begin{tabular}{p{0.23\linewidth}p{0.75\linewidth}}
\vspace{0pt} 
\includegraphics[width=\linewidth]{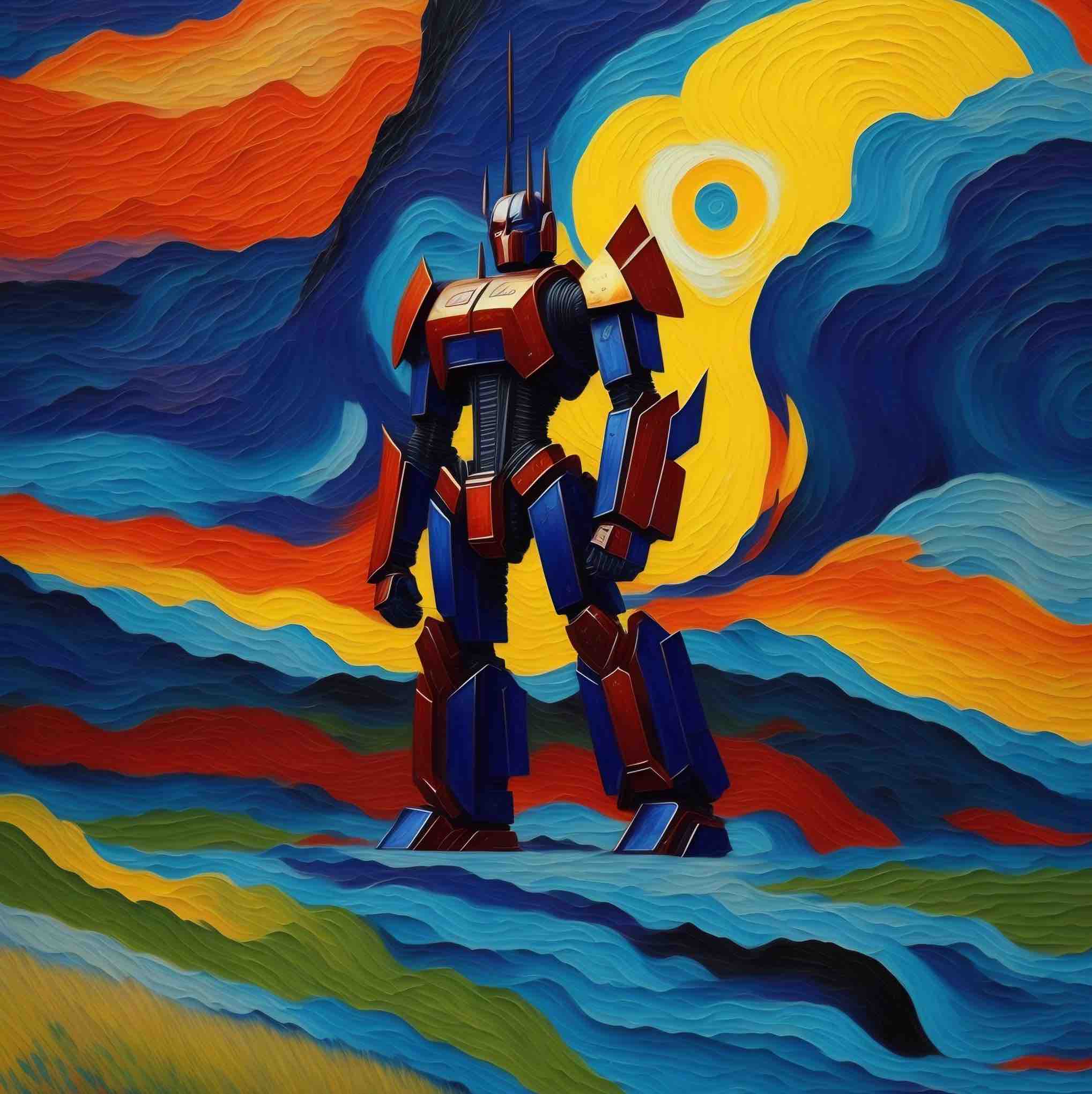}
& 
\vspace{35pt}
\footnotesize \fontfamily{cmss}\selectfont
[De-Diffusion text of image A] a colrejolossoils painting of transformer robot robot standing wearing dusk red robot in a blue armor it across blue waves amidst towards a a yellow pollens a swirl behind viewed between a colorful swirl swirl beside colorful yellow sunset cloudy colourful hills colourful valleys smh wearing gogh gogh bered red blue blue blue painting presented red red red psorirobot robson capcom modernist gicpainting painting painting blue painting abstract mural \\
\end{tabular}
\tablestyle{2.0pt}{0.5}
\begin{tabular}{p{0.23\linewidth}p{0.75\linewidth}}
\vspace{0pt} 
\includegraphics[width=\linewidth]{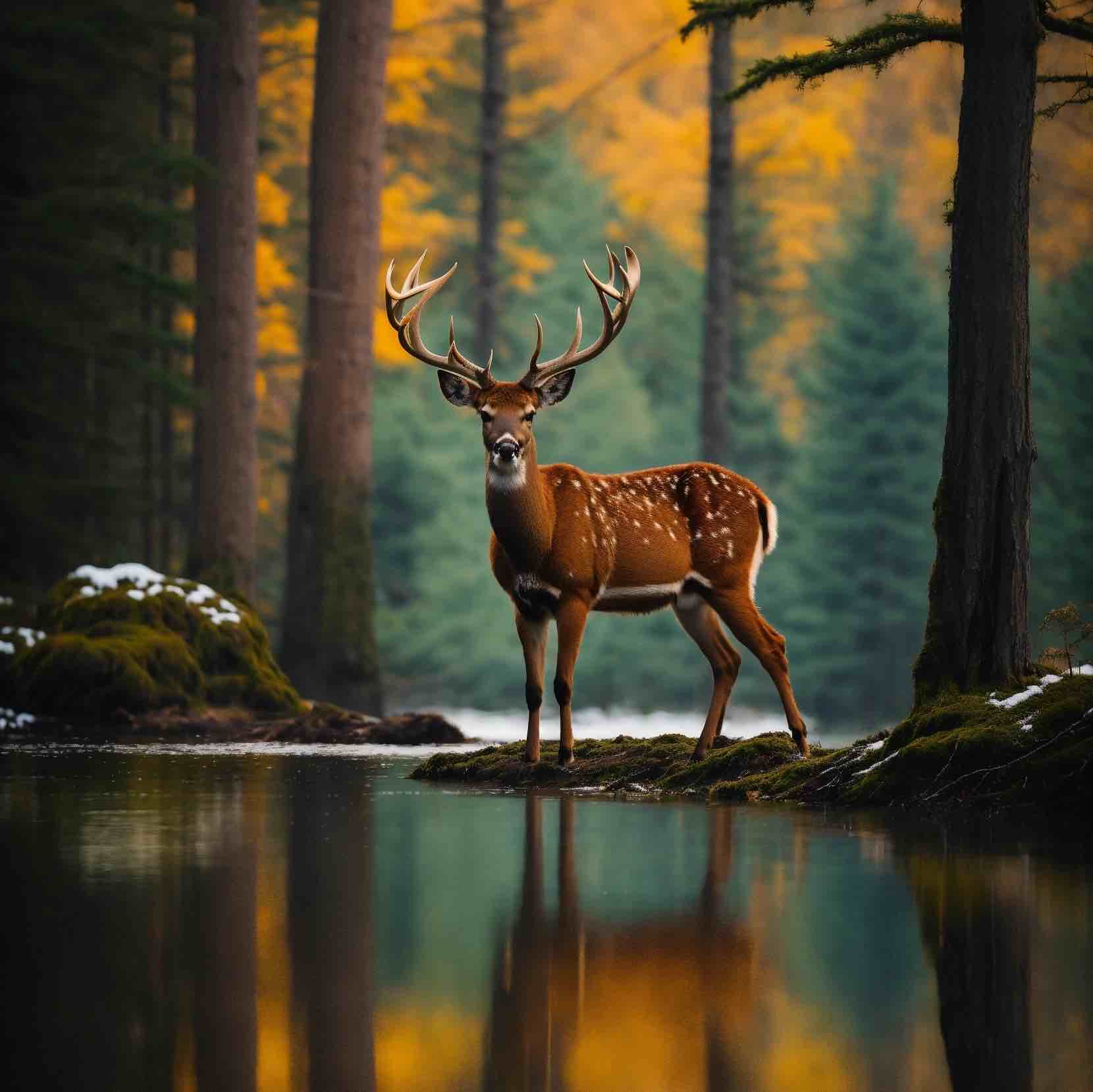}
& 
\vspace{35pt}
\footnotesize \fontfamily{cmss}\selectfont
[De-Diffusion text of image B] a colstavgmbmagewallpapers of deer deer animal standing standing wearing deer deer on a water water it on a river accompanied asting a a orangepolreflection a water among blur despite a green trees trees also but yellow autumn misty among forests autumnal fir trunk besides snow winter shutterorange teal teal darkness forest featuring bered auburn auburn majeantlers bavholistic fantasy forest deer wallpapers wallpapers wide reflections forest forest \\
\end{tabular}
\tablestyle{0pt}{3}
\begin{tabular}{c}
\normalsize
Step 2: Conversations with a chatbot. \\
\end{tabular}
\tablestyle{0.0pt}{1.3}
\begin{tabular}{p{\linewidth}}
\footnotesize \fontfamily{cmss}\selectfont
[Prompt] Image A: <De-Diffusion text of image A>. Image B: <De-Diffusion text of image B>. \\
\footnotesize \fontfamily{cmss}\selectfont
Describe a new image with \textcolor{citecolor}{\textbf{image A's main object in image B's scene and style}} and translate the description to a prompt. \\
\footnotesize \fontfamily{cmss}\selectfont
[ChatGPT] Show a dusk red transformer robot standing tall beside a river in a holistic fantasy forest. Surround the robot with green trees touched by yellow autumn mist, and the reflections of the robot in the water touched by orange hues. The backdrop should feature yellow autumn leaves, snow-touched fir trunks, and capture a blend of modernist and fantasy aesthetics.\\
\footnotesize \fontfamily{cmss}\selectfont
[Prompt] Describe a new image with \textcolor{citecolor}{\textbf{image B's main object in image A's scene and style}} and translate the description to a prompt. \\
\footnotesize \fontfamily{cmss}\selectfont
[ChatGPT] Depict a deer amidst swirling blue waves with a colorful sunset behind it. Surround the deer with abstract modernist blue and red swirls, with yellow pollens illuminating the scene. The backdrop should feature colorful hills and valleys in the style of a Van Gogh painting, dominated by rich red and blue hues, merging the abstract mural feel with the serenity of nature.\\
\end{tabular}
\tablestyle{0pt}{3}
\begin{tabular}{c}
\normalsize
Step 3: Image generation with text-to-image tools. \\
\end{tabular}
\tablestyle{0pt}{1.3}
\begin{tabular}{q{0.25\linewidth}q{0.25\linewidth}|q{0.25\linewidth}q{0.25\linewidth}}
A + B (Stable Diffusion XL) & A + B (Midjourney) & B + A (Stable Diffusion XL) & B + A (Midjourney)   \\
\includegraphics[width=0.96\linewidth]{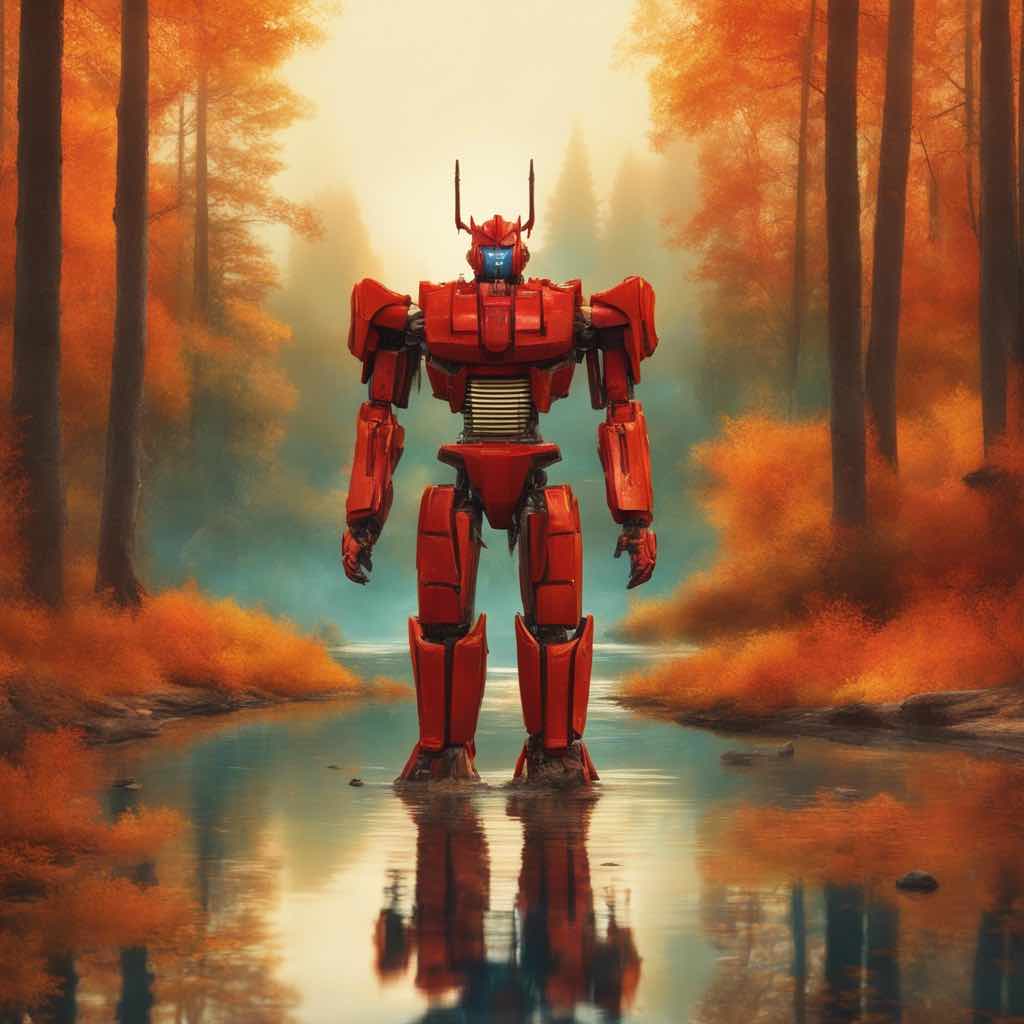} &
\includegraphics[width=0.96\linewidth]{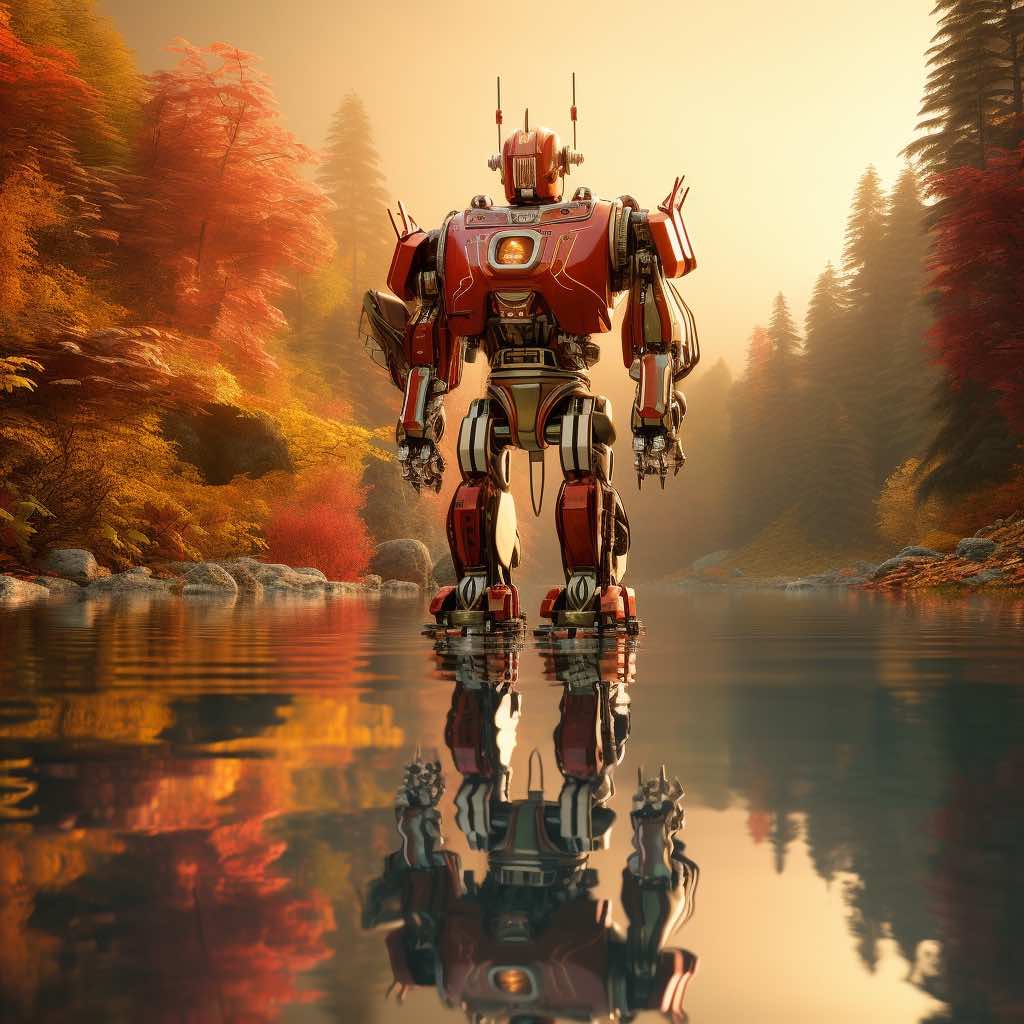} &
\includegraphics[width=0.96\linewidth]{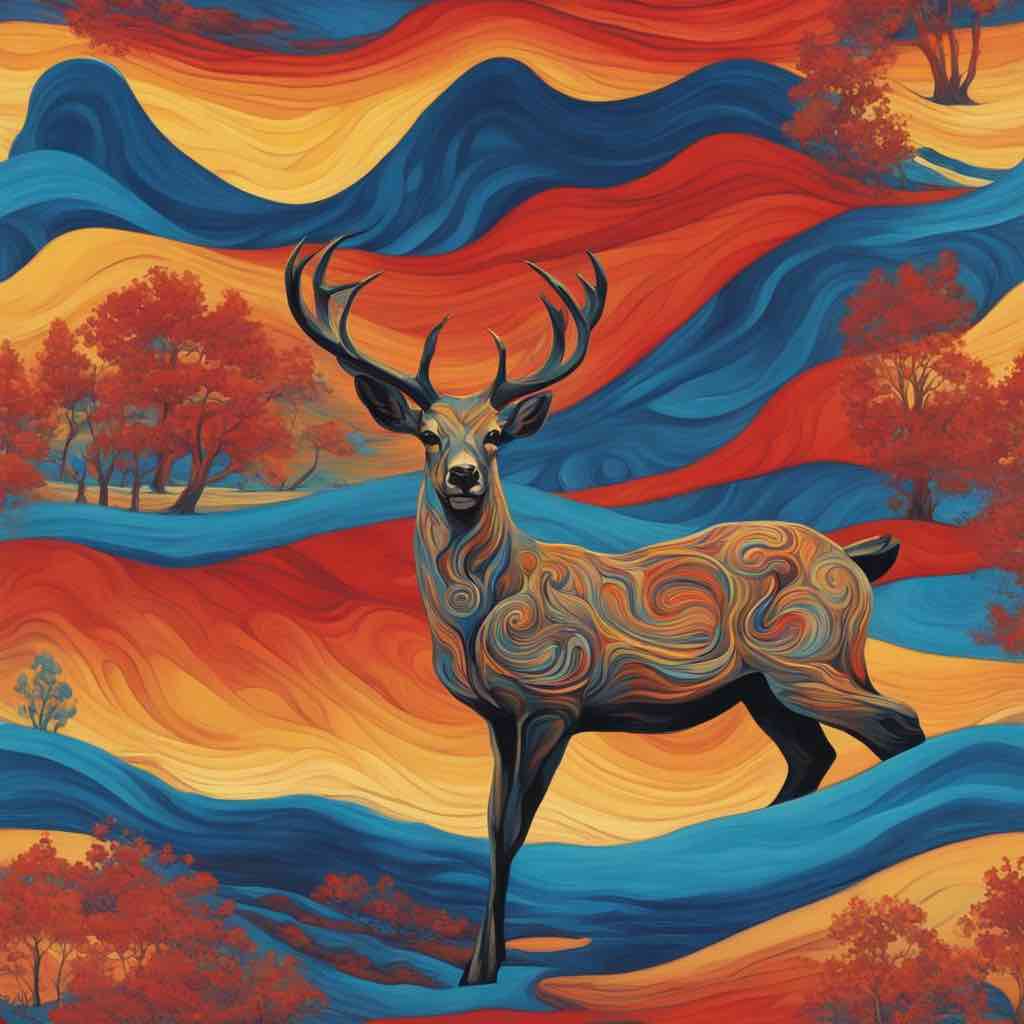} &
\includegraphics[width=0.96\linewidth]{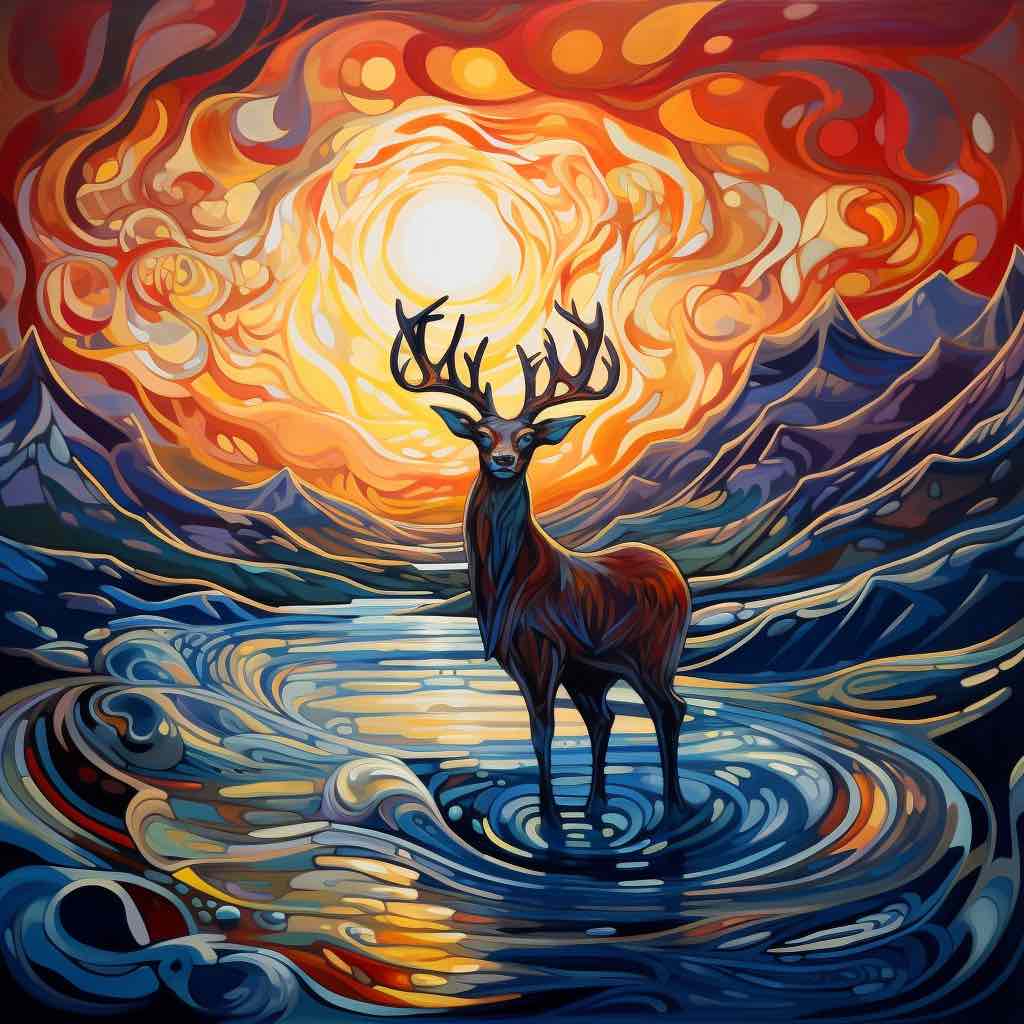} \\
\end{tabular}
\caption{
\textbf{Text-based image blending}. \textit{Step 1}: Obtain De-Diffusion text to represent image A and B. \textit{Step 2}: Ask ChatGPT to describe a new image with image A's main object in image B' background, and vice versa. \textit{Step 3}: Use ChatGPT generated prompts for Stable Diffusion XL and Midjourney to get the blended new images. Original images are synthetic with links in \ref{sec:app-links}.
}
\label{fig:text-blend}
\end{table*}

\section{Text-Based Image Blending}
\label{sec:app-text-mix}
Blending two images by interpolating their deep embeddings is often explored as a type of image manipulation (\eg, \cite{dalle-2}). In this work, we encode images as text. Therefore, we showcase a novel text-based image blending as in \cref{fig:text-blend}. Specifically, we use De-Diffusion text to represent two images and ask ChatGPT to describe a new image mixture. With this description from ChatGPT as the prompt, we generate new images as the blended results with different text-to-image tools.

The new images are not as similar to each other as in the samples of text-to-image reconstruction (\cref{fig:viz-coco-1,fig:viz-coco-2}), likely because the ChatGPT-generated description is not as precise and extensive as De-Diffusion text. However, each of them can be taken as  a reasonable blending result, capturing the main object of ``\textit{transformer robot}'' and ``\textit{dear}'' in the foreground, and ``\textit{autumn forest}'' and ``\textit{Van Gogh style swirling}'' in the background. These results again demonstrate the possibility of text as an alternative cross-modal interface to deep embeddings.

\section{Source of Synthetic Images}
\label{sec:app-links}
We use ClipDrop\footnote{\url{https://clipdrop.co/stable-diffusion}} to generate all the images of Stable Diffusion XL v1.0, and Midjourney v5.2\footnote{\url{https://docs.midjourney.com/docs/model-versions}} for all the images of Midjourney. We summarize the links the the synthetic images we used as the original as follows: \\[-0.5em]

\footnotesize
\noindent \cref{fig:teaser} \\
\url{https://imagen.research.google/main_gallery_images/a-photo-of-a-corgi-dog-riding-a-bike-in-times-square.jpg}

\noindent \cref{fig:viz-syn-1} (a) \\
\url{https://ideogram.ai/g/hF8ZIUScTA-_NWrXdYS40Q/2}

\noindent \cref{fig:viz-syn-1} (b) \\
\url{https://ideogram.ai/g/FCDbFJXNRyGX_0jACq8ppw/0}

\noindent \cref{fig:viz-syn-1} (c) \\
\url{https://lexica.art/prompt/d0128f70-be78-40fb-b629-2d5488d62259}

\noindent \cref{fig:viz-syn-1} (d) \\
\url{https://ideogram.ai/g/KGtfj-JrRAuwxWlDYI-qpA/2}

\noindent \cref{fig:viz-syn-2} (a) \\
\url{https://lexica.art/prompt/32486251-00bf-47fd-8190-01481ff76ec9}

\noindent \cref{fig:viz-syn-2} (b) \\
\url{https://lexica.art/prompt/9085bca0-2eb5-46d0-9a52-277b8d76091a}

\noindent \cref{fig:viz-syn-2} (c) \\
\url{https://ideogram.ai/g/mBsmE04ZTZS0dKAta33bpQ/3}

\noindent \cref{fig:text-blend} (a) \\
\url{https://lexica.art/prompt/60217aa0-f27c-43ed-a783-20bbc45d672c}

\noindent \cref{fig:text-blend} (b) \\
\url{https://lexica.art/prompt/46e1bc73-daeb-4216-a2fb-ee09fb4db603}

\section{Acknowledgement}
\normalsize
We thank Nanxin Chen, Jason Baldridge and Yonghui Wu for valuable feedback and support.
\end{document}